\documentclass{sfuthesis}

\usepackage{amsmath}                            
\usepackage{amssymb}                            
\usepackage{amsthm}                             
\usepackage{graphicx}                           
\usepackage[pdfborder={0 0 0}]{hyperref}        
\usepackage{hyperref}
\usepackage{multirow}
\usepackage[table]{xcolor}
\newcommand{\cmark}{\ding{51}}%
\newcommand{\xmark}{\ding{55}}%
\usepackage{pifont}
\RequirePackage{tipa}
\usepackage{wrapfig}
\usepackage{float}
\usepackage{tabularx}
\usepackage{longtable} 
\usepackage{array} 
\usepackage{cite}
\usepackage{CJKutf8}
\usepackage{booktabs}
\usepackage{caption}
\usepackage{soul}
\usepackage{color}

% ...

\begin{document}

%

%

\title{I Know You're Listening: Adaptive Voice for HRI}
\thesistype{Thesis}
\author{Paige Irène Tuttösí}
\previousdegrees{%
    B.Sc., Simon Fraser University, 2021\\
    B.A. (Hons.), Simon Fraser University, 2016}
\degree{Doctor of Philosophy}
\department{School of Computing Science}
\faculty{Faculty of Applied Sciences}
\copyrightyear{2025}
\semester{Spring 2025}

\keywords{social robotics; L2 teaching robots; L2-tailored TTS; adaptive speech synthesis; L2 speech perception; accessible speech synthesis; expressive speech synthesis; ambient appropriate voice}

\committee{
    \chair{Anders Miltner}{Assistant Professor, Computing Science}
    \member{Angelica Lim}{Supervisor \\ Assistant Professor, Computing Science}
    \member{Jean-Julien Aucouturier}{Committee Member \\ Directeur de Recherche, FEMTO-ST Institute \\Centre national de la recherche scientifique (CNRS)\\ Besançon France}
    \member{Mo Chen}{Committee Member \\ Associate Professor, Computing Science}
    \member{Lawrence Kim}{Examiner \\ Assistant Professor, Computing Science}
    \member{Tony Belpaeme}{External Examiner \\ Professor, Faculty of Engineering and Architecture\\ Ghent University \\ Ghent Belgium}
}

\frontmatter
\maketitle{}
\makecommittee{}


\begin{abstract}
With increased globalization, the desire to learn a new language is putting increased pressure on the need for second language (L2) teachers. Social robots are particularly well suited for teaching tasks as they can create engaging social interactions in a physical form, for example, using action stories (i.e., acting out an action matching the word) when learning new vocabulary. While the use of social robots for language teaching has been explored, there remains limited work on a task-specific synthesized voices for language teaching robots. Given that language is a verbal task, this gap may have severe consequences for the effectiveness of robots for language teaching tasks. We address this lack of L2 teaching robot voices through three contributions: 1. We address the need for a lightweight and expressive robot voice. Using a fine-tuned version of Matcha-TTS, we use emoji prompting to create an expressive voice that shows a range of expressivity over time. The voice can run in real time with limited compute resources. Through case studies, we found this voice more expressive, socially appropriate, and suitable for long periods of expressive speech, such as storytelling. 2. We explore how to adapt a robot's voice to physical and social ambient environments to deploy our voices in various locations. We found that increasing pitch and pitch rate in noisy and high-energy environments makes the robot's voice appear more appropriate and makes it seem more aware of its current environment. 3. We create an English TTS system with improved clarity for L2 listeners using known linguistic properties of vowels that are difficult for these listeners. We used a data-driven, perception-based approach to understand how L2 speakers use duration cues to interpret challenging words with minimal tense (long) and lax (short) vowels in English. We found that the duration of vowels strongly influences the perception for L2 listeners and created an ``L2 clarity mode'' for Matcha-TTS that applies a lengthening to tense vowels while leaving lax vowels unchanged. Our clarity mode was found to be more respectful, intelligible, and encouraging than base Matcha-TTS while reducing transcription errors in these challenging tense/lax minimal pairs.
\end{abstract}

\begin{dedication}
This thesis is dedicated to my soul, my feelings of hope and joy, and any love that I once had for academia and research. You were once great friends who allowed me to have a sense of purpose in my life, but sadly you did not survive the long and harrowing journey of my doctoral studies. You are greatly missed and I write this thesis in your loving memory.

\end{dedication}

\begin{acknowledgements}
First, I would like to thank my dogs Ilona and Nymuë, who put up with missing walks for me to work all the time but also keep me sane. I want to thank my cat, Lily. She didn't make the move to France, but she has literally been with me for the entire 15 years of my university career and has put up with all of it. It may sound silly, but really, they all kept me going through all of this.

I want to thank my supervisor, Dr. Angelica Lim. I know so many people who have had awful experiences with their graduate studies, getting in fights and not feeling supported... I may have had some awful experiences, but not a single one of them could ever be due to Angelica. She is incredibly supportive, which is probably hard to do with me because I can be stubborn, thinking I know the best way to do something. I wanted to do crazy things, change my mind about what I was doing, do tons of work in linguistics, leave for France, and not come back. She supported me and not only supported but championed me through all of it. Without her, I would not have my connections, my job, or anything I have coming out of my PhD. She is really everything a supervisor should be, and I wish everyone could have this kind of experience for their graduate studies. She works so hard to be an amazing researcher, teacher, woman in STEM activist, supervisor, mother to her daughter, and second mother to all her students (she probably worried more about me getting a gun waved at me at a conference in Detroit than my own mother did). I am so lucky to not only have her as my supervisor but also to have her in my life in general.

I would also like to thank Dr. JJ Aucouturier, who allowed me to invade his lab and his country and never leave. Letting me argue with him about how I think things need to be a certain way and letting me complain about French bureaucracy. You have been such a wealth of knowledge in a domain I knew hardly anything about before I arrived here, and that help has been immeasurably beneficial to my PhD. I want to thank Dr. Mo Chen, who, despite being from a discipline slightly removed from my own, has been here throughout my PhD, giving me helpful feedback and direction along the way. Thank you to Dr. Henny Yeung and Dr. Yue Wang for giving me such excellent feedback and providing references and guidance as I stumbled my way through learning about linguistics.

I want to thank Shivam for allowing me to message him incessantly about Matcha-TTS and TTS in general. He is the reason why I can even call myself a TTS researcher now; I have learned so much from his help. I would like to thank Emma for letting me help her with her ambiance paper; this basically kickstarted me knowing what I wanted to do with my PhD. Also, thank you for the general conversation and venting; it has been pretty... pretty... pretty good. I want to thank all my other Rosie lab mates, Bita, Shay, Bermet, Yasaman, Zhitian, Payam, Boey and Zach, and Aynaz, and Zhenxing in Neuro-Group who have helped me with projects, figuring out how to sort out my thesis, use solar and a myriad of other support. A special shout out to Rudra, who has helped to leave the lab early and give my dogs medication so I can go into Paris for work. Thank you to my colleagues, Dr. Thomas Janssoone, Dr. Mohammed Hafsati, Dr. Waldez Gomes, and Lucas, who have helped me learn how to integrate my work into our robots, given me helpful feedback, and helped me run my studies. I also want to thank Dr. Anita Tino, who is the only reason I made it as far as I did in engineering and helped me see how to be a great teacher for challenging topics. Also, a thank you to the people in my past lives, Dr. Hugo Cardoso and Shannon Wood. Even though I left archaeology behind, the 8 years I spent digging in jars of teeth cataloging human remains for repatriation has profoundly impacted the rest of my career. There is also a plethora of people in the HRI and speech communities that have given me useful support, feedback and direction, and I am sure no matter how many people I name I will forget someone, so thank you to all my mentors and peers for the engaging discussions and invaluable feedback.

I would also like to thank all the undergrads I have supervised, without whom I would never have finished so many great projects: Blackfoot Team: Danny and Liyang, Citizen Science: Eric, Lina, and Julie, and BERSt: Maya, Mantaj, Shawn, Flora, Avni, Poorvi, Luna, and Minh.

I would also like to thank my friends and partners throughout my studies. My best friend Elle who is always there to talk me off a ledge, support me when there is a real problem, but also tell me when I am being absolutely ridiculous and unreasonable. She is also always there to tell me to work less, even though she knows I will never listen. Thank you to Correy, who is basically the only reason I know how to code and code well. Everything I know, and everything that makes me a good coder I learned from him. Thank you to Paul, who, in classic French fashion, will blatantly tell me with no sugar coating when my results are bad but will also listen to 100 hours of my voice saying the same sentence over and over again until I get it right. Finally, thank you to my parents, who may not think academia is the best route but are happy to see me succeed either way; whenever I want to give up, I hear my dad's voice in my head (when I failed chemistry my first semester at SFU 15 years ago), ``do you really want to give up... you know where people who give up go... to Subway School of Sandwich Making, that's where.''

\end{acknowledgements}

\addtoToC{Table of Contents}%
\tableofcontents%
\clearpage

\addtoToC{List of Tables}%
\listoftables%
\clearpage

\addtoToC{List of Figures}%
\listoffigures%
\clearpage

%
%

\mainmatter%

\chapter{Introduction}

\begin{figure}[t]
  \centering
  \includegraphics[width=\linewidth]{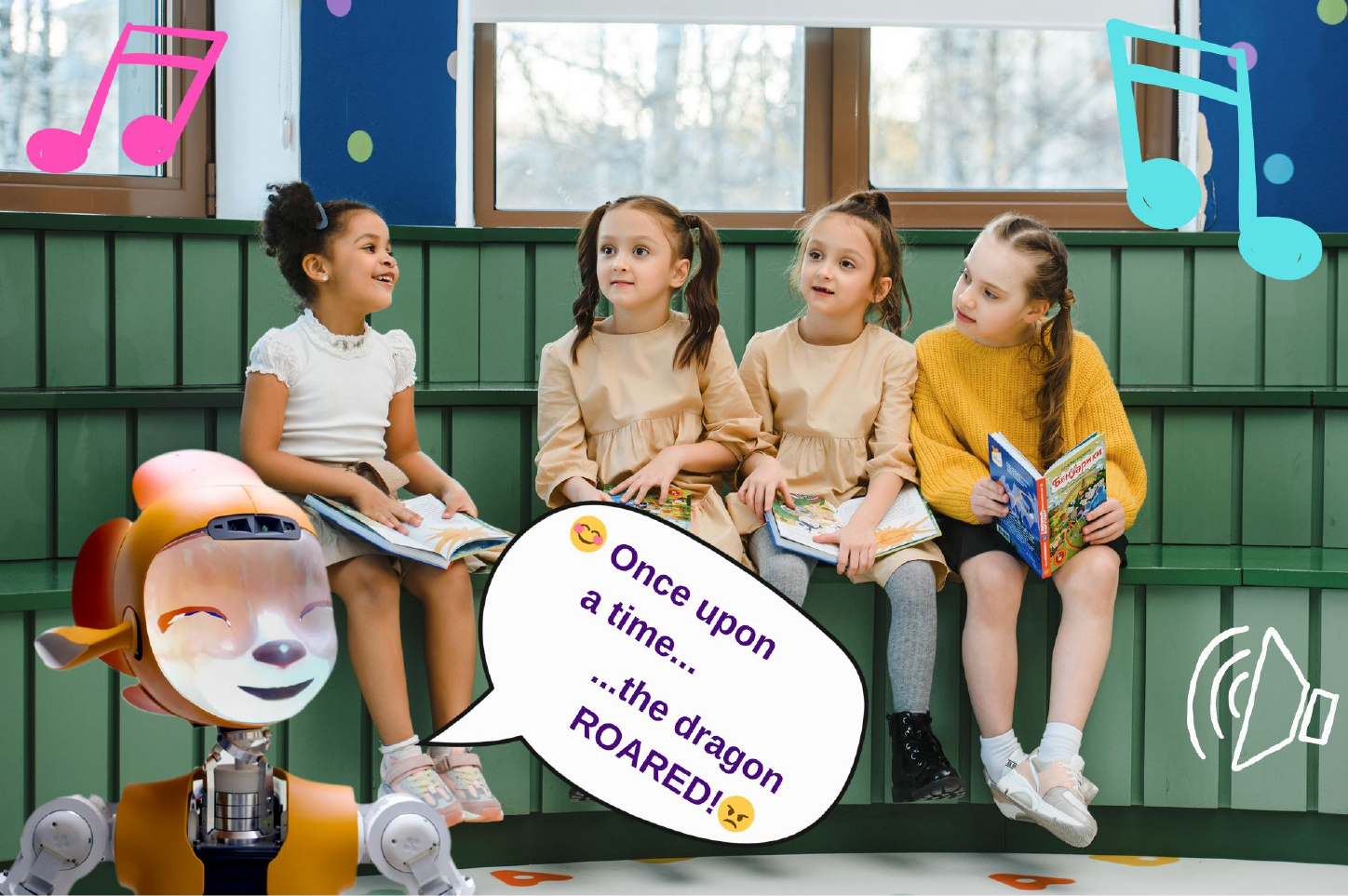}
  \caption{Example deployment of our TTS system. The Miroka Robot is using emoji's to add expressivity when reading a story to children in a high energy, noisy environment. At the same time it will adapt it's voice to be heard over the background noise and music.}
\end{figure}

Imagine you are an immigrant family arriving in a new country. There are pressures to have your children learn the new language and conform to societal norms, but you also want to maintain your cultural heritage and preserve your native language. An important choice that these families make is whether to continue to speak their native language in their home or to begin speaking the language of their new country. Researchers have found that although immigrant families perceive their native languages positively and express a desire for their children to continue speaking their native language, they worry their children will be isolated, struggle with their peers, and fall behind in their education \cite{MucherahWinnie2008IPot}. Moreover, the maintenance of a native language necessitates enormous commitment and encouragement from both parents \cite{doi:10.3138/cmlr.58.3.341} to both transfer language knowledge and nourish a positive sentiment towards their first language. As such, there is a growing need for support in learning new languages and the maintenance of native languages not only among immigrant communities but also in the case of bilingual countries such as Canada, where a shortage of French language teachers is ubiquitous across the country \cite{shortteachers, shortteachers2}, and in the perseveration of Indigenous languages \cite{trushrec, blackfoot, piegan}.

For several years, the robotics academic and industrial communities have proposed that social robots would be particularly apt for use in language acquisition, as this task is most effectively learned through social interactions \cite{MOHAMADNOR2018161, hrisurvey, learninghri}. There has been rather extensive academic work on the use of social robots for language learning; for a survey, see \cite{hrisurvey}, and for a large-scale project, see L2TOR\footnote{http://www.l2tor.eu/}. Yet, there has been little to no work on improving these robots' voices within robotics and speech communities. Speech is the primary means of social communication among humans. Therefore, it is particularly surprising that little care has been given to designing voices specific to second-language (L2) listeners for teaching tasks. Often out-of-box commercial or on-robot text-to-speech (TTS) \cite{6301971, catchcur,nagata10_l2ws, 1652373, 5e2c9c02-ef39-3495-94bd-bbe390e50dff, Lee_Noh_Lee_Lee_Lee_Sagong_Kim_2011, Mubin2013RobotAL}, or recorded human speech is used \cite{7745196, Kanda01062004, 6100846, Westlund2015LearningAS, KORYWESTLUND20171} for robot assisted learning. These TTS systems are always assessed by native speakers, and although some work has been done to improve clarity, it is always in the context of background noise \cite{10.1145/1228716.1228732, 10.1007/978-3-030-98358-1_43, novitasari21_interspeech, cohn20_interspeech, 8343873} and not listener comprehension or language ability. To the best of our knowledge, only one previous work has looked at how to adapt a robot's voice to a user's comprehension abilities and ambient noise \cite{renclarity}. Yet, they did not look deeply into mechanisms to support second-language comprehension specifically.

To improve the perception of synthesized voices for second-language problems with TTS voices need to be addressed to aid the comprehension of L2 listeners. The first problem is the technical limitation of deploying voices on robots in the wild; we need voices that are stable, fast, and lightweight; second, we must have an expressive voice that is able to engage users; third, we need the voice to be usable in multiple ambient and social contexts while remaining audible and appropriate; and fourth, we need a voice that is specifically tailored to the comprehension capabilities of L2 listeners.

\section{Problem 1 - Hardware Limitations}

Robotic platforms are frequently limited by hardware and connectivity, making state-of-the-art speech models impractical for deployment. As such, hardware capability continues to be a bottleneck in the explosive field of artificial intelligence (AI), where new, faster, and more efficient systems are often achieved at the expense of reliability, security, and cost \cite{gnad2024}. Many social robots deployed in HRI studies (e.g., Nao and Pepper: Aldebaran, United Robotics Group; iCub: Istituto Italiano di Tecnologia; LOVOT: GROOVE X; Furhat: Furhat Robotics) do not have an on-board GPU or TPU. A new generation of robots with GPUs is becoming available (e.g., Reachy: Pollen robotics; Mirokaï: Enchanted Tools), yet less than 20 (7\%) of the 263 robots available on IEEE Robots database\footnote{\url{https://robotsguide.com/}} specify that they are equipped with GPU hardware. Given that robots often run many systems at once (navigation, balance, vision, grasping, speech, and more), deploying foundation models with an enormous number of parameters on-board is often infeasible. AI models deployed on robots ideally have a small footprint and require a limited amount of resources at inference time \cite{aisize, distai}; it takes time and careful planning to balance the computational usage of all these models simultaneously. In addition, the ecological footprint of cloud computing \cite{green} and the production of chips needed for powerful GPUs \cite{chips} has become increasingly concerning. To reduce costs and the use of power and resource consumption for an ecologically and economically sustainable future in robotics, we must strive to develop smaller and more efficient AI models.

Hardware limitations are often circumvented through the use of cloud computing. Although this may be an adequate solution for in-lab research deployment, in-the-wild, one cannot always rely on a steady nor fast internet connection. This can lead to latency or a complete inability to deploy interactive systems in-the-wild. In addition, using cloud resources can result in several security and privacy concerns \cite{privacy, MANIAH20191325}. Both subscriptions to cloud services and the inclusion of powerful hardware increase the cost of a robot, increasing the barrier to entry and decreasing the equity of using powerful models in robotics. There is, therefore, a pressing need for lightweight, efficient, and customizable speech models tailored for robotics.

\emph{To address this limitation, we introduce the EmojiVoice toolbox. This free, open-source toolkit enables social roboticists to build their own custom TTS with only 3 minutes of training data per voice style. The TTS is small (20.9M parameters with a 78MB checkpoint), efficient, and runs in real-time, even on a CPU.}

\section{Problem 2 - Expressive Voice}
It is known within pedagogy that immediacy (i.e., a decrease in the perceived psychological distance between a teacher and their students) and expressivity, both verbally and non-verbally, can increase both affective and cognitive outcomes in the classroom \cite{learning1, learning2, learning3}. This has been confirmed for virtual and robotic teaching agents as well \cite{robotlearn1, robotlearn2, robotlearn3, robotlearn4}. Yet, in these studies, similar to other studies of expressivity in robot voices outside of the context of teaching \cite{torre}, the range is often limited to changes in intonation and out-of-box voices, such as those built into Nao and Pepper or commercial TTS such as Amazon Polly\footnote{https://aws.amazon.com/polly/}. These voices may contain, for example, only one expressive ``joyful'' voice or claim to be ``expressive'' without providing the user with control over this expressivity. The use of a high-pitched and ``joyful'' voice\footnote{http://doc.aldebaran.com/2-4/naoqi/audio/altexttospeech-tuto.html?highlight=joyful} is common when attempting to give the impression of an expressive robot \cite{hennig2012expressive}. Although it is true that more engaging speakers use a higher-than-average pitch and greater pitch range \cite{NIEBUHR2016366}, joy is only a single expression, and humans do not convey expressivity through one single emotion instead, they change their expression over time \cite{clark2019makes, cowie2003describing}. Moreover, without any fine-grained control over the expressivity of these voices, tasks requiring extended periods of speech, which may commonly occur in a teaching environment, e.g., telling a story or giving a lesson, will use the same tone of voice throughout the interaction, without considering the appropriateness of this generalized ``expressivity'' for the phrase nor the task.

The expression and interpretation of emotions is an integral part of human speech \cite{beller10_speechprosody}, and the use of expressive speech not only contributes to a more pleasant interaction but also increases engagement and learning outcomes in a classroom environment. Although ``expressive'' and ``emotional'' TTS exist, they have limited controllability, and their use in complex and long-term interactions is poorly understood. Moreover, they are often not customizable, not allowing users to add their own expressive styles relevant to their use cases. There is a need for a synthesized voice that is not only ``joyful'' but temporally expressive, selecting the correct expressive style given the task and the linguistic context and understanding how and when different expressive styles should be deployed.

\emph{To create a voice that remains expressive over time, we built several temporally variable, expressive TTS voices using our EmojiVoice toolbox. In addition, we explored, through case studies, when and where temporally variable expressive and singularly expressive voices are most effectively deployed.}

\vspace{10pt}
\noindent \fbox{\begin{minipage}{\textwidth}
These first two problems are addressed in our in-review article: \emph{EmojiVoice: A TTS Toolkit for Real-time Expressive Speech on Robots, submitted to Robotics and Automation Letters} \cite{emojivoice} and a demo: \emph{Take a Look, it’s in a Book, a Reading Robot in the Companion Proceedings to IEEE/ACM International Conference on Human-Robot Interaction 2025} \cite{reading_robot}.
\end{minipage}}

\section{Problem 3 - Social and Ambiance Adaptation}

To ensure a robot voice can be deployed in multiple environments, from a child's bedroom to a noisy classroom, we must understand what adaptations improve the perception of robot voices in context. When it comes to humans, from the moment we wake up to the moment we go to sleep, our day contains a variety of social situations in different contexts. For example, our first stop may be to the local bustling café, picking up our usual coffee to start the day. The end of the day may involve a date night at a fancy romantic restaurant or a loud, rhythmic nightclub. In either case, these environments contain different lighting and music to set the mood and people with whom we, hopefully, have the pleasure of interacting.

In all such situations, humans have the unconscious ability to adapt their voice appropriately to the different physical and social characteristics of the environment. A significant portion of adaptive communication results from non-linguistic vocal features that can alter the meaning of phrases \cite{maruri21_interspeech}. A waiter may ask a customer, ``Can I take your order?'' and this phrase may carry a different connotation depending on whether the waiter asks the phrase in a casual cafe during a rush or an upscale bar with highbrow clientele. As such, the ambient surrounding plays a role in how one changes their voice to convey meaning to others, and being able to reproduce and interpret these non-linguistic features plays a significant role in human social intelligence.

 The need for a robot to adapt itself appropriately in physical and social ambient environments proves crucial when integrating robots into humans’ everyday lives \cite{context_and_robot_voice,progress_on_robotics_lit_review,what_makes_a_robot_social}. We suspect that correct adaptation could improve both the robot's perceived awareness and intelligence. Yet, the exact way to adapt robotic voices has not been thoroughly explored due in part to the difficulty in collecting clean recordings of realistic data in noisy and crowded environments, i.e., to separate voice and background noise. As such, source voice data for TTS is typically recorded and tested in quiet office environments.
 
 \emph{To study human adaptation while retaining high-quality, clean audio, we utilized a readily available video-conferencing platform equipped with ambient sounds to aid actors in adapting their voice to fit the target environment while recording from participant's headphones. We then explored how the participants adapted their voices and designed a set of ambient-specific voices to mirror these features, finally conducting a comprehensive user perception study of the voices on a robot in varying virtual environments.}

\vspace{10pt}
\noindent \fbox{\begin{minipage}{\textwidth}
The third problem is addressed in our article: \emph{Read the Room: Adapting a Robot's Voice to Ambient and Social Contexts, published in the Companion Proceedings to the IEEE/RSJ International Conference on Intelligent Robots and Systems 2023} \cite{readroom}, as well as a workshop paper: 
\emph{Read the Room: Adapting a Robot’s Voice to Ambient and Social Contexts, published at Sound for Robotics at the IEEE International Conference on Robotics and Automation 2022}, and a demo: \emph{I’m a Robot Hear Me Speak in the Companion Proceedings of the ACM/IEEE International Conference on Human-Robot Interaction 2023}\cite{hear_speak}.
\end{minipage}}

\section{Problem 4 - Second Language TTS}

Finally, given the understanding that humans modify their voices to improve interactions, we ask how we can modify TTS to suit the perceptual needs of second-language listeners. The use of voice-based technology, both in perception, production, and end-to-end systems, is seeing expansive growth. These systems can help improve accessibility, such as in-home automation systems for the physically and visually impaired \cite{VIEIRA2022121961, deoliviera}, or as a socially supportive device to allow elderly individuals to maintain their autonomy \cite{london}. However, there remains much work to ensure the ethical validity of such systems in terms of transparency, bias, fairness, and many other factors \cite{ethicsreview}. One aspect of fairness lacking in systems with synthesized speech is their inability to adapt to those with different hearing and comprehension abilities, such as L2 listeners. In human-to-human interaction, the ability of a speaker to adapt to an interlocutor is invaluable. Humans will modify their speech to interlocutors with reduced comprehension abilities, e.g., babies \cite{McClay22} or L2 speakers \cite{redmon2020cross}. Moreover, taking cues from multiple modalities, a speaker is able to perceive when they are not being understood and make adjustments to their speech production to increase clarity both in terms of linguistic \cite{Biro22, Maniwa09} and para-linguistic (e.g., emotional \cite{Banse96, Bachorowski99}) contexts. There is, however, evidence that human adaptation to L2 listeners does not, in fact, aid the L2 listeners in their comprehension \cite{AOKI2024101328,ROTHERMICH201922}. Therefore, to improve the fairness of voice technology and move towards adaptive synthesized speech, we must first understand how speech can be adjusted to facilitate comprehension in a data-driven, perception-based manner.

In English, L2 speakers often struggle to differentiate tense (long) and lax (short) vowels. For example, French and Japanese first language (L1) speakers will often replace the lax \textipa{/I/} (ship) with the tense \textipa{/i/} (sheep) \cite{2019japanese, iverson12}, while Chinese speakers replace both vowels with the Chinese vowel \textipa{/I/}, which requires a higher and more frontal position of the tongue. Classically, vowel perception is considered as categorizing speech sounds by comparison with a pre-learned auditory representation, presumably a spectral one in formant space \cite{LIB00}. However, it is widely documented that perception also operates relative to its acoustic context \cite{Stilp20, MCMU11}. For instance, following a phrase spoken quickly, a sound can be perceived to be longer \cite{Newman96}. Yet, how such context effects can interact or compete, and the exact timescale at which they occur, remains poorly understood \cite{Reinisch13, Newman96, Johnson99, Stilp20}.

\emph{To improve the comprehension of TTS from an L2 perspective, we use the data-driven approach of psychophysical reverse correlation \cite{Ponsot18} to reconstruct the prosodic profile of these difficult vowel sounds over a sentence for both L1 and L2 speakers. Accordingly, we ask: How can the perception of vowels that are known to be difficult for L2 speakers be controlled and comprehension be improved through modifications of pitch and duration across a word or phrase? We then apply these learned profiles to be automatically applied in a TTS system to improve comprehension for L2 Listeners.}

\vspace{10pt}
\noindent \fbox{\begin{minipage}{\textwidth}
This problem has been addressed in our article: \emph{Mmm whatcha say? Uncovering distal and proximal context effects in first and second-language word perception using psychophysical reverse correlation at Interspeech 2024} \cite{tuttosi24_interspeech}, an article under review: \emph{You Sound a Little Tense: L2 Tailored TTS Using Durational Vowel Properties for Interspeech 2025} \cite{l2tts}, and an article in preparation for submission to PNAS.
\end{minipage}}

\vspace{10pt}

Taken together, this thesis' contribution is a lightweight TTS with temporally-variant expressivity, social and ambiance adaptation, and L2 clarity mode, providing the building blocks for a synthesized robot voice that can improve human-robot interactions with L2 English speakers.

\section{Thesis Organization}

In the rest of this manuscript, Chapter 2 presents a review of existing literature in the domains of 1) lightweight and expressive TTS, 2) ambient adaptive robot voices, and 3) understanding linguistic features of English vowels for first and second-language speakers. This provides a theoretical background for our work, both providing evidence for the methods used and displaying gaps in the existing literature. Chapter 3 then presents our EmojiVoice toolbox along with our expressive voices and case studies of their deployment (Problems 1 \& 2 above). Chapter 4 explores how to adapt a robot's voice to physical and social ambient contexts (Problem 3 above). Chapter 5 presents the linguistic studies that provide the evidence and background for our L2 TTS and the design and testing of the L2 TTS system (Problem 4 above). Finally, in Chapter 6, we present conclusions, future directions, and suggestions on how these improvements can and should be integrated in the future.

\chapter{Related Work}

\section{Text-to Speech Synthesis (TTS) for Robotics: Performance and Long Term Expressivity}

Robotic platforms for HRI are frequently limited by hardware and connectivity, which severely constrain the deployment of state-of-art TTS systems on robots. In addition, the HRI context creates a stringent need for voice expressivity and adaptiveness, for which speech models developed in other usage scenarii (e.g. vocal assistants) are often not designed. This creates an opportunity for research and innovation, which only a handful of recent work has started to address.

\subsection{TTS limitations for robotics applications} \label{tts-limit}

State-of-the-art TTS models often contain hundreds of millions of parameters (coqui-XTTS: 750M \cite{coqui}, VoiceCraft: 830M \cite{peng2024voicecraft}, Parler-TTS: 878M \cite{lyth2024naturallanguageguidancehighfidelity}). For models of this size, a non-negligible part of the robot's storage space and RAM are required for their deployment. Moreover, due to the large computational requirements, these models frequently do not run in real-time, often having a real-time factor (RTF) greater than 1, meaning it takes at least as long as the duration of the audio to generate the speech, disrupting the flow of a conversation. Yet, large models' size and computational requirements are not the only barriers for robotics deployment of state-of-the-art TTS models. Many large models, especially those able to perform zero-shot voice cloning, such as CoquiXTTS \cite{coqui} and VoiceCraft \cite{peng2024voicecraft}, are auto-regressive. These models are extremely human-like, VoiceCraft having a naturalness mean opinion score (MOS) nearly akin to the original human voice (VoiceCraft: 4.16±0.08, original human speech: 4.29±0.08 on Spotify speech \cite{peng2024voicecraft}), yet are not readily controllable in a fine-grained manner. For example, it is not possible to control the duration or pitch on a phoneme level as one can with Fastpitch \cite{fastpitch, ju22_interspeech} and Fastspeech-based models \cite{ren2022fastspeech2fasthighquality, diatlova23_ssw, strickland23_ssw}. Yet these controllable models are considered to have low naturalness \cite{mehta2024matcha}.
Additionally, auto-regressive models often hallucinate, resulting in pure noise or gibberish \cite{hallucinate}. Some models, such as Parler-TTS \cite{lyth2024naturallanguageguidancehighfidelity}, have reduced the number of hallucinations but still return nonsensical phrases when met with an unknown word or symbol. 

Lastly, many of the high-performance TTS systems are either not open source or lack documentation on modifying or fine-tuning the model. Even those that do, once again noting Parler-TTS and its excellent documentation, require large amounts of data to fine-tune to a specific use case. For engaging and robust human-robot interactions, we need a voice to be lightweight, generate in real time, and be free of hallucinations. 

\subsection{Long-term variable expressivity}

Many expressive TTS systems are controlled through a series of complex models and hours of expressive training data \cite{cho24_interspeech, bott24_interspeech}, which is contrary to the need for minimalist systems and is not flexible to fast fine-tuning to a specific use case. Moreover, so-called expressive TTS systems mostly focus on increasing pitch and pitch range to create a more ``joyful" sounding voice \cite{cho24_interspeech, bott24_interspeech}, as previous research has found that these modifications drive expressivity in human speech \cite{anotherstory, gustafson01_eurospeech}. However, there is a lack of work exploring how this type of expressive voice performs over long periods of speech; for example, does listening to a highly expressive TTS for a long time become monotonous? Recent advancements in long-term TTS have improved smoothing over sentences by including contextual information from previous phrases \cite{xiao23_interspeech}, but do not use this to control the expressivity. A recent dataset has been released to help train Chinese expressive storytelling TTS through multiple annotations, including emotional colouring, with the goal of creating a ``storytelling voice" \cite{storytts}. They, however, do not provide an open-source model nor information on the model size, stability, and synthesis time of their baseline. Further, their use of an encoder to control emotion expressivity does not allow the user explicit expressive control, as may be needed for HRI studies. In Chapter 3, we will address these gaps though a lightweight, customizable TTS with temporally variational expressive control.

\section{Ambient Intelligence for Robotics: How Humans and Machines Adapt Their Voice to Context}

Humans' ability to assess and adapt to their physical and social environment is innate and a primary factor in smooth and pleasant interactions. Yet, in human-robot-interactions, we have yet to see extensive work on the adaptation of a robot, particularly their voice, to a given context. This gap provides an opportunity to expand the research on robotic vocal adaptations, improving the perception of a robot's awareness and appropriateness, and the interaction overall.

\subsection{Context adaptation in human interaction}

In human-to-human interaction, it is well documented that interlocutors will both adapt their speech to the context. One important goal of vocal modifications is creating ``deliberately clear speech'' \cite{confluent_talker_and_listener}. These modifications may notably occur in difficult listening environments \cite{vocal_effort_acoustic_environments, acoustic_phonetic_adverse_listening_conditions}, but may also occur to communicate a specific social signal \cite{polite}. Modifications are also often listener-specific, as is the case in infant, child \cite{whats_new_pussycat, piazza2017mothers, de2003investigating, hirsh1982doggerel}, hearing impaired \cite{mommy_speak_clearly}, and machine-directed speech  \cite{prosodic_and_intelligibility}. Vocal modifications are typically produced without conscious effort to elicit a specific auditory feature; rather, they are produced as a result of achieving the aforementioned goals. Moreover, a speaker is able to perceive when they are not being understood and make adjustments to their speech production \cite{Biro22, Maniwa09,Banse96, Bachorowski99}.

\subsection{Contextual Voices in robotics}

Although TTS has become an inexpensive and efficient means to create realistic voices for a variety of robotic behaviors \cite{TTS_relevancy_of_voice_effect, cost_effective_cvocoder, TTS_and_robots}, it is not entirely versatile. For many years, the primary concern for TTS was intelligibility. Because of this, state-of-the-art TTS voices could be mistaken for a human voice, but they still lack the ability to adapt to physical and social contexts. Furthermore, much controllability in stable TTS (i.e., those that do not hallucinate) is constrained by Speech Synthesis Markup Language (SSML). Although the available features have broadened and include loudness, pitch, and rate-of-speech\footnote{https://cloud.google.com/text-to-speech/docs/ssml}, it is not clear whether these features are sufficient for a robot to flexibly and automatically adapt its voice to context.

Several studies have shown that humans show preferences for a robot's voice depending on context and task \cite{context_and_robot_voice, perceptual_effect_ros, providing_route_directions, adaptive_speech_cog_robots}. In \cite{context_and_robot_voice}, participants rated the appropriateness of different robot voices in various contexts, including schools, restaurants, homes, and hospitals. They found that, even given the same physical appearance, participants selected varying voices depending on context and concluded that a robot voice created for a specific context is likely not generalizable. Further studies suggest the incorporation of context-based methods such as sociophonetic-inspired design \cite{voice_as_a_design_material}, acoustic-prosodic adaptation to match user pitch \cite{voice_adaptation_robot_learning_companion}, or incremental adaptation of loudness to the user's distance \cite{HRI_incremental_speech_adaptation}. 

While human modifications are well studied, their application to robot voices has been limited. One study has explored the adaptation of a robot’s voice to its acoustic environment by adjusting volume based on environmental noise levels \cite{volume_adaptation_telepresence_system}. Another used the analysis of the ambient annoyance and user information to modify the robot's voice \cite{renclarity}. Overall, the literature has focused on the adaptation of loudness to ambient environments. 

In Chapter 4, we expand on the literature by applying knowledge from studies on human voices, where we understand that contextual appropriateness relies on more than loudness features to address the observed need for more adaptive robot voices. 

\section{Second-Language Speech Perception: Evidence from Linguistics}

One critical aspect to improve the fairness of voice technology is to make it adaptive to a diverse set of perceptual needs, perhaps foremost facilitating comprehension for second-language (L2) listeners. While much of linguistic and cognitive science research has explored L2 language perception in human-to-human interaction, currently, there is no existing TTS adaptation strategy for L2 listeners. We review here this literature, and identify a data-driven methodology, reverse correlation, which appears well-suited to develop such a strategy.

\subsection{Vowel perception in L1 and L2: local and context effects}

Classically, vowel perception is viewed as a local (i.e., time-resolute) sensory process that categorizes individual speech sounds by comparing them to a pre-learned auditory representation, presumably spectral in formant space \cite{kuhl97}. Formants are the prominent frequencies (F0 being pitch) in a sound that determines a vowel's phonetic quality and are a primary cue for L1 speakers \cite{kuhl97}. Speech sounds, however, are highly variable within and across speakers, and it is now widely documented that word perception also operates relative to its acoustic context \cite{Stilp20}. For instance, following a context spoken quickly, a sound can be perceived to be longer than it actually is \cite{Newman96}. Similarly, if a speaker's first formant (F1) range is low, a subsequent sound that is ambiguous between \textipa{/I/} and \textipa{/E/} may be perceived as having a relatively high F1 and thus subjectively sounds more like \textipa{/E/} \cite{Watkins94, Reinisch13}. Such context effects appear to be active in both a forward and backward manner - e.g., they apply whether the word begins or ends the sentence or even if the word is mid-sentence \cite{miller79}. While it appears difficult to help L2 listeners overcome less spectrally-resolute representations by manipulating the spectral content of the target sound itself (although see \cite{chen2012effect} for a similar idea for hearing-impaired listeners), the existence of such context effects opens the opportunity to use pitch and duration manipulations within and around the target sound strategically to \emph{bias} the perception of difficult L2 sounds into their correct interpretation.

There is debate, however, about the exact temporal characteristics of such context effects in ecological sentences (i.e., how far the effects are from the contextual changes) and whether the location of the word in the phrase affects which speech cue is involved (e.g., pitch v.s. duration) \cite{Stilp20, Reinisch13}. Previous research has, for instance, suggested that the influence of the preceding speech rate on phoneme distinction may be limited to a temporal window of one or two adjacent phonemes \cite{Newman96}. Yet other work suggests that spectral context effects result from a form of speakers' vocal tract length normalization (i.e., a listener will inherently attempt to judge the length of a speaker's vocal tract to more clearly interpret their specific pronunciation of formants), which benefits from exposure to long-term spectral cues accumulated over possibly several sentences, or even non-auditorily, from a silent video recording of the speaker \cite{Johnson99}. 

In addition, it is poorly understood whether non-native language processing is able to recruit similar acoustic context mechanisms to those found in native language. For instance, Kang, Johnson, and Finley \cite{KANG16} found that French-L2 speakers failed to use vowel context (i.e., to compensate for coarticulation) when judging fricative sounds when such vowels were unfamiliar. Still, other groups of participants (e.g., English and Tamil, or English and Japanese) behaved identically on different types of sound contrasts \cite{MANN86,VIS13}.

\subsection{Hypothesis- vs data-driven paradigms in speech perception}

Determining the acoustic and temporal characteristics of information intake in sentence perception has always been methodologically challenging for speech perception research. Studies such as \cite{Heff13} have used hypothesis-driven experimental paradigms to establish the causal influence of specific cues on word contrasts by systematically varying their intensity. Yet, these methods are applicable only in a limited scope as assessing the relative perceptual weight of several temporal regions of interest quickly becomes impractical \cite{Newman96}.

To document such temporal dynamics, several studies have relied on concurrent eye-tracking in visual search tasks (e.g., printed words or objects corresponding to each word interpretation) and compared the time course of eye fixations to the occurrence of contextual cues \cite{SHATZ06,Reinisch13}. However, this type of study remains relatively uninformative when a specific listener or listener group (e.g., L1 vs L2 listeners \cite{KANG16}) fails to show the effect, as the researcher is left wondering what other possible mechanism may be at play. What is needed is a methodological way to automatically extract, in a data-driven manner, a listener’s mental representation of what acoustic profile (i.e., what cue and where) drives a specific vowel contrast one way or another.

\begin{figure}[H]Update index.html
  \centering
  \includegraphics[width=\linewidth]{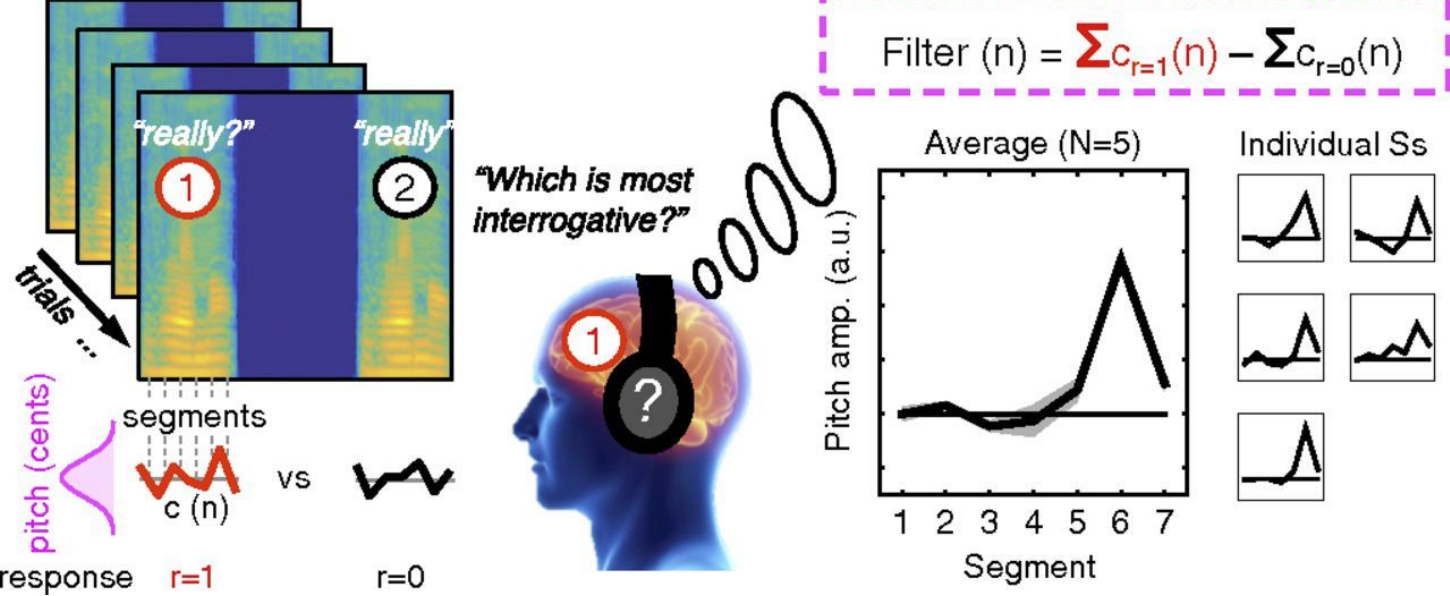}
  \caption{Visualization of the reverse correlation process from \cite{Ponsot18}. (Left) Stimulus with random pitch contours, the participants made a judgment on a pair of sounds (center) resulting in a prosodic mental representations of the judgment.}
  \label{rev-cor-examp}
\end{figure}

Psychophysical reverse correlation \cite{AHU71} is an experimental paradigm used to reconstruct the signals and features that form the mental representation of the decision-making taken by humans when responding to a set of stimuli. The procedure involves a participant observing hundreds of stimuli with random modifications to a target dimension, e.g., facial action units for images \cite{Yan2023}, or social impressions of a speaker’s dominance \cite{Ponsot18} or confidence \cite{Goupil21}. From their responses, one can reconstruct the prosodic profile that maximizes the likelihood of responding to one option or the other (for a review, see \cite{Mur11}). A visualization of the reverse correlation process can be seen in Fig. \ref{rev-cor-examp}.

In Chapter 5, we introduce the use of reverse correlation to investigate whether and how duration and pitch cues, both within a vowel and within the context surrounding a vowel, can facilitate the perception of difficult English tense(long)-lax(short) vowel pairs (e.g., peel vs pill) for L1 and L2 listeners. Finally, we use this novel information to create a ``L2 clarity'' mode within a state-of-art TTS, and test whether it improves L2 comprehension.

\chapter{EmojiVoice Toolbox}

\begin{figure}[h]
  \centering
  \includegraphics[width=0.85\linewidth]{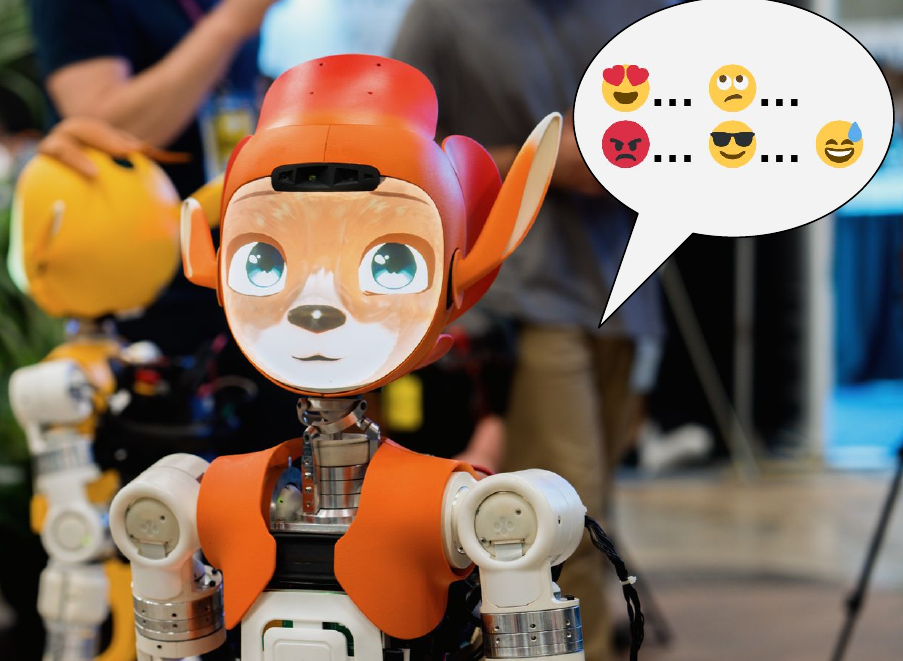}
  \caption{Miroka robot speaking with EmojiVoice expressive TTS using emojis to colour the text over phrases.}
  \label{header-emoji}
\end{figure}

\section{Early Exploration of Expressive Teaching Voices}
\begin{figure}[]
      \centering
      \includegraphics[width=0.78\linewidth]{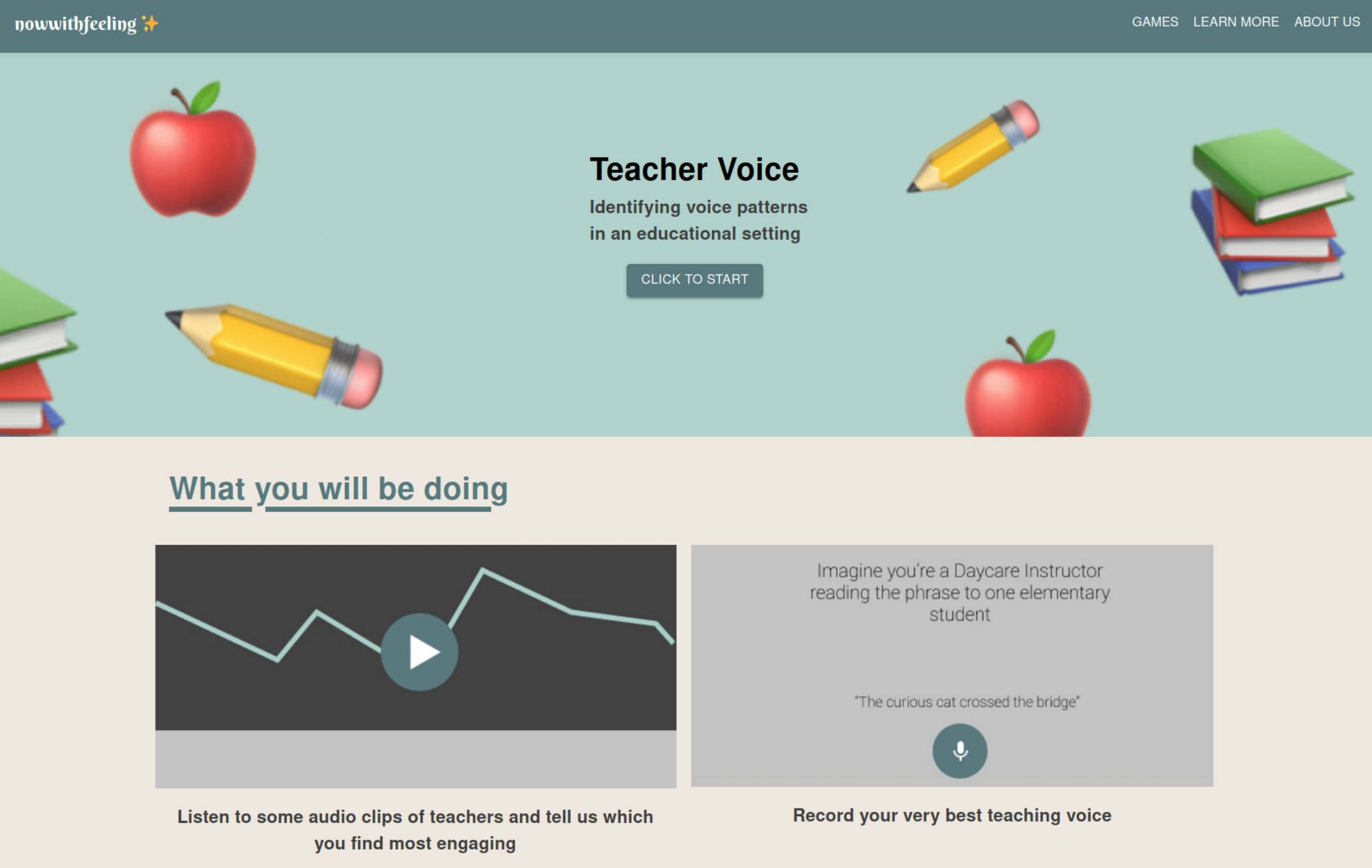}
      \caption{Example image from the Now With Feeling webpage where users around the world can submit their best teaching voice.}
      \label{nowwfeeling}
\end{figure}

We created a citizen science website called \emph{Now With Feeling}: \url{https://nowwithfeeling.com/teacherVoice} to get an initial idea of how expressive teaching voices should sound. Here, participants listened to clips of different teaching voices taken from YouTube lectures. They had to rate the voices from most to least engaging. Then, the participants read different text prompts directed at various audiences. For example, one would read a passage from a storybook to a group of 5-year-olds or read a short lecture passage on mathematics to a university class of 100 students. The participants were prompted with an image of the situation and could practice and play back their speech before submitting the final recording. The site extracted voice features and compared them to voice features of well-known teachers, such as Bill Nye, letting the participants know which teacher they sounded most like. An image of the website can be seen in Fig. \ref{nowwfeeling}. The website was shared to target teachers, such as on a Reddit page for elementary school teachers.

It was through these audio files that we began to understand the difference between an expressive and temporally expressive voice, especially in a storytelling scenario. We found that the voices that were the most engaging did not read entire pages with a pleasant and excited voice; instead, from phrase to phrase, they changed the type of expression in their voice, keeping us engaged long-term.

To address this and the technical limitations of large TTS models for robot deployment, we built the EmojiVoice toolbox \cite{emojivoice}. Our toolbox uses an implementation of Matcha-TTS, a fully open-source, lightweight, and fast TTS model that requires as little as three minutes of data to fine-tune a new voice and has only 20.9M parameters. EmojiVoice runs in real-time on a robot's hardware and can be easily modified to HRI use cases. 

\section{TTS model}
\begin{figure}[t]
  \centering
  \includegraphics[width=\linewidth]{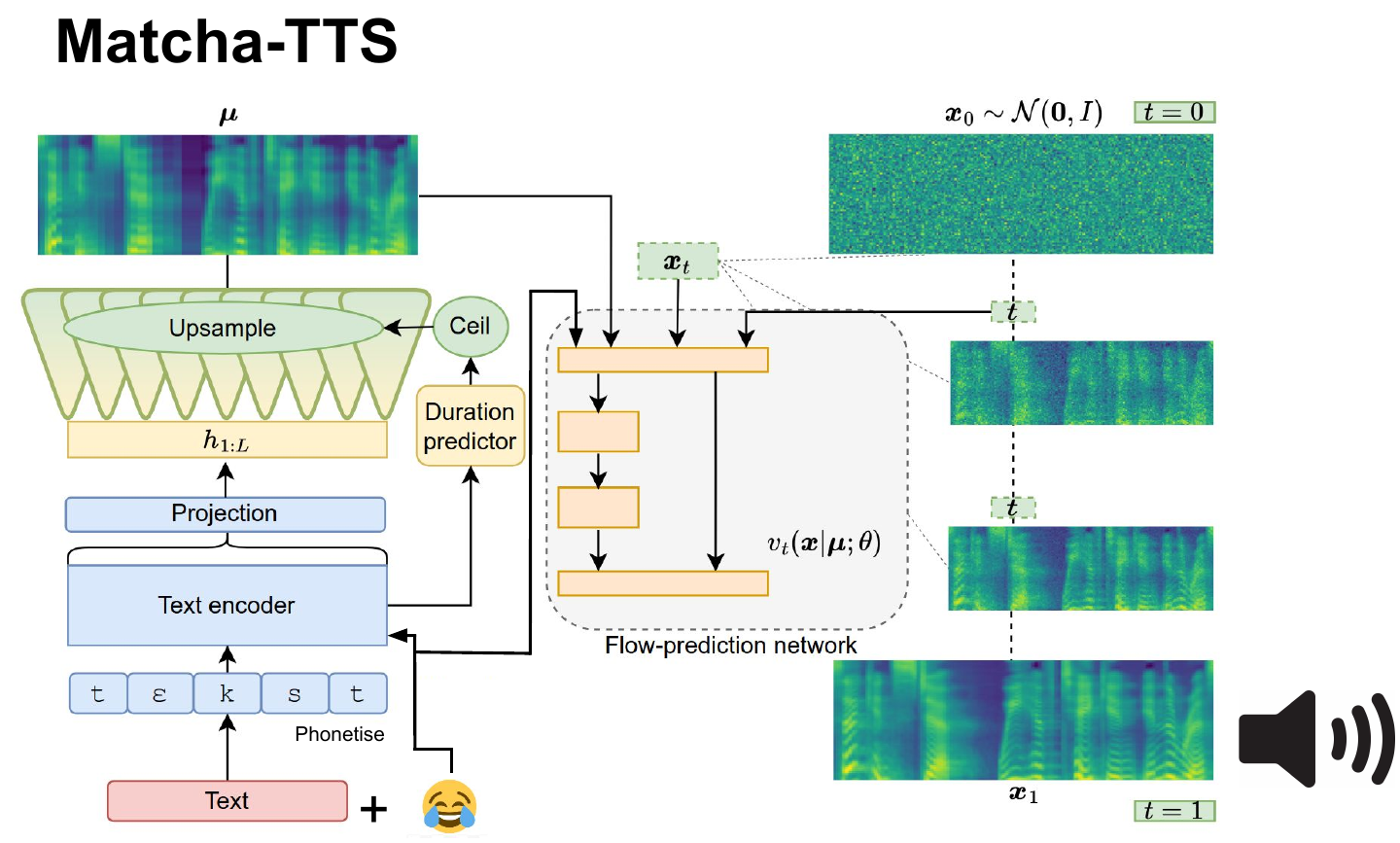}
  \caption{Multi-speaker Matcha-TTS with EmojiVoice architecture.}
  \label{arch}
\end{figure}

We presented, for the first time, a multi-speaker version of Matcha-TTS \cite{mehta2024matcha}\footnote{https://github.com/shivammehta25/Matcha-TTS} that has been fine-tuned to create EmojiVoice. Pre-fine tuning the multi-speaker TTS was trained on the VCTK corpus \cite{VCTK}, with 110 speakers. Behind the scenes, Matcha-TTS is a neural TTS trained using optimal-transport conditional flow matching (OT-CFM). The model is probabilistic rather than auto-regressive, allowing us to circumvent the aformentioned weakness of auto-regressive models (see Section \ref{tts-limit}). Moreover, the model is ideal for robot applications as it has a compact memory footprint of only 20.9M parameters, in comparison to 830M for VoiceCraft and 41.2M for Fastspeech 2, and has an RTF of 0.3 (an RTF $<1$ is considered real-time \cite{pratap20b_interspeech}) on a Nvidia Jetson AGX Orin 64GB, a GPU with similar power and memory to those applied in production robotics. EmojiVoice takes a label, in this case, an emoji number, along with the input text. This is concatenated both to the text encoder input and the flow-prediction network (decoder) input to contain multiple voices in a single, small (78MB, in our case) checkpoint. The architecture can be seen in Fig. \ref{arch}. Having multiple voices in a single checkpoint reduces space and allows for efficient switching between voice styles without reloading the model. A comparison of Matcha-TTS and other TTS systems can be seen in Table \ref{compare}. 

\begin{table*}[t]
\centering 
\caption{Comparison of open source TTS systems and robot-centric features.}
\resizebox{\textwidth}{!}{
\begin{tabular}{|l|l|l|l|l|l|l|}
\hline
  & VoiceCraft & CoquiXTTS & Parler-TTS & FastPitch & Matcha-TTS & EmojiVoice \\
 \hline
$<$ 100k parameters & \xmark & \xmark & \xmark & \cmark & \cmark & \cmark \\
No Hallucinations & \xmark & \xmark & \xmark & \cmark & \cmark & \cmark\\
Real Time Synthesis  & \xmark & \xmark & \cmark & \cmark & \cmark & \cmark\\
Quick Fine Tuning*  & \xmark & \xmark & \xmark & \xmark & \cmark  & \cmark\\
Consistent Speaker Control & \xmark & \xmark & \xmark & \cmark & \cmark & \cmark\\
Controllably expressive  & \xmark & \xmark & \cmark & \cmark** & \xmark & \cmark\\
\hline
\end{tabular}}
\label{compare}
\parbox{6in}{
\footnotesize{*on under 1 hour of fine-tuning data, **control pitch and duration by hand rather than intention or category}
}
\end{table*}

\section{Emoji Representation}
\begin{figure}[]
  \centering
  \includegraphics[width=\linewidth]{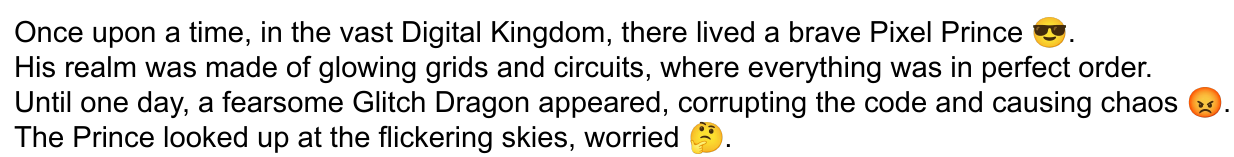}
  \caption{Example of text with emojis. Sample of the script provided to the TTS for Case Study 2: Storytelling task.}
  \label{case1script}
\end{figure}

\begin{figure}[]
  \centering
  \includegraphics[width=0.7\linewidth]{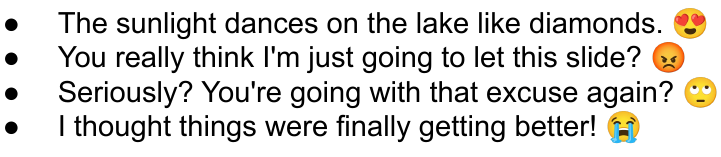}
  \caption{Example prompts for voice actors for data collection and training.}
  \label{prompts}
\end{figure}

We suggest emojis as a means to prompt expressive styles that can be used for temporal variational control, i.e., selecting a different emoji style for each phrase generated. In 2019, \cite{bai2019} stated that were were 3,019 emojis in Unicode, and this number has now grown to 3,782 in only 5 years. While not all emojis can be considered representations of human expression, considering 106 emojis are smileys alone, with the addition of non-smiley emojis such as ``thumbs up" and ``flexed biceps,'' we have a rich representation of both emotions and social signals larger than, or at least as large as the most commonly used discrete emotion models \cite{schroder2006first, ShaverPhillip1987EKFE, PLUTCHIK19803}.  The variation provided by the standard Unicode emoji set is especially useful in that they can express emotions: \includegraphics[scale=0.3]{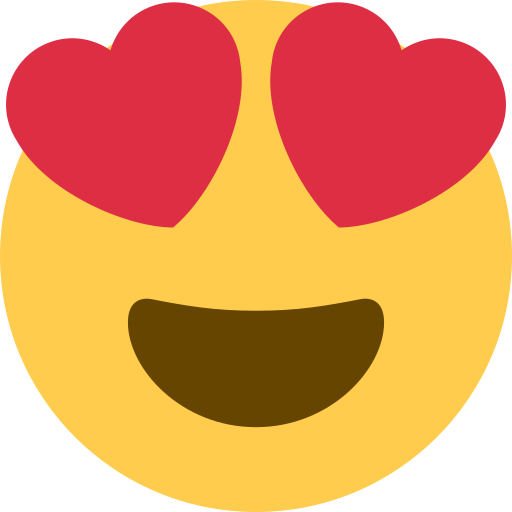}, \includegraphics[scale=0.018]{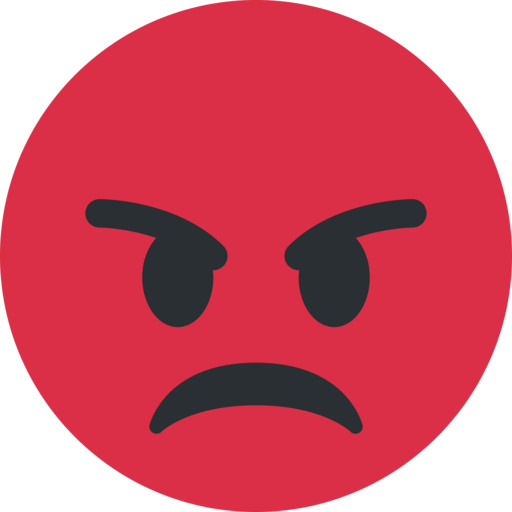}, attitudes: \includegraphics[scale=0.04]{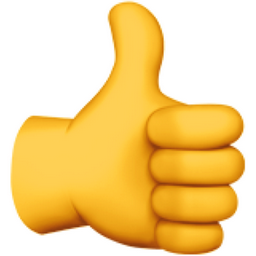}, \includegraphics[scale=0.018]{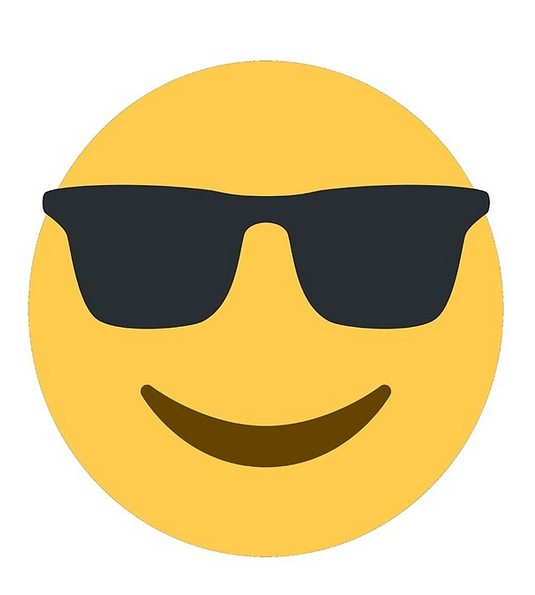}, mental states: \includegraphics[scale=0.018]{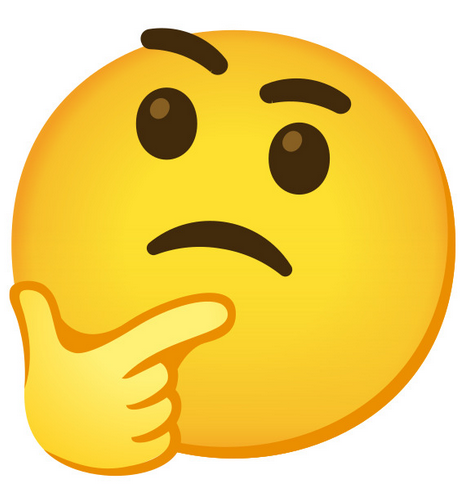}, \includegraphics[scale=0.018]{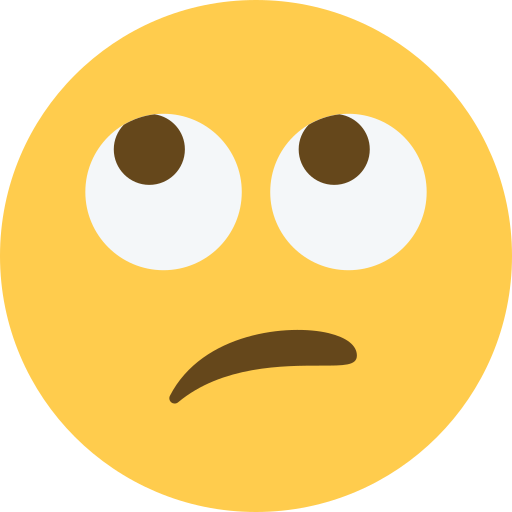}, and bodily states: \includegraphics[scale=0.1]{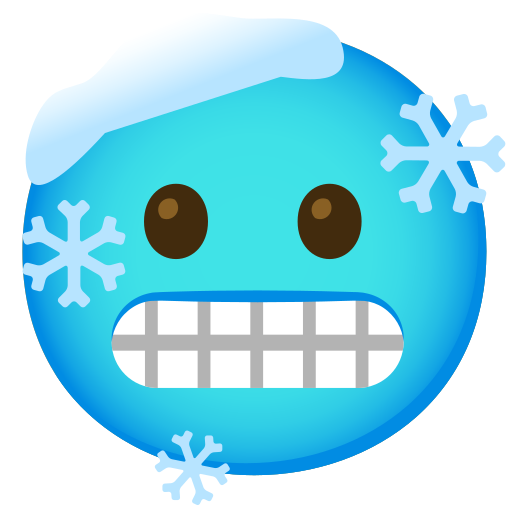}, \includegraphics[scale=0.005]{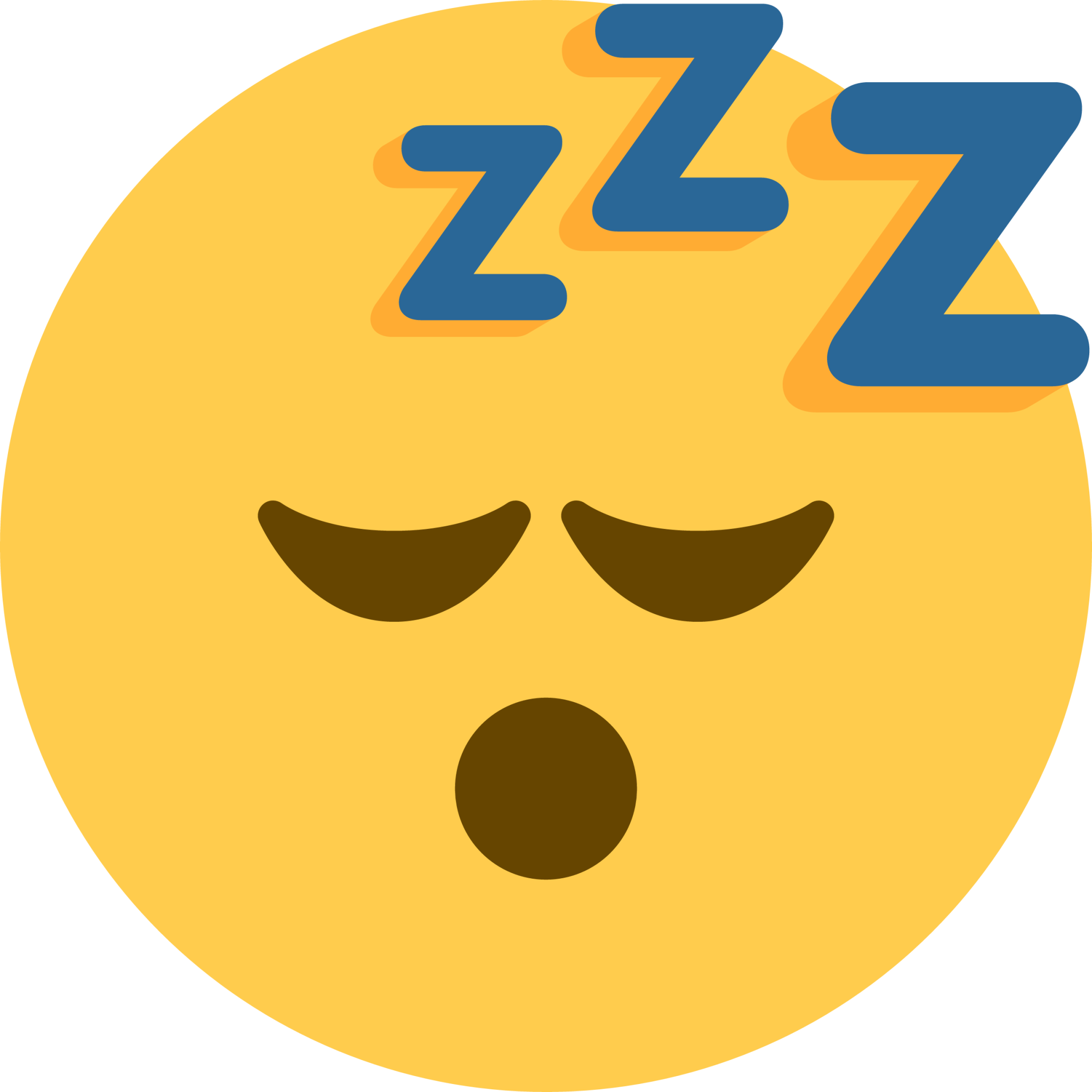}. Moreover, we have become accustomed to the use of emojis in our everyday written expression; for example, almost half of all Instagram posts with text contained an emoji in 2015 \cite{dimson2015emojineering}. This has resulted in a large amount of the training data for large language models (LLMs) containing emojis along with the text; hence, LLMs can readily produce emojis within text in a rather convincing manner \cite{savage2024real, dunngood}. Although LLMs are large and costly, they are increasingly incorporated into robotic conversational systems and are a promising option for generating dialogue~\cite{10611232, 10599883, ZHANG2023100131, 10.1145/3568294.3580040}. This means that emoji-based voice selection can be easily integrated into systems using an LLM with a negligible added cost, where the emotional intention is interpretable by humans. Thus, we used emojis to 1) collect emotional vocal expressions in humans to generate training data for our expressive TTS (Fig. \ref{prompts}) and 2) control the TTS voice selection on a phrase-by-phrase basis to ensure long-term temporal variability as seen in Fig. \ref{case1script}.

\begin{figure}[]
  \centering
  \includegraphics[width=0.8\linewidth]{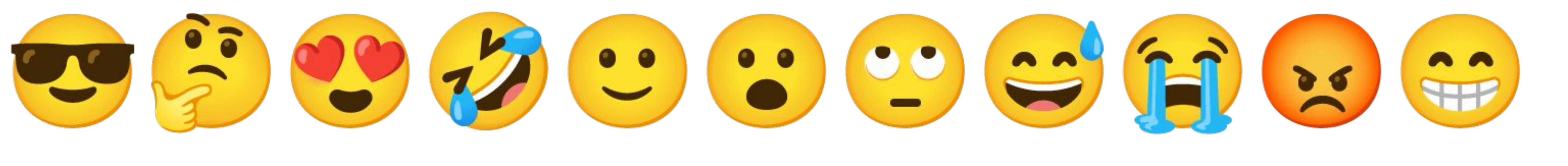}
  \caption{Emojis used for voice prompting and voice selection.}
  \label{emojis}
\end{figure}

 To select the emojis, we aimed to have more positive than negative emojis, unlike the basic emotions, which are primarily negative. We selected an initial list of 23 emojis grouped into negative, neutral, and positive. We further filtered the list with a goal of $<$15 emojis to maintain a reasonable workload for the speaker while maintaining a range of expressions. Through initial exploration, we found that maintaining a consistent production of an emoji was essential for reducing the amount of training data and adequately reproducing the tone of the speaker's expression, i.e., higher variability in the data requires more data to capture these nuances. As such, we kept the emojis where the speaker had a precise mapping of an emoji to a vocal expression and, in turn, was able to produce the most consistent voice. We settled on 11 emojis (Fig. \ref{emojis}), a repertoire that outnumbers typical emotional voice datasets (e.g., 9 for IEMOCAP \cite{BussoCarlos2008Iied}, 8 for RAVDESS \cite{LivingstoneStevenR2018TRAD}, 8 for MSP-PODCAST \cite{Lotfian_2019_3}). We do not claim that the present set of emojis is the optimal set; rather, it is a starting point for researchers to explore, expand, or even limit the expressivity of robots using this representation, and researchers can use as little as 3 minutes of speech per emoji to customize the TTS. Moreover, emojis could be substituted with an emotion label or text prompts, and we invite researchers to use our toolbox to do so if it suits their needs. 

\subsection{Data collection and training}
We provide information on data collection for other researchers to extend or retrain EmojiVoice depending on their needs. To collect training data, we created a prompting script for the speaker. We generated 50 sentences for each emoji using ChatGPT\footnote{https://openai.com/} with the prompt: \emph{Please provide 50 short phrases that reflect this emoji: X. The phrases should use all the English phonemes and should not be repetitive, using a variety of words.} These phrases appeared on the screen one at a time with the target emoji appended to the end of the sentence, and the speaker pressed a button each time they were ready to record. Example phrases can be seen in Fig. \ref{prompts}. The data was split into 40 sentences of training data and 10 for validation. We fine-tuned off of the multi-speaker VCTK checkpoint available with Matcha-TTS\footnote{\url{https://drive.google.com/drive/folders/17C\_gYgEHOxI5ZypcfE\_k1piKCtyR0isJ}}. When fine-tuning 11 of the 110 speakers were overwritten, choosing voices that match the target gender. Other speaker numbers remain available after fine-tuning, however, they tend to be an unstable mix of several speaking voices. We trained on a NVIDIA GeForce RTX 4070 GPU with an Adam optimizer and a learning rate of 1e-4, unchanged from the checkpoint, with a batch size of 20. The model was trained for 85 epochs, approximately 20 minutes. Using our toolbox, we trained three checkpoints with emoji voice data from three actors, two female (Paige and Olivia) and one male (Zach). The toolbox extends Matcha-TTS by including 1) Training files setup: examples, raw data, and 3 checkpoints (with and without optimizers), 2) Additional information on the amount of data needed to fine-tune, 3) Scripts to record data, 4) Wrappers to parse emojis in text to prompt the voices, and 5) a conversational agent. Our toolbox and voices are available free and open source\footnote{https://github.com/rosielab/emojivoice}.

\section{Automatic Speech-to-Speech System}\label{s2s-model}
\begin{figure}[]
  \centering
  \includegraphics[width=300pt]{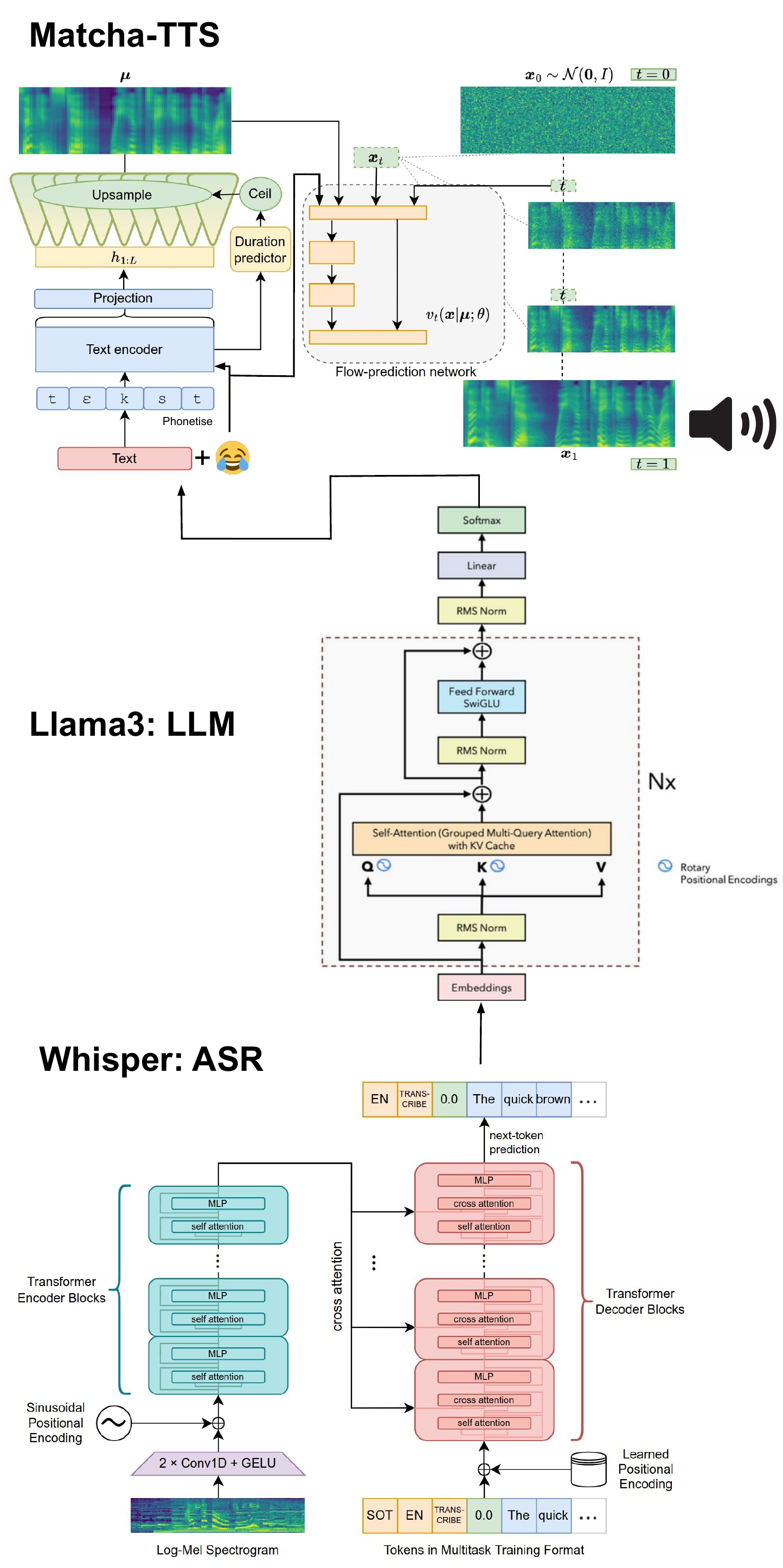}
  \caption{Architecture of the pseudo speech-to-speech interactive agent combining: Whisper for ASR (bottom), Llama 3.2 as the LLM (middle), and EmojiVoice as the TTS (top).}
  \label{arch2}
\end{figure}

In addition to EmojiVoice and the code required to fine-tune and synthesize with it, we include an autonomous, pseudo-speech-to-speech interactive agent chaining automatic speech recognition (ASR), a large language model (LLM) and our TTS with emoji selection to create an autonomous dialogue system. We say ``pseudo'' as each model runs end to end without linking the embedding spaces between models. The interactive agent runs entirely locally and does not require a connection to the internet. There may remain some latency issues in the LLM on low-power GPUs, but it ran in real-time for our testing on an RTX 4060. We use OpenAI's Whisper \emph{tiny.en} model \cite{whisper2023} for speech recognition (39M parameters). To mitigate the complexity of deciding when the user has finished speaking \cite{speakingturn}, we use a push-to-talk system, recording the voice and producing a transcription. The transcription is the input to the LLM, currently Meta's Llama3.2 \cite{dubey2024llama3herdmodels} (1.23B parameters) implemented via Ollama\footnote{https://ollama.com/} and LangChain\footnote{https://www.langchain.com/} as a chatbot. Although we have used the smallest Llama3 model available, this remains a bottleneck in terms of memory footprint. In the toolbox, the voice designer is able to control the 1) prompt to the LLM to a specific use case and 2) emoji set. This prompt controls the concatenation of the emoji to each response. We extract the appended emoji and use a mapping from the emoji to the associated emoji style numbers, providing this to the TTS to synthesize the appropriate voice. The architecture can be seen in Fig. \ref{arch2}

\section{EmojiVoice Case Studies}

We explore three possible use cases for EmojiVoice to understand how an expressive voice can be effectively deployed in HRI. The first case is a scripted conversation with a robot; the second is a short story told by the robot; and the third is an autonomous interactive agent playing a storytelling game with the user. The first two case studies were conducted on the Pepper and the Miroka (Enchanted Tools) robots, and the third was with an image of a Miroka robot. Each case study compared three voices (using the ``Paige'' voice): 1. \textbf{Baseline}: the original voice from the Matcha-TTS VCTK checkpoint for speaker one for all phrases; 2. \textbf{Pleasant}: the EmojiVoice voice that represents the \emph{slightly smiling} emoji, which has an increased pitch range over the baseline as is the focus with other expressive TTS systems, for all phrases; 3. \textbf{Emoji}: the complete set of 11 emoji voice styles with each phrase containing an emoji to select a specific voice. Both the Baseline voice and training data for the EmojiVoice were of speakers perceived to be 25-35 years old and female.

\begin{figure}[]
    \centering
    \begin{minipage}{0.49\linewidth}
  \includegraphics[width=\linewidth]{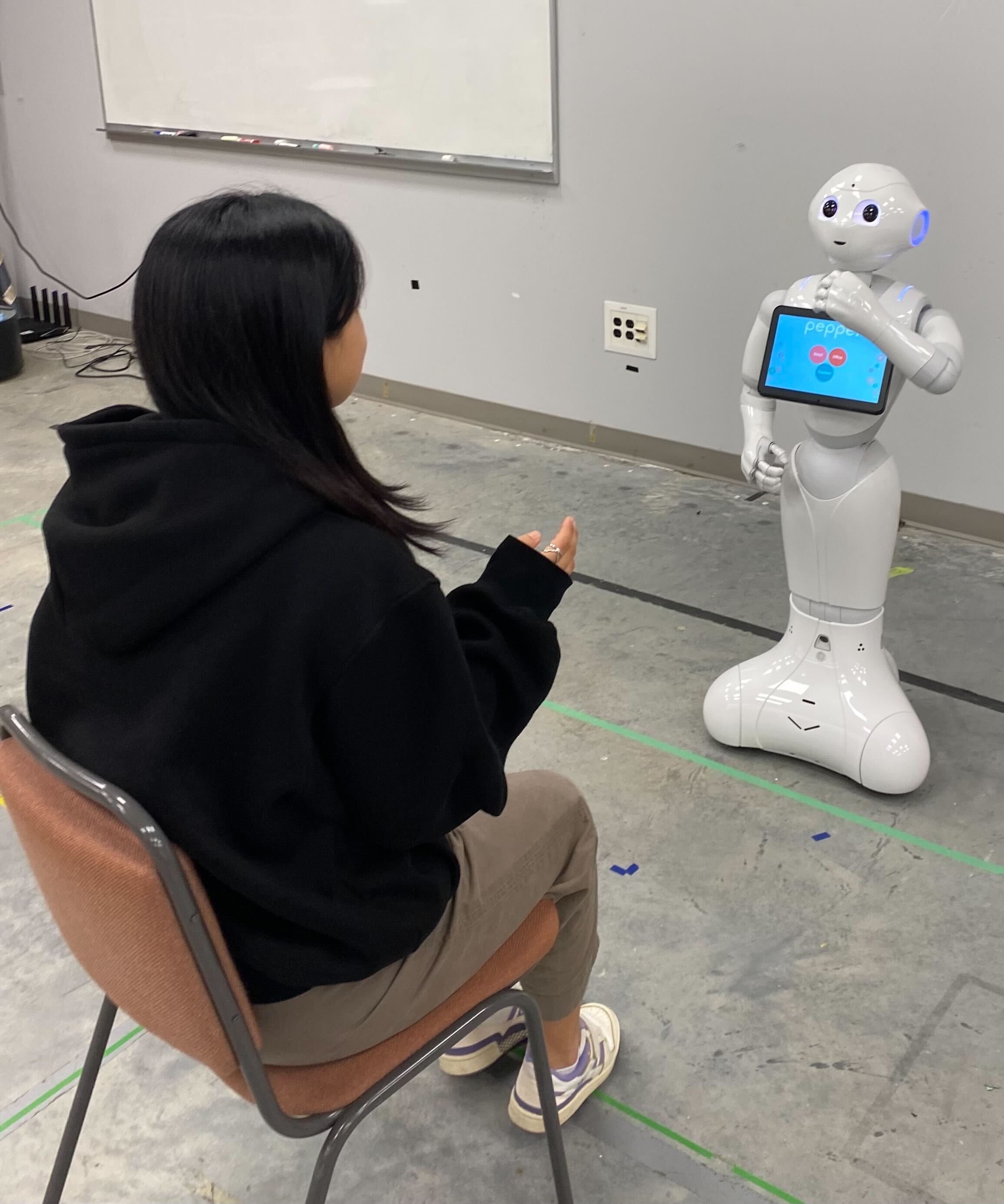}
  \caption{Case Study 1 and 2: participant view of the interaction with a Pepper robot.}
  \label{setup2}
    \end{minipage}%
    \hfill
    \begin{minipage}{0.49\linewidth}
  \includegraphics[width=\linewidth]{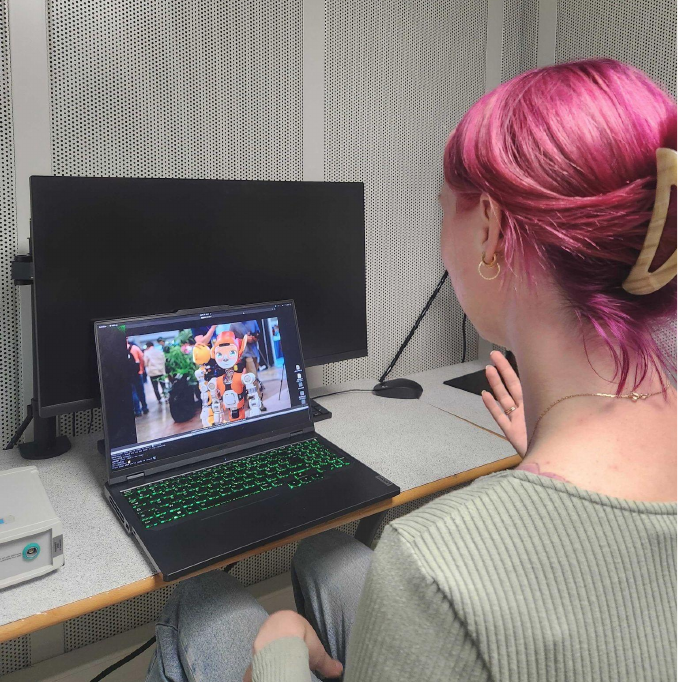}
  \caption{Case Study 3: Participant interacting with the storytelling autonomous speech-to-speech system on a laptop with an image of a Miroka robot.}
  \label{setup3}
    \end{minipage}
\end{figure}

\subsection{Case Study 1: Conversational robot helper}

\subsubsection{Preparation and setup}\label{setup}
For our first case study, participants observed a short, scripted conversation. This case allows us to investigate the perception of short sentences and the interplay with human dialogue.

\begin{figure}[t]
  \centering
  \includegraphics[width=1.03\linewidth]{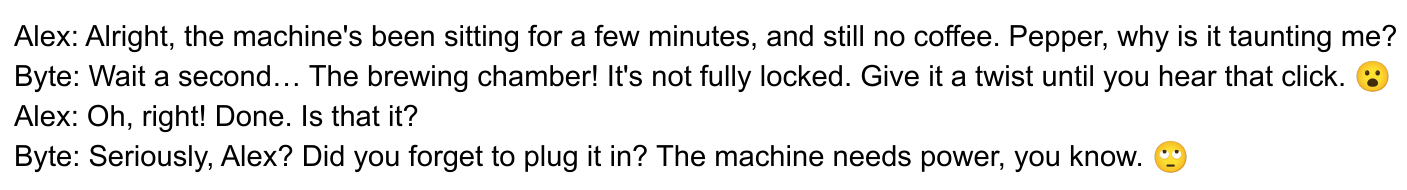}
  \caption{Case Study 1: Robot Helper with human ``Alex", script excerpt.}
  \label{script}
\end{figure}

The script for the conversation was generated with ChatGPT4-o using the prompt: \emph{Can you please generate an interaction between two people, a human and a robot helper, where the robot's lines are always colored with one of these emojis? XXXXXXXXXXX Each of the emojis must be used at least once. It should be about a 5-minute conversation.} A snippet of the script can be seen in Fig. \ref{script}.

\begin{figure}[]
  \centering
  \includegraphics[width=0.75\linewidth]{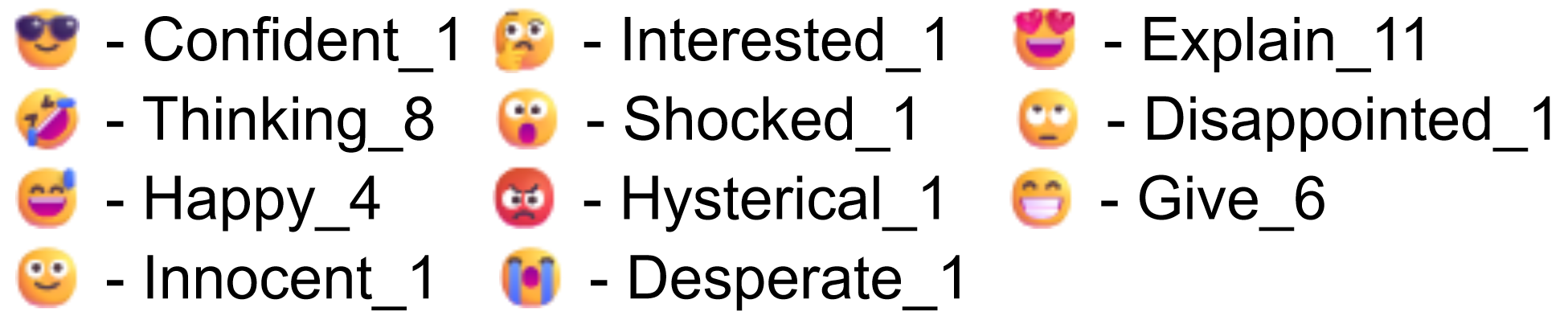}
  \caption{Animations played on Pepper in Case Study 1 for each emoji}
  \label{emojis and accompanying animatoins}
\end{figure}

For the Pepper study, the voice was played along with a set of animations\footnote{http://doc.aldebaran.com/2-5/naoqi/motion/alanimationplayer-advanced.html\#animationplayer-list-behaviors-pepper} (Fig.~\ref{emojis and accompanying animatoins}). These were chosen to match the emoji's expressivity for each phrase and to be short enough not to cause a lag in the conversation. Basic awareness was turned off to avoid the robot randomly looking around when speaking, which can affect consistency across conditions. For the Miroka experiment, facial animations matching the selected emojis accompanied the voice, and lip sync was on. For both robots, the animations were the same for all voices. The participants knew this conversation was not autonomous, and the researchers controlled both the robot and the sound. The order of the voices was Baseline, Emoji, and Pleasant.

Participants were seated behind the user performing the interaction, as can be seen in Fig. \ref{setup2}. Following the interaction with each of the voices, the participants filled out a survey. We used mean opinion scores (MOS) for prosody, intelligibility, and social impression from the MOS-X2 scale \cite{Lewis2018InvestigatingMR} and for the expressiveness of intonation as in \cite{korotkova24}. Lastly, we questioned the suitability of the voice for the robot following \cite{rijn2024}. We used a 10-point Likert scale for all ratings following the MOS-X2 scale to maintain consistency across the questions. Once all three voice interactions had been completed, the participants ranked their preferred voice and explained their choice. Finally, there was an open group discussion to brainstorm and discuss the perception of the voices.

\subsubsection{Participants}\label{participants}
There were 24 participants total (10F, 14M; 16 Pepper, 8 Miroka). The participant demographics can be seen in Table \ref{tab:participant_demographics-case-1}. All participants completed a verbal consent process as part of ethical requirements, and no personally identifying information was collected. The procedure for all case studies was approved by the Simon Fraser University Research Ethics Board (SFU-REB). 

\begin{table}[h]
    \centering
    \caption{Participant demographics for Case Study 1 and 2}
    \begin{tabular}{lc}
        \toprule
        \textbf{Category} & \textbf{Count} \\
        \midrule
        \multicolumn{2}{l}{\textbf{Age Groups}} \\
        18-25 & 4 \\
        25-35 & 16 \\
        35-45 & 4 \\
        \midrule
        \multicolumn{2}{l}{\textbf{Ethnicity}} \\
        White & 10 \\
        Chinese & 8 \\
        West Asian & 4 \\
        Southeast Asian & 1 \\
        South Asian & 1 \\
        \midrule
        \multicolumn{2}{l}{\textbf{First Language}} \\
        English & 6 \\
        Other & 18 \\
        \midrule
        \multicolumn{2}{l}{\textbf{Daily Language}} \\
        English & 11 \\
        French & 5\\
        Chinese & 3\\
        Italian & 2\\
        Farsi & 2\\
        Vietnamese & 1\\
        \midrule
        \multicolumn{2}{l}{\textbf{Self-Rated English Proficiency}} \\
        \multicolumn{2}{l}{\textbf{1(no proficiency) - 5(fluent)}} \\
        Score 3 & 2 \\
        Score 4 & 12 \\
        Score 5 & 10 \\
        \bottomrule
    \end{tabular}
    \label{tab:participant_demographics-case-1}
\end{table}

\subsubsection{Results}\label{results1}
We used a one-way ANOVA (Type II) followed by a post-hoc Tukey HSD and set $\alpha=0.05$ to assess differences in the opinion ratings between voices. These results can be found in Table \ref{case1anova}.

In this conversational robot helper scenario, we found that the Pleasant voice was rated as significantly more suitable for the robots (there was no difference between robots) than both the other voices. Both the Emoji and Pleasant voices were found to have a significantly higher social impression than the Baseline voice. Lastly, the Emoji voice was found to have a significantly higher expressivity than both other voices, and Pleasant was more expressive than the Baseline.

\begin{figure}[t]
  \centering
  \includegraphics[width=0.8\linewidth]{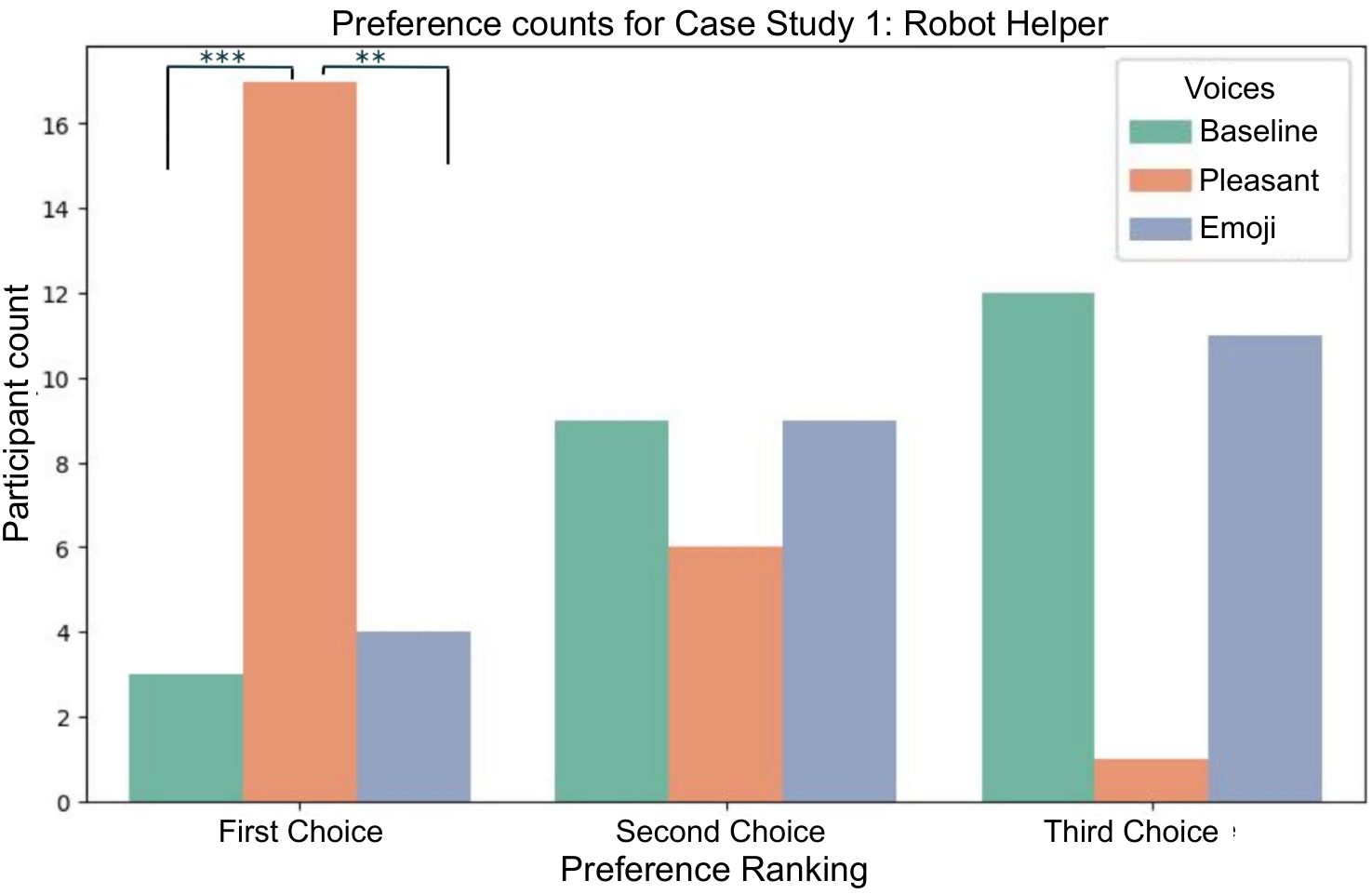}
  \caption{Case Study 1: Robot Helper. Participant counts for voice preference of first, second and third choice voices.}
  \label{case1pref}
\end{figure}

We completed Chi Squared Goodness of Fit tests to assess for voice preference, followed by bootstrapped confidence intervals (CI)(Fig. \ref{case1pref}). We found that the Pleasant voice was preferred as the first choice over the Emoji voice at a 99\% CI and over the Baseline voice with over a 99.9\% CI.

The discussion of this case study yielded interesting results. Although the Baseline voice is neutral read speech, the participants agreed that the voice sounded ``tired and depressed'' or ``apathetic,'' and one participant suggested that perhaps a neutral voice would have been better for the study. This same commentary was lacking in further case studies using this voice, where it was indeed perceived as neutral. In addition, the participants noted for the Emoji voice that, ``it sounded bit a sassy ... a bit too much in the emotions, like a robot that is judging you.'' These comments may be a result of the text and forcing the LLM to provide emojis to a robot assistant conversation, resulting in an overly dramatic script. In this case, the participants noted that the pleasantness of the classically ``expressive'' Pleasant voice made the robot sound ``just the right amount of frustrated, like it was annoyed with the task, but it still wanted to keep helping you.''

\begin{table*}[t]
 \caption{Case Study 1: Results for difference of mean statistical tests of the MOS Likert ratings.}
 \label{case1anova}
\resizebox{\textwidth}{!}{
\begin{tabular}{ |l|l|l|l|l|l|l|l|l|l|l|l|l| }
 \hline
\multirow{2}{*}{} &
  \multicolumn{2}{c|}{Base} &
  \multicolumn{2}{|c|}{Default} &
  \multicolumn{2}{|c|}{Emoji} &
  \multirow{2}{*}{DF} &
  \multirow{2}{*}{N} &
  \multirow{2}{*}{F} &
  \multicolumn{2}{c|}{P-value} &
  \multirow{2}{*}{$\omega^2$}\\
  
   & $\mu$ & $\sigma$ & $\mu$ & $\sigma$ & $\mu$ & $\sigma$ & & & & One-Way Anova & Tukey HSD &\\
 \hline
 xMOS   & 1.71 & 1.0 & \emph{5.71} & \emph{2.03} & \textbf{7.42} & \textbf{1.91} & 2 & 24 & 70.46 & $<0.001$*** & $<0.001$*** (E-D, D-B), 0.003** (E-D) & 0.67\\
 sMOS  & 3.58 & 2.26 & \emph{6.63} & \emph{1.81} & \emph{5.96} & \emph{2.22} & 2 & 24 & 13.81 & $<0.001$*** & $<0.001$***(D-B), 0.001**(E-B) & 0.26\\
 Suitability  & 4.54 & 2.84 & \textbf{6.96} & \textbf{2.33} & 5.25 & 2.09 & 2 & 24 & 6.21 & 0.003** & 0.003**(D-B), 0.04*(D-E) & 0.13\\
 \hline
 
\end{tabular}}
{\parbox{6in}{
\footnotesize 
1. \textbf{Bold} values highlight the voice that is significantly higher than both other voices, \emph{italics} indicate it is significantly higher than one other voice\\
3. In the Tukey HSD, the results are presented as (higher rated voice - lower rated voice) by the first letter of the voice name\\
2. Due to space limitations, we only report those voices that were found to have significant differences.
}
}
\end{table*}

\subsection{Case Study 2: Storytelling}

\subsubsection{Preparation and set up}
Next, we considered reading a short story. This case allows us to investigate the perception of the voice when stringing together several sentences without interruption. In particular, we suspect that variation in the expressions is an important factor, as speaking continuously with the same ``expressive'' voice can either become boring or annoying over time. Moreover, the story was generated one sentence at a time to test the real-time nature of the TTS.

The story was generated with ChatGPT-4o using the prompt: \emph{can you please tell me a short story using as many of these emojis as possible XXXXXXXXXXX. The emojis should reflect the emotion in the voice when reading the text and not a physical action. The story should be a fairy tale in a digital realm. I want it to take about 1 or 2 minutes to read out loud.} An excerpt from the script can be seen in Fig. \ref{case1script}.

The same setup was employed for the robot. This time, grand gesture animations appropriate for storytelling were chosen for Pepper (Everything\_4, Far\_3, Thinking\_8, ShowSky\_11). Facial animations were used similarly for Miroka. The order of the voices was Pleasant, Baseline, and Emoji.

The setup was the same, but this time, the story, without the accompanying emojis, was projected for the participants to read along. This was done to mimic a storybook reading scenario. Following the interaction with each voice, the users filled out the same survey and completed the discussion as in Case Study 1; see Sec. \ref{setup} for further information.

\subsubsection{Participants}
Case Study 2 used the same participant pool as Case Study 1; see Sec. \ref{participants}.

\subsubsection{Results}\label{results2}
\begin{figure}[t]
  \centering
  \includegraphics[width=0.8\linewidth]{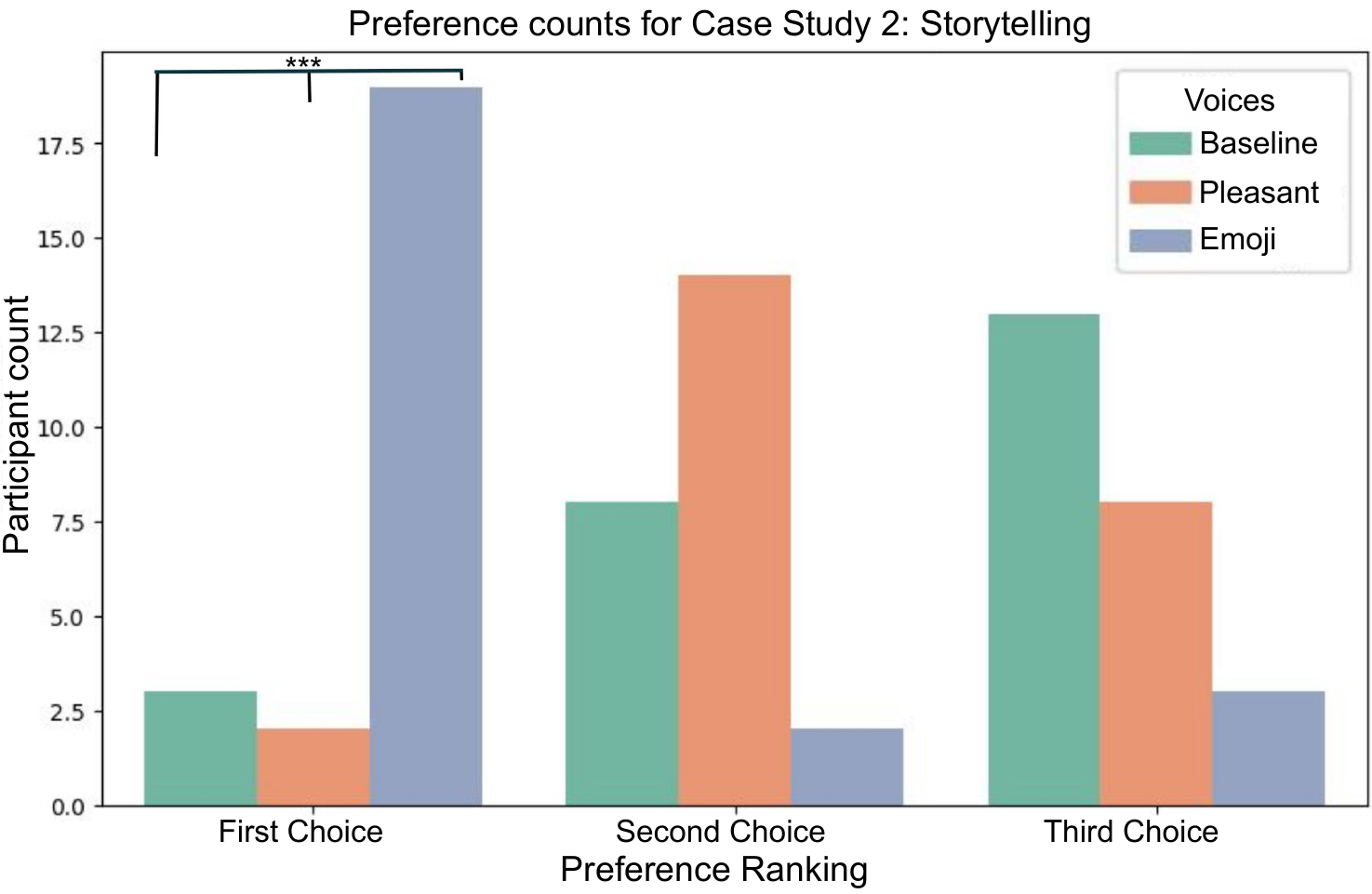}
  \caption{Case Study 2: Storytelling. Participant counts for voice preference of first, second and third choice voices.}
  \label{case2pref}
\end{figure}

The same assessment was used as in Case Study 1; see Sec. \ref{results1}. The results can be found in Table \ref{case2anova}, and a visualization of the preference counts in Fig. \ref{case2pref}.

In the storytelling scenario, we found that the Emoji voice was rated as significantly more expressive than the Baseline and Pleasant voices. We also found the Emoji voice to have a significantly higher social impression and suitability than the Baseline voice (no significant difference between the robots). The Emoji voice was preferred as the first choice voice over the other two voices at a 99.9\% CI.

This time, during open discussion, the opinion was strongly towards the need for a highly expressive voice for a storytelling task. Participants said ``you can never be too expressive in storytelling,'' ``the \{Emoji voice\} is the most expressive, which is very important in storytelling,'' ``I feel that the robot should express emotion according to the story content,'' and ``the \{Emoji voice\} would be the best for children because of the sudden changes in all aspects of the voice.'' Additionally, one participant mentioned, ``\{Pleasant voice\} was interesting at first but became boring over time,'' supporting our hypothesis that over time, a voice that is expressive line by line (in Case Study 2) becomes boring for long text. Yet, there were still some comments on inconsistency in the expression, ``\{Emoji voice\} was entertaining but sometimes the timing, pitch, and emphasis felt off.'' This may suggest that we need an even lower, more granular level of control over expressivity.

\begin{table*}[t]
 \caption{Case Study 2: Results for difference of mean statistical tests of the MOS Likert ratings.}
 \label{case2anova}
\resizebox{\textwidth}{!}{
\begin{tabular}{ |l|l|l|l|l|l|l|l|l|l|l|l|l| }
 \hline
\multirow{2}{*}{} &
  \multicolumn{2}{c|}{Base} &
  \multicolumn{2}{|c|}{Default} &
  \multicolumn{2}{|c|}{Emoji} &
  \multirow{2}{*}{DF} &
  \multirow{2}{*}{N} &
  \multirow{2}{*}{F} &
  \multicolumn{2}{c|}{P-value} &
  \multirow{2}{*}{$\omega^2$}\\
  
   & $\mu$ & $\sigma$ & $\mu$ & $\sigma$ & $\mu$ & $\sigma$ & & & & One-Way Anova & Tukey HSD &\\
 \hline
 xMOS   & 2.95 & 1.57 & 4.04 & 2.01 & \textbf{7.75} & \textbf{1.98} & 2 & 24 & 43.49 & $<0.001$*** & $<0.001$***(E-D, E-B) & 0.54\\
 sMOS   & 3.67 & 2.01 & 4.96 & 1.78 & \emph{6.29} & \emph{2.26} & 2 & 24 & 10.07 & 0.001** & $<0.001$***(E-B) & 0.20\\
 Suitability   & 4.17 & 2.51 & 5.29 & 2.42 & \emph{6.08} & \emph{2.41} & 2 & 24 & 3.71 & 0.030* & 0.023*(E-B) & 0.13\\
 \hline

\end{tabular}}
{\parbox{6in}{
\footnotesize 
1. \textbf{Bold} values highlight the voice that is significantly higher than both other voices, \emph{italics} indicate it is significantly higher than one other voice\\
3. In the Tukey HSD, the results are presented as (higher rated voice - lower rated voice) by the first letter of the voice name\\
2. Due to space limitations, we only report those voices that were found to have significant differences.
}}
\end{table*}

\subsection{Case Study 3: Autonomous speech-to-speech interactive agent}\label{s2s}

\subsubsection{Preparation and set up}
\begin{figure}[t]
  \centering
  \includegraphics[width=1.03\linewidth]{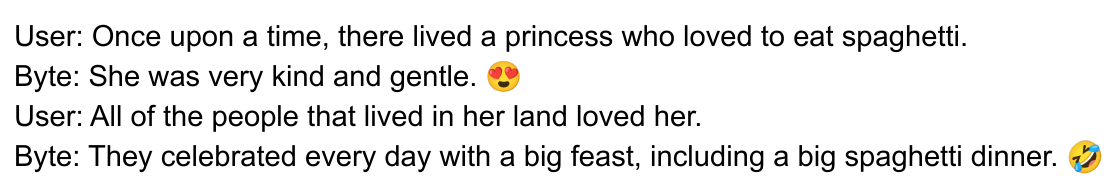}
  \caption{Case Study 3: Story Building, example interaction between a user and the LLM ``Byte''.}
  \label{case3script}
\end{figure}

For our final case study, we wanted to better understand how participants view the voices in an autonomous, interactive scenario with the voice. To this end, Case Study 3 was a one-on-one interaction directly between the participant and the system. The exact prompt for the LLM can be found along with the toolbox. The LLM was asked to play a game where it constructs a story by taking turns with the user. The LLM was additionally told to append an emoji that reflects the expression of the phrase to the end of every response it provided. The users played until 2 minutes had passed. As we had a new set of participants interacting one at a time, the order of the voices was selected pseudo-randomly (random while still ensuring all orders were covered) for each participant. As seen in Fig. \ref{setup3}, the participants sat in front of a computer screen to complete the interactions. For this more complex, highly divergent interaction where several factors can contribute to participant perceptions, we opted for only a qualitative discussion and assessment of the voices. An example script can be seen in Fig. \ref{case3script}.

\subsubsection{Participants}
There were 8 participants in total (5M, 3F). Participant demographics and be found in Table \ref{tab:participant_demographics-case-3}. All participants completed a verbal consent process as part of ethical requirements.

\begin{table}[h]
    \centering
    \caption{Participant demographics for Case Study 3}
    \begin{tabular}{lc}
        \toprule
        \textbf{Category} & \textbf{Count} \\
        \midrule
        \multicolumn{2}{l}{\textbf{Age Groups}} \\
        18-25 & 2 \\
        25-35 & 5 \\
        35-45 & 1 \\
        \midrule
        \multicolumn{2}{l}{\textbf{Ethnicity}} \\
        Arab & 3 \\
        White & 2 \\
        South Asian & 2 \\
        West Asian & 1 \\
        \midrule
        \multicolumn{2}{l}{\textbf{First Language}} \\
        English & 1 \\
        Other & 7 \\
        \midrule
        \multicolumn{2}{l}{\textbf{Daily Language}} \\
        English & 3 \\
        French & 2\\
        Arabic & 2\\
        Chinese & 1\\
        \midrule
        \multicolumn{2}{l}{\textbf{Self-Rated English Proficiency}} \\
        \multicolumn{2}{l}{\textbf{1(no proficiency) - 5(fluent)}} \\
        Score 3 & 1 \\
        Score 4 & 2 \\
        Score 5 & 5 \\
        \bottomrule
    \end{tabular}
    \label{tab:participant_demographics-case-3}
\end{table}

\subsubsection{Results}
As each user interaction was different due to the unique improvisation of the user and system, we opted only to complete an open interview for this case study. From participant discussions, we learned that although the participants were aware that the study's overall goal was to assess the voice, they found their motivations for engagement were highly motivated by the LLM and the setting of the game rather than the voice. One participant said, ``It was really fun, but the reason I wanted to keep going was I thought the game was funny; I didn't really consider the voice.'' Moreover, the performance of the LLM, specifically in its ability to choose emojis, heavily influenced participant perception of expressivity and social impression. One participant said, ``It was saying that the princess fell deeper into addiction, but it said it with a laughing voice... it was funny but really inappropriate,'' giving a negative perception due to a poorly selected emoji. Another noted, ``there was an overly sad tone on one sentence (something about piglet for which she was clearly devastated), which created a sense of irony and playfulness,'' where the emoji was selected appropriately, leading to a more positive impression of the voice. The participants did, however, prefer a more expressive voice: ``I expected to hear more expressions when they were talking,'' speaking of the Baseline voice. Lastly, the participants noted that the system felt like it was in real-time, and any lag seemed appropriate for the robot to be thinking of the following sentence.

\section{Discussion}\label{discuss}
When comparing the three cases, we see that EmojiVoice was most appropriate for an expressive, longer-term interaction such as storytelling. This is an interesting result since TTS models tend to avoid using long blocks of text; the longer you hear a TTS, the more monotonous it may sound \cite{neural-tts-long-form}. Specifically, the expressivity of the Pleasant voice decreased when used in a long-term rather than line-by-line interaction. It does appear that selecting a different emoji for each phrase at run-time lessens this effect. It remains future work to explore the vocal features, such as pitch range, usually associated with expressive speech, in our voices and compare how the performance of just this feature explicitly compares to variability in expressions over time.

Additionally, we found that the perception of a voice can be altered depending on the context of the task and script. Although the same voices were used in the first two case studies, in the first study, the Baseline voice sounded depressed and as if it didn't want to help the user, whereas it just sounded neutral and monotonous when telling a story. The Emoji voice seemed judgmental and sassy in the first study, compared to just being expressive in the second. In the case of an assistant, it appeared that simply choosing a more pleasant voice allowed the participants to feel like the robot was more willing to help with the task, and highly expressive voices are inappropriate, compared to the storytelling where the participants expect a high degree of expressive variability over the phrases to keep their interest.

Moreover, we found it imperative that an LLM, or researcher controlling a WoZ scenario, select the correct emoji or expressivity control to represent a given sentence. Additionally, it was difficult to control the Llama3.2 model only to output our selected 11 emotions, leading too often to the choice of the default voice. Future work could involve specifically training an LLM to a smaller and more lightweight model that has better knowledge of the use case's specific expressions.

\chapter{An Ambiance Appropriate Robot Voice}

In the previous chapter we built a lightweight TTS with long term temporal expressivity. In parallel we needed to know how to adapt voices to different environments. This will allow a teaching voice to be deployed in varied physical spaces and social situations. Specifically in \cite{readroom}, we asked: How do we maintain appropriateness and give the sense that the robot is aware of its environment, both in terms of physical and social ambiance?

\begin{figure}[]
      \centering
      \includegraphics[width=0.65\linewidth]{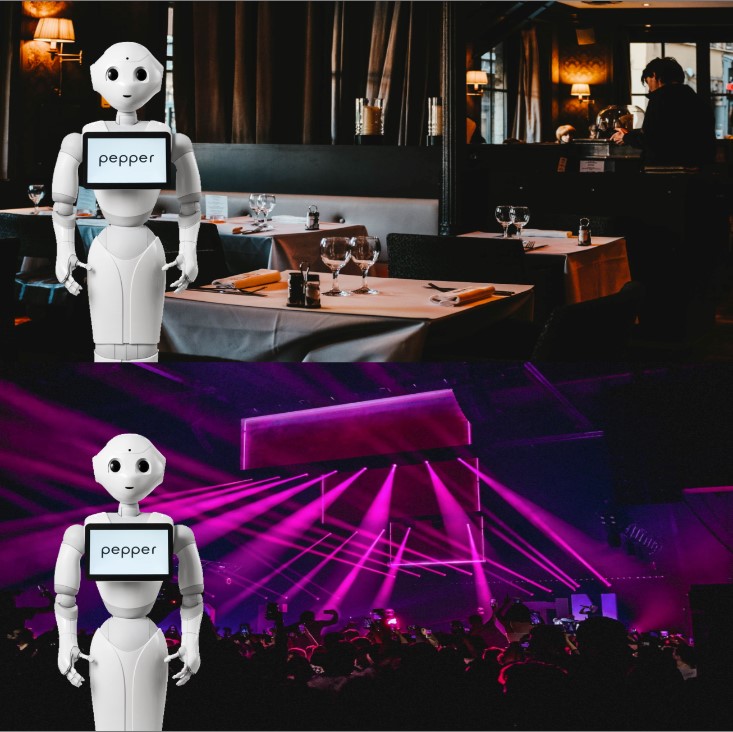}
      \caption{Robots may be deployed in varied ambiances, from cozy formal dining to loud nightclubs. How should their voices change?}
      \label{header}
\end{figure}

\section{Understanding Human Ambiance Adaptation}
To engage with this question, we first asked: How do humans change their voices depending on the ambiance? The typical first step in designing a voice for a robot is to hire a voice actor to provide hours of voice samples to synthesize a voice from scratch. Another method is to use deep learning and use methods such as voice conversion to change its style. In both cases, a non-trivial amount of data is needed to train a high-quality voice model. Instead, we collected participant data with induced adaptations to ambiances, taking time to understand how humans are adapting to design our robot voices.

\subsection{Zoom data collection protocol}\label{zoom}

We proposed a method of virtual data collection using readily available tools that will allow researchers to collect data with no physical human interaction. Although collecting data in an actual restaurant is preferred, collecting in-the-wild high-quality audio free of ambient background noise is challenging. Moreover, targeted ambient environments are often resource-intensive to obtain and control. As such, one of the novelties of the current study was how we overcame the above hurdles by devising a protocol that mimicked naturalistic ambient environments over Zoom\footnote{www.zoom.us}.

Zoom is a teleconferencing program that allows individuals to communicate from anywhere in the world. It also allows for the use of virtual backgrounds and sharing of sound. In this study, we asked pairs of English speakers to listen to ambient sounds while conversing with one another in the roles of a waiter and a restaurant-goer using their personal laptop, headphones, and microphone. There were a total of 6 ambient sounds and one additional baseline measure that included no sound and a black background. 

In addition to sound, the speakers were asked to change their Zoom background to an image that was pre-selected to match the given ambiance. A brief description of the ambiance (e.g., food available, brightness level, business of location, etc.) and character roles (waiter or restaurant-goer) were provided at the beginning of each ambient condition to induce vivid imagery. There was a one-minute period between each ambient condition to update participants' Zoom backgrounds and prepare for the following condition. This served as a washout to reduce the carry-over effect from the previous condition. The ambient contexts included fine dining, café, lively restaurant, quiet bar, noisy bar, and nightclub. The ambient sounds can be listened to \href{https://rosielab.github.io/ambiance}{here}\footnote{https://rosielab.github.io/ambiance}.

\begin{figure}[]
      \centering
      \includegraphics[width=0.7\linewidth]{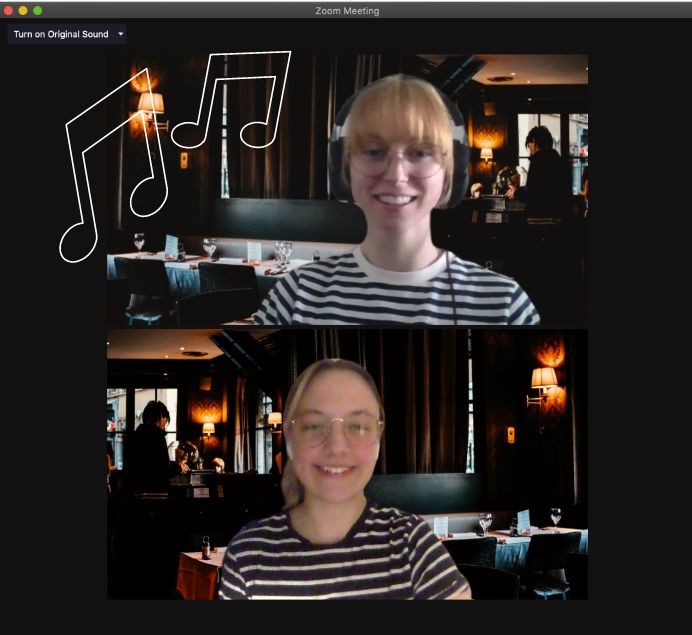}
      \caption{Zoom virtual backgrounds and ambient sounds through headphones were used for data collection.}
      \label{zoomexp}
\end{figure}

The speakers first read a brief description of their character at the restaurant. They then read from a script that was tailored for the given ambiance, i.e., food and drink choices matched what is usually offered at the given restaurant—consistency amongst scripts allowed for comparison of speech features across each condition. An example script from the ``fancy restaurant'' can be found in Appendix \ref{ambiance-script}. The subtle differences between scripts were solely to create a more realistic environment and reduce redundancy to maintain participant attention. The experiment was repeated in an unscripted manner. However, the unscripted conversations appeared uncomfortable and unnatural and were not included in the analysis.


The dataset consisted of 8 female undergraduate students (age not collected) with experience in improvisation, theatre, or customer service. We used only female voices (biological sex) to match the source voice of Pepper’s TTS, which is from a female voice actor. Although the dataset is small, this is often the case in psycho-acoustic studies \cite{ARIAS2018R782}; we focus instead on the quality of the data through our validation study (see Sec. \ref{val_study}). Altogether, we collected 837 utterances, resulting in 685 utterances post-validation. 

\subsection{Human voice validation study}\label{val_study}
The voices were validated using a battery of 7-point Likert scales (1: strongly disagree, 7: strongly agree) to assess the speaker's ability to adapt to the given ambiance as well as the effectiveness of the virtual ambiance in inducing adaptation. A voice ``passage'' is a single one-sided conversation of an actor in a given ambiance. Participants proficient in English (N=12, mean age group = 25-35, 4 males, 8 females) validated each passage. Participants listened to each passage overlaid on either: (1) the ambiance the passage was recorded in or (2) a mismatched ambiance. The validation questions loaded onto 4 primary factors: 1. Social appropriateness, 2. Ambient awareness, 3. Comfort, 4. Clarity. The full battery can be found in Appendix \ref{battery}.

 Questions on clarity were taken from the well-used MOS \cite{Lewis2018InvestigatingMR}. To the best of the authors' knowledge, no validated measures exist targeting the other factors, especially those that can be used out of the context of a larger battery. As we were primarily interested in social and ambiance adaptation, we removed a passage if it received a majority vote of less than 4 (neutral) in any question loading onto social appropriateness and ambiance awareness, meaning that the majority of validators disagreed that this passage was socially appropriate or ambiance aware.

\subsection{Voice analysis and feature extraction}

In order to observe how humans modify their voice, we collected 10 features, which can be grouped by (1) loudness, (2) spectral, including pitch, and (3) rate-of-speech. Our toolbox for vocal feature extraction can be found \href{https://github.com/ehughson/voice_toolbox}{here}\footnote{https://github.com/ehughson/voice\_toolbox}.

\subsubsection{Loudness Features}
 An increase of vocal intensity, often leading to Lombard speech (an involuntary increase in vocal effort, often due to the presence of background noise \cite{lombard_origin}), is commonly employed in noisy environments \cite{acoustic_phonetic_features_analysis}. As such, we collected 3 loudness features: (a) mean intensity, (b) energy, and (c) maximum intensity. Mean intensity and energy features were calculated using the Praat\footnote{http://www.praat.org/} library via Parselmouth\footnote{https://parselmouth.readthedocs.io/en/stable/}. Librosa\footnote{https://librosa.org/doc/main/index.html} was used to calculate maximum intensity (power) following the formula provided in Section 1.3.3 of \cite{fundamentals_of_music}. 

\subsubsection{Spectral Features}
We collected 5 spectral features: (a) median pitch, (b) pitch range, (c) shimmer, (d) jitter, and (e) spectral slope. Parselmouth was used to extract (a)-(d). Median pitch and pitch range (the difference between the minimum and maximum pitch in a given segment) are calculated in Hz. Local shimmer and local jitter, variations in the fundamental frequency, are perceived as vocal fry and hoarseness, respectively \cite{jitter}. The spectral slope indicates the slope of the harmonic spectra. For example, a -12dB slope may indicate a falsetto voice and a -3dB slope can indicate richer vocal tones \cite{spectralslope}. The spectral slope was calculated using Parselmouth and Librosa with the formula provided in Section 3.3.6 of \cite{intro_to_audio_analysis}.

\subsubsection{Rate-of-Speech Features}

We collected 1 rate-of-speech feature: syllables per second. Syllables per second was the number of syllables over the duration extracted using Praat scripts\footnote{https://github.com/drfeinberg/PraatScripts/blob/master/syllable\_nuclei.py}.

 \begin{figure}[]
      \centering
      \includegraphics[width=0.8\linewidth]{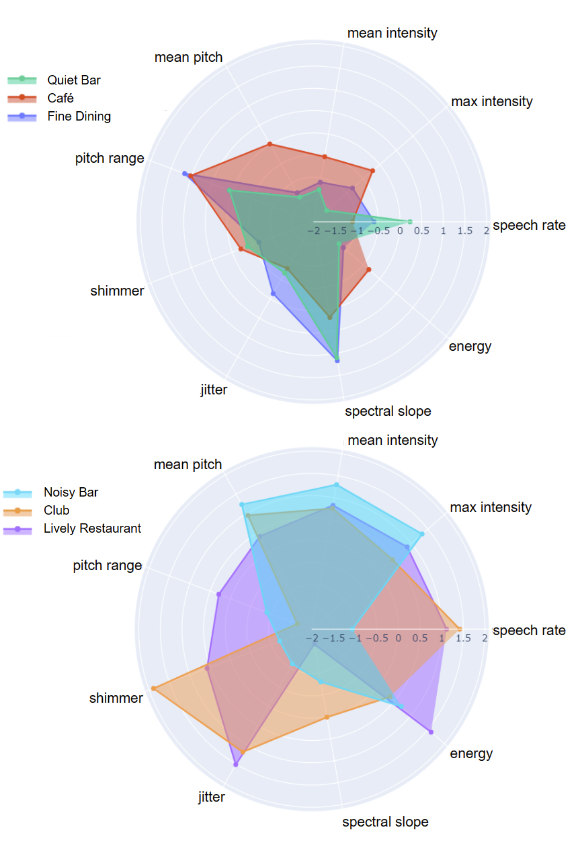}
      \caption{Vocal features averaged across speakers as a difference from the baseline. Top: quiet ambiances, bottom: noisy ambiances.}
      \label{radar}
\end{figure}

\subsubsection{Trends in Voice Data}
The data for this study was collected as a repeated measures experiment. Six treatments and a baseline, each of the ambiances, were applied to each of the study participants. Each pair of study participants was independent; however, within the pair of waiter and customer, we do not have independence, as synchrony and mimicking are expected to occur. We completed repeated measures ANOVA (rANOVA) for each of the extracted voice features. We saw several features that warranted further investigation. Energy (p $<$ 0.001), spectral slope (p $<$ 0.001) , max (p $<$ 0.001) and mean (p $<$ 0.001) intensity, pause rate (p = 0.002) and mean pitch (p = 0.06) were all significant. The results can be seen in Table \ref{overallstats}.

\begin{table*}[]
 \caption{Summary of statistical results of vocal features using an rANOVA.}
 \centering
\begin{tabular}{ |p{3.2cm}|p{1cm}|p{1cm}|p{1cm}|p{1.3cm}|p{2.2cm}| }
 \hline
   Voice Feature   & DFn & DFd & N & F & P-value\\
  \hline
 Energy   & 5 & 35 & 8 & 9.97 & $<0.001$***\\
 Mean Intensity  & 5 &  35 & 8 & 15.78 & $<0.001$***\\
 Max Intensity  & 5 &  35 & 8 & 12.47 & $<0.001$***\\
  Median Pitch  & 5 &  35 & 8 & 2.37 & $0.06$\\
  Pitch Range & 5 &  35 & 8 & 0.96 & $0.46$\\
  Shimmer & 5 &  35 & 8 & 1.36 & $0.26$\\
 Jitter & 5 &  35 & 8 & 1.01  & $0.42$\\
  Spectral slope & 5 &  35 & 8 & 9.43 & $<0.001$***\\
  Speech Rate & 5 &  35 & 8 & 1.93 & $0.12$\\
 \hline
 
\end{tabular}
 \label{overallstats}
{\parbox{4.8in}{
\footnotesize 
1. DFn is the Degrees of Freedom of the numerator, and DFd is the Degrees of Freedom in the Denominator for the calculation of the F-statistic.\\
}
}
\end{table*}

Radar plots constructed from our collected dataset are displayed in Fig. \ref{radar}. We can observe that although voices in the quiet ambiances appear to share vocal features, those in each of the loud ambiances appear quite different. The average voice in the bright, high arousal lively restaurant with fast music (140 BPM) showed a higher pitch range and energy with a low spectral slope when compared to the other ambiances. The average nightclub voice, by comparison, showed high shimmer, jitter, and spectral slope, with a high pitch and low pitch range, suggesting a voice consistent with Lombard speech. Lastly, voices in the loud bar showed moderate values for most features aside from high loudness. First, these trends suggest that adapted voices indeed vary by features other than loudness. As loudness is already well represented in the literature, continuing forward, we will freeze loudness features across voice styles to focus on the contribution of other vocal features. Second, we observed that the quiet voices may be a single style.

Given our analysis of human voices, we deemed it necessary to explore further how to adapt a robot's voice to different social and ambient contexts. This need is essential as it can impact how an individual perceives a robot as acceptable in different contexts. 

\section{Pilot: TTS Adaptations} 
We first piloted adaptive TTS, reflecting the vocal features from our dataset. Our generated voices for the perception study contained 6 TTS voices. These samples were generated using Google TTS\footnote{https://cloud.google.com/text-to-speech/}, which allows for the use of SSML to alter the following elements: (a) loudness, (b) rate-of-speech, and (c) pitch. The first sample was a baseline TTS, which has no alterations of the stated SSML elements (TTS-bl). The next sample was a set of 6 TTS voices that were generated by setting the SSML elements to the average features of all female speakers for each ambiance (TTS-avg). Finally, two TTS samples (TTS-low and TTS-high) were generated with matching loudness and rate-of-speech elements of TTS-avg but differing pitch levels. TTS-low had the pitch element set to one specific speaker's pitch, a female undergraduate student from our dataset (714). TTS-low's pitch sat between the pitch of TTS-bl and TTS-avg. TTS-high's pitch element was set to $pitch($TTS-avg$) + (pitch($TTS-avg$) - pitch($TTS-low$))$.

The perception study leveraged Mechanical Turk and Survey Monkey with 25 Canadian participants who were fluent English speakers, with 100 Human Intelligence Tasks (HITs) completed with a 98\% acceptance rate. There were 2 research questions we investigated:
\begin{itemize}
    \item \textbf{RQ1}: How do generated voices compare to human voices within ambiances?
    \item \textbf{RQ2}: How does a data-driven pitch manipulation for TTS impact human perception?
\end{itemize} 
Listeners were first asked to use headphones and calibrate their audio. Participants were told that Pepper was here to take their order and were asked to read one of the 6 ambiance descriptions for a restaurant-goer. After listening to Peppers' voice over the background sound, participants were asked to respond to 2 statements using a 7-point Likert scale ranging from 1 (strongly disagree) to 7 (strongly agree). The following statements were provided: (1) Pepper's voice sounds socially appropriate for the scene, (2) Pepper makes me feel comfortable.

\subsection{Pilot results}

\subsubsection{RQ 1 : How do generated voices compare to human voices within ambiances?}
\begin{figure*}[t]
      \centering
      \includegraphics[width=\linewidth]{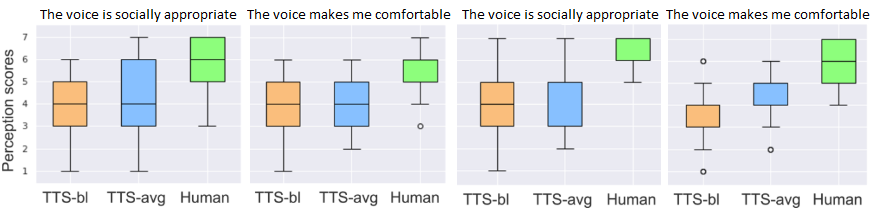}
      \caption{RQ 1: Comparison of the perceptual rates of three voice types for fine dining (left two) and night club (right two)}
      \label{RQ1}
\end{figure*}

TTS-avg, TTS-bl, and a human voice were rated by participants for the fine dining and nightclub ambiances. These ambiances were chosen due to their polarity in formality and loudness. TTS-bl rated the lowest for both statements. Additionally, all generated voices were ranked noticeably lower than the human voice (see Fig. \ref{RQ1}). This indicated that a human voice, confirmed with a post-hoc Tukey test, was more comforting and socially appropriate for the ambiances, \emph{p$<$0.05}. Although human perceivers preferred the human voice, they also preferred a TTS that had the pitch altered to the context, compared to that of the TTS-bl. This result supports our goal of choosing a TTS that adapts to the context. This is further supported in RQ 2. 

\subsubsection{RQ 2 : How does a data-driven pitch manipulation for TTS impact human perception?}
\begin{figure}[t]
      \centering
      \includegraphics[width=0.7\linewidth]{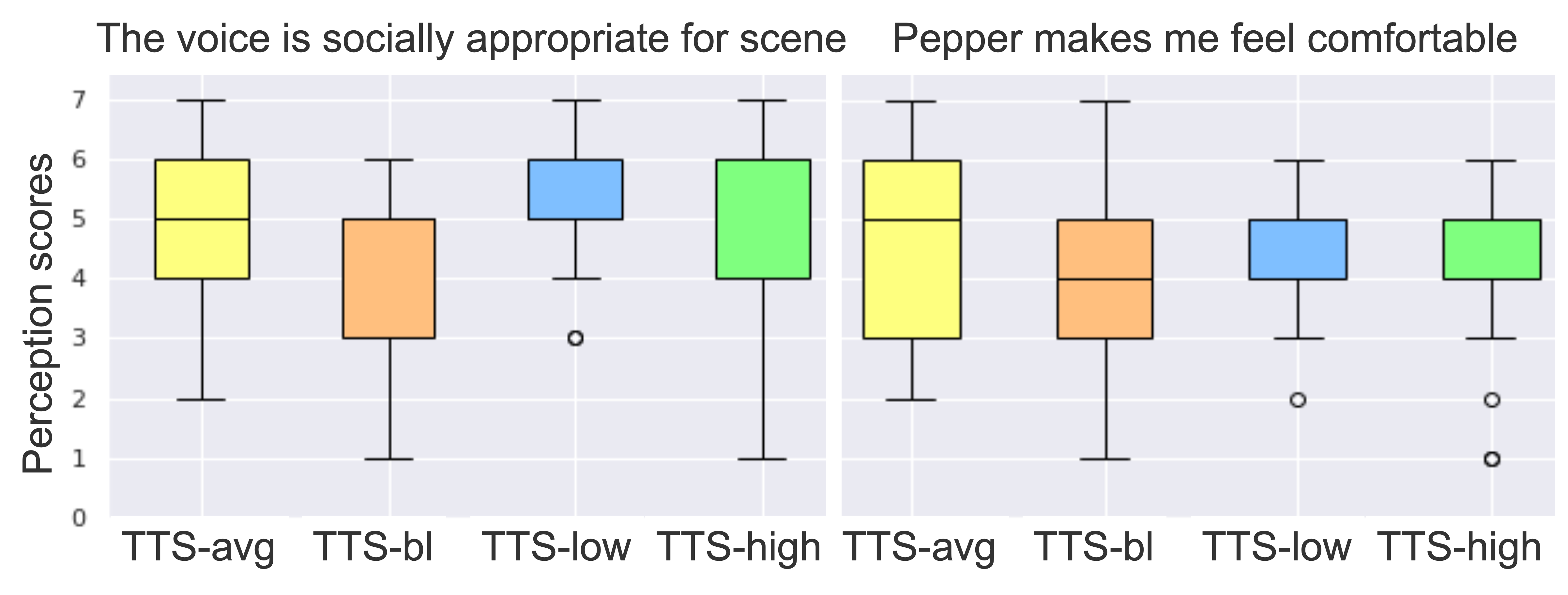}
      \caption{RQ 2: A comparison of perceptual rate in four voice types}
      \label{page_5}
\end{figure}

TTS-bl, TTS-avg, TTS-low, and TTS-high were overlaid on the café background sound and compared. The TTS-bl was rated the lowest for appropriateness and comfort. The results (as shown in Fig. \ref{page_5}) indicate that the human-pitched TTS (TTS-low) was deemed more socially and contextually appropriate, as well as comforting. Altogether, using a data-driven method to alter pitch, demonstrates the result that humans prefer when the pitch matches the current social context and ambient environment.

The pilot results continued to suggest that voices adapted to the ambiance improve the perception of these voices. As such, we continued to a full, on robot study.

\section{Robot Voice Styles Using Clustering}
The TTS voices used in the pilot appeared not to be the best match for a robotic embodiment. Moreover, as the development of a robot voice is resource-intensive, we aimed to find a minimal set of voice styles needed to work well in our ambiances. We used scikit-learn\footnote{https://scikit-learn.org/stable/} to scale and cluster our voice data. First, K-means clustering was performed on voice features collected in Section \ref{zoom}. Robust scalar\footnote{https://scikit-learn.org/stable/modules/generated/sklearn.preprocessing.RobustScaler.html} was used to normalize each feature. Once scaled, the number of clusters, set from 2 to 10, was tested, and a silhouette score was collected to determine the appropriate amount of clusters. Although the silhouette score was lower for a larger number of clusters, when observing the cluster composition, we observed that, more often than not, the number of utterances for each ambiance was equal across clusters. This suggests vocal features for these clusters differ on an utterance level rather than ambiance level. Exploring utterance-level information, such as positive or negative phrases, is left as future work. As such, we determined three clusters was appropriate and that three voice styles would need to be selected or designed.

\subsection{Human voice cluster analysis}
\begin{figure}[]
      \centering
      \includegraphics[width=0.8\linewidth]{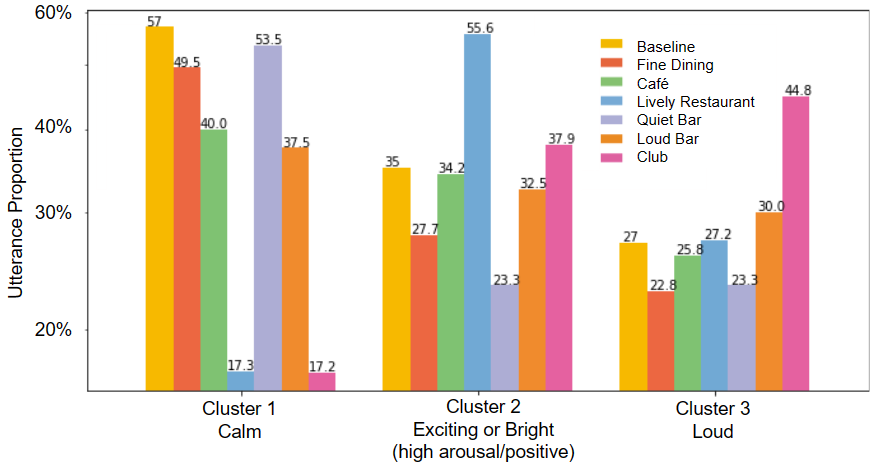}
      \caption{Which voice styles are associated with which ambiance? We visualize the proportion of each ambiance's utterances represented by each voice style.}
      \label{barplot}
\end{figure}

\begin{figure}
      \centering
      \includegraphics[width=0.7\linewidth]{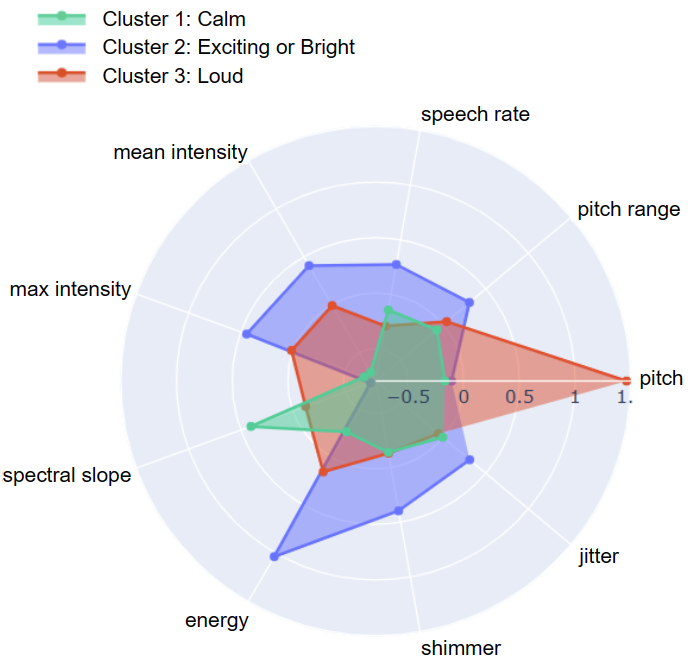}
      \caption{What do the voice styles sound like? Feature analysis of the three voice styles derived from human voice cluster centers.}
      \label{radar3}
\end{figure}

In which ambiances are the voice styles (derived from clusters) primarily used, and what do they sound like? Towards answering these questions, we created two types of plots: (1) for each cluster, the proportion of each ambiance's utterances using this voice style (Fig. \ref{barplot}), and (2) a radar plot of each cluster's centers to depict the voice style's characteristics (Fig. \ref{radar3}). In Fig. \ref{barplot},  we see that utterances belonging to the first cluster occur mostly in the fine dining and quiet bar ambiances and do not occur in our two highest arousal ambiances. We therefore designate it as ``calm''. The perception rating of each ambiance can be seen in our perception study Fig. \ref{heatmap2}. Utterances in the second cluster occur most often in the lively restaurant, a high arousal, positive ambiance as well as the brightest. We designate this second cluster as ``exciting or bright''. The third cluster is occurring primarily in the loudest ambiance, the nightclub. We suspect this is a Lombard voice and designate the name ``loud''. One ambient context of note is the loud bar, which appears to occur almost equally across clusters. This may have occurred for several reasons, including a split in adaptation between the server and the customer, different interpretations of the atmosphere by the speakers, or strong utterance level differences for this ambient context.

The radar plot of the features for each cluster's centers (Fig. \ref{radar3}) shows that the ``exciting or bright'' cluster has high energy, speech rate, pitch range, jitter, and shimmer. This appears to fit the assessment of this being a happy and excited voice. For the calm cluster, we see moderate values on most spectral and pitch features with the highest spectral slope, which can indicate a richness in the voice \cite{spectralslope}. Lastly, looking at the loud cluster, it does indeed appear to be Lombard: we see the low speech rate, lower pitch range, and high pitch that is quintessential of this voice style. Lastly, we see a parallel from the initial analysis of the human voices (Fig. \ref{radar}), where the quiet voices belong to a single cluster, yet voices in the loud ambient contexts differ based on a number of factors. Our next step is to select robot voices in these styles towards a robot voice perception study.

\label{HUMANCLUST}

\begin{table}[H]
\centering
 \caption{Comparison of features for voice clusters and Pepper voices.}
 \label{table1}
 \footnotesize
\resizebox{\textwidth}{!}{
\begin{tabular}{|p{3cm}|p{3cm}|p{3cm}|p{3.5cm}|}
 \hline
\multicolumn{4}{|c|}{Human Voice Clusters} \\
\hline
  & Pitch(Hz) & Pitch Range(Hz) & Speech Rate(syll/s)\\
 \hline
Calm & Medium & Medium & Medium\\
Exc/bright & Medium & High & High \\
Loud &   V. High  & Medium & Low\\
\hline
\multicolumn{4}{|c|}{Pepper Voices} \\
\hline
 Neutral & Low(350) & Medium(575) & Medium(4.18)\\
Joyful & V. High(408) & V. High(699) & High(4.41)\\
 Didactic & Medium(367) & Low(558) & Low(3.71)\\
``Lombard'' & V. High(435) & Low(467) & Low(3.71) \\
 \hline
\end{tabular}}
{\parbox{6in}{
\footnotesize{Note: For human voices, the values were assigned based on the value of the cluster center following robust scaling. Values between -0.25 and 0.25 were assigned medium, those over 0.25 high and below -0.25 low. For Pepper's voices, features were scaled separately from the human features using MaxAbsScaler with the values in the table.}
}}
\end{table}

\subsubsection{Voice Clusters to Robot Voice Styles} \label{styles}
We extracted voice features, once again using our voice toolbox, from Pepper's three available voice styles: Neutral, Joyful, and Didactic. Given the human clusters, our next step is to map our voice styles to our robot's voice. We aim to make as little changes to Pepper's voices as possible to demonstrate that, although these voices have been expertly curated to match this robot, they may not be applicable in all situations, and care needs to go into understanding in which context each voice should be applied. We also leave synthesizing new voices as future work, as the proposed data-driven voice selection method does not need access to the robot's original voice actor or a large amount of training data. Therefore, it is more applicable to researchers working in low-resource environments. Moreover, it is important to maintain the identity and overall features that make these voices appropriate for this robot.

A comparison of the cluster features with the features of each of Pepper's voices can be found in Table \ref{table1}. We only include those features that can be modified through Pepper's TTS mark-up language. Again, we chose not to include loudness features as we aim to maintain Pepper's voice at a constant volume to explore how the alteration of non-loudness features can affect the perception of the voice. Based on the results in the table, we decided to equate the ``exciting or bright'' cluster with Pepper's Joyful voice and the ``calm'' cluster with Pepper's Neutral voice. For the ``loud'' cluster, we did not have a particularly close match given the high pitch and low pitch range. We noted that Pepper's Didactic voice had the lowest pitch range and speech rate. Therefore, we decided to build our ``Lombard'' voice with this Didactic base, increasing the pitch by 130\%.


\section{Perception Study}

We conducted a perception study to better understand human perception of our selected ambient robot voices. 
We aimed to address three research questions (RQs):\\
\textbf{RQ 1}: What voice styles are preferred for the target ambiances?\\
\textbf{RQ 2}: What ambiance-specific voice styles can increase perceived appropriateness and user comfort?\\
\textbf{RQ 3}: Does improved appropriateness and comfort occur with an increased perception of competency, awareness, and human-likeness, i.e. overall intelligence?

We recruited 120 students to participate: 71 men, 44 women, 4 non-binary, and 2 who preferred not to answer. The participants were primarily between the ages of 18 and 24 (N=114). Participants were required to have normal hearing, be over 18, and have a proficient grasp of English in order to participate. The students were primarily recruited through an introductory computing science course. All participants provided their informed consent and the procedure was approved by the SFU-REB. 

\begin{table*}[]
 \caption{Lighting (Lux) and sound (Db) settings for each condition.}
 \centering
\resizebox{\textwidth}{!}{
\begin{tabular}{ |p{1.7cm}|p{2cm}|p{1cm}|p{2cm}|p{1.7cm}|p{1.7cm}| p{1.9cm}|}
 \hline
   Ambiance   & Formal Restaurant & Café & Lively Restaurant & Quiet Bar & Noisy Bar & Night Club\\
  \hline
 Lux Level   & 44 & 17 & 17 & 44 & 44 & 0\\
\hline
 Db Level  & 50 &  50 & 55 & 50 & 55 & 60\\
 \hline
 
\end{tabular}}
 \label{lightingsound}
\end{table*}

\begin{figure}[]
      \centering
      \includegraphics[width=\linewidth]{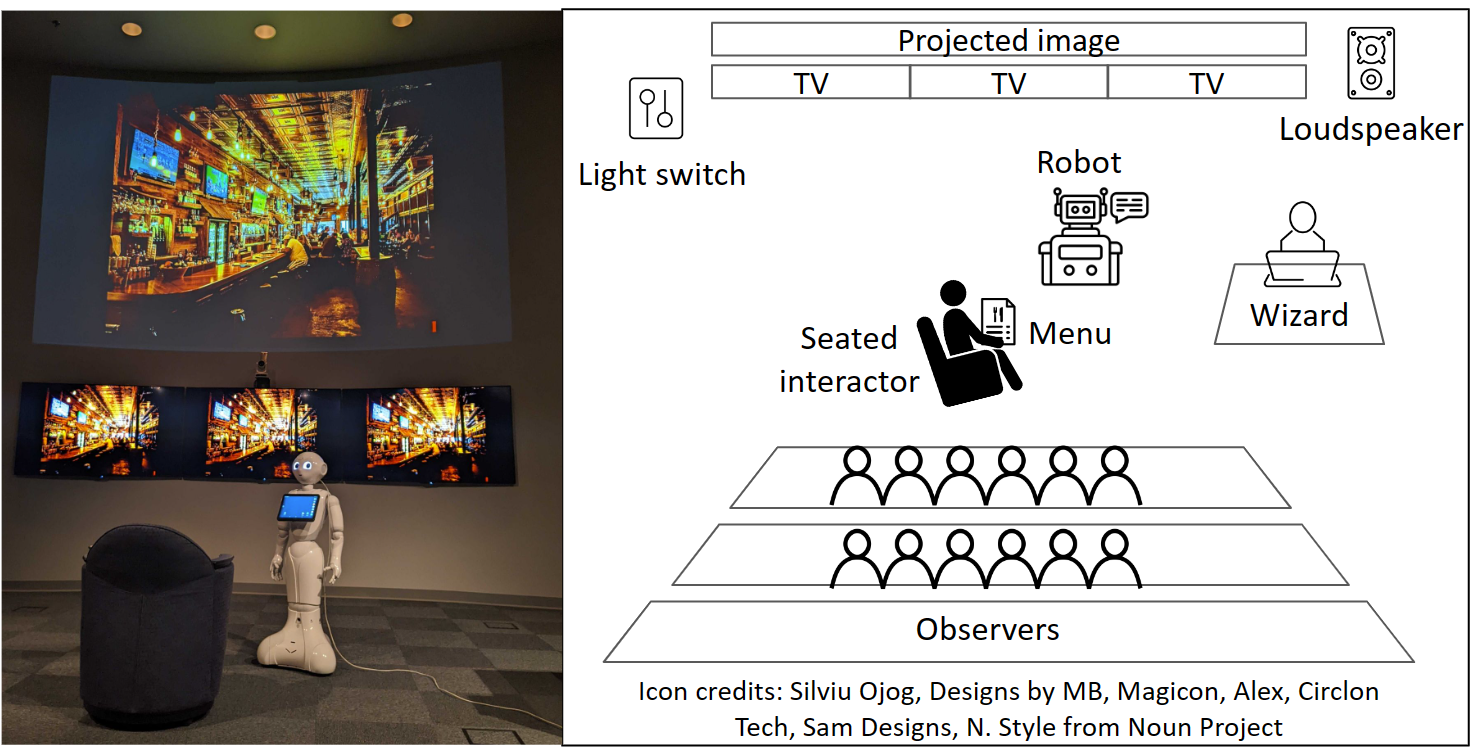}
      \caption{Experiment setup for user perception study, and the ``noisy bar'' condition (left).}
      \label{diagram}
\end{figure}

We ran our study in a presentation theatre within a university campus as a controlled proxy for actual restaurants. In order to replicate our desired ambiance, we projected the same restaurant image used to collect voice data on the wall, as well as three TV screens at the front of the hall. Additionally, we used the ambient speakers available within the room to play the music used in the data collection. Fine dining, café, and quiet bar were set to 50 dB, lively restaurant and noisy bar were set to 55 dB, and nightclub was set to 60 dB using a decibel meter. Lastly, we controlled the lighting by keeping the blinds closed and turning on and off lights to replicate `dark,' `warm and dim,' or `bright' environments. The level of light and volume of the sound used for each ambiance can be found in Table \ref{lightingsound} and an image of the setup in Fig. \ref{diagram}. 
The ambiances were presented in random order for each trial. 

The participants interacted one at a time, for a single interaction with the Pepper robot, while the others observed. For each ambiance, three interactions took place using the script below, each with a different voice style. Pepper was controlled using Wizard of Oz, with Pepper's ALTextToSpeech module. Pepper said, \emph{``Hello, I hope you are doing well. I hope you have had a chance to look at the menu. I recommend the daily special. Anyways, what can I get you?''} The interacting participant (hereafter, ``interactor'') then verbally replied, choosing an item from the paper menu in front of them. Pepper then replied, \emph{``Sure, I can definitely do that. I will be right back with your order.''} During the conversations, Pepper's tablet displayed a black screen, and no gestures were used. Basic Awareness and idle body motions were activated to allow the robot to appear lifelike while also allowing the participants to focus on the voice.

Once the conversation was completed, all participants, both the interactor and the observers, completed an online questionnaire. A black screen was shown during this time, and no music was played. The questions began by asking whether you were the one who interacted with the robot. The remaining questions used a 7-point Likert scale (1: strongly disagree, 7: strongly agree) as follows:

\begin{itemize}
 \footnotesize
\item Pepper's voice is socially appropriate for the scene
\item Pepper is aware of the surrounding ambiance
\item Pepper makes me feel comfortable
\item Pepper sounds human-like
\item Pepper is a competent server
\end{itemize}

To the best of the authors' knowledge, no validated measures exist clearly targeting all of these factors, specifically social appropriateness within a scene and ambiance awareness, especially without the inclusion of numerous extraneous questions. At the end of each ambiance, the participants answered a ``scene change'' question set while the next ambiance was set up. These questions included which of the three voices they would prefer as their server and questions to assess the accuracy of the ambiance. First, we asked whether ``The ambiance matched the prompt'' with a 100-point slider from disagree to agree. The subsequent questions were on a sliding scale from 1 to 100 with the prompt ``The ambiance is:'' with slider scales of Dark-Bright, Quiet-Noisy, Casual-Formal, Calm-Exciting, Cold-Warm, Negative-Positive, Enclosed area-Open area. We completed 18 of the previously described short conversations for each trial—the three voices from Sec. \ref{styles}, Neutral, Joyful, and ``Lombard'' were used in each of the 6 different ambiances: a fine dining restaurant, a café, a lively restaurant, a quiet bar, a loud bar, and a nightclub. 

\section{Results}
We first tested all results for significant differences between the ratings of the observer and the interactor, and none were found; therefore, moving forward, observers and interactors will be treated as members of the same group.
\subsection{Ambiance analysis}
\begin{figure}[t]
      \centering
      \includegraphics[width=0.7\linewidth]{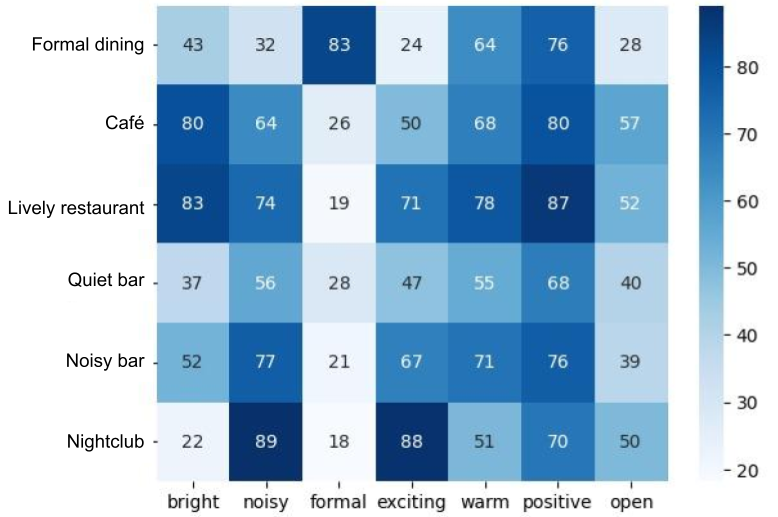}
      \caption{Perception of ambient characteristics (rated from 1-100) for each condition based on participant results.}
      \label{heatmap2}
\end{figure}

We found that, in general, our physical ambiances were a good match to the given prompt (N=714, $\mu$=81.25, $\sigma$=20.10) when rated on a scale from 1 to 100. The agreement was lowest for the quiet bar (N=120, $\mu$=77.34, $\sigma$=23.56) and highest for the fine dining restaurant (N=120, $\mu$=85.63, $\sigma$=17.86). The average ratings for each ambiance feature on a scale from 1-100 can be seen in Fig. \ref{heatmap2}.

\subsection{RQ 1: Preferred voices for ambiances}
\begin{figure}[t]
      \centering
      \includegraphics[width=0.7\linewidth]{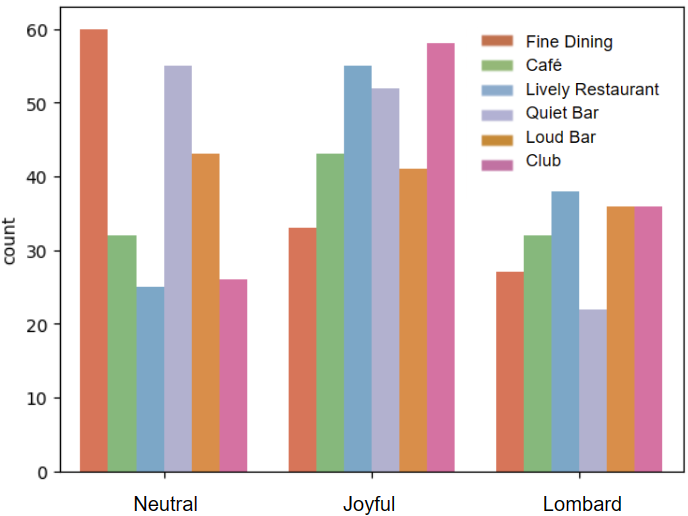}
      \caption{Number of participants choosing each voice as first preference for each ambiance.}
      \label{1st}
\end{figure}

We completed Chi Squared Goodness of Fit tests, followed by confidence intervals (CIs) to assess preference for specific voices in a given ambiance. At $\alpha$ = 0.001, we found that a Neutral voice was least often selected as the preferred server in a lively restaurant and a nightclub and that the ``Lombard'' voice was least often selected as the preferred server in the fine dining and quiet bar. Moreover, the Joyful voice was least often selected as the last choice in the lively restaurant. So, although Joyful showed no significance as the preferred voice, we know that it is not the least preferred. Full significant results can be seen in Table \ref{chi}, and a plot of first-choice voices can be seen in Fig. \ref{1st}.

\begin{table*}[t]
 \centering
 \caption{Chi-Square preference results}

 \label{chi}
 \footnotesize
\resizebox{\textwidth}{!}{
\begin{tabular}{ |p{2cm}|p{1.3cm}|p{1.3cm}|p{1.3cm}|p{1.3cm}| p{1.3cm}| p{1.3cm}| }
 \hline
\multirow{2}{*}{Ambiance} &
  \multicolumn{3}{c|}{Count} &
  \multirow{2}{*}{$\chi^{2\;2}$} &
  \multirow{2}{*}{P-value} &
  \multirow{2}{*}{CI$^3$}\\
  
   & Neutral & Joyful & 'Lomb.' &  & &    \\
\hline
 \multicolumn{7}{|c|}{\MakeUppercase{First Choice}} \\
 \hline
 Fine Dining   & 60 & 33 &  \textbf{27} & 15.45 & 0.001** &  13-32\% \\
 Lively Res.   & \textbf{25} & 55 & 38 & 11.51 & 0.003** & 12-31\% \\
 Quiet Bar  & 52 & 49 & \textbf{19} & 15.49 & 0.001** & 5-22\% \\
 Nightclub  & \textbf{26} & 58 & 36 & 15.49 & 0.001** & 12-31\%\\
 \hline
  \multicolumn{7}{|c|}{\MakeUppercase{Last Choice}} \\
 \hline
  Lively Res.   & 58 & \textbf{25} & 35 & 14.65 & 0.001** & 12-31\% \\
 \hline
 
\end{tabular}}
{\parbox{6in}{
\footnotesize 
1. \textbf{Bold} values highlight the voices selected at a significantly different ratio.\\
2. DF = 2, N = 120 (118 for Lively Restaurant)\\
3. Confidence interval on selection proportion is reported only for the selected voice at a significantly different ratio.\\
4. Only significant values are reported.
}
}
\end{table*}

\subsection{RQ 2: Social appropriateness and comfort}
For both RQ1 and RQ2, we used a one-way ANOVA followed by a post-hoc Tukey HSD and set $\alpha$=0.001 to assess improvements in Likert ratings between voices within each ambiance for our targeted questions. A full table of these results can be found in Table \ref{anova}.

We found that a Neutral voice was more socially appropriate for fine dining and that a Joyful voice was more socially appropriate in a lively restaurant, noisy bar, and nightclub. In addition, a Joyful voice made participants feel more comfortable in a nightclub.

\begin{table*}[t]
 \caption{Results for difference of mean Statistical tests}
 \label{anova}
\resizebox{\textwidth}{!}{
\begin{tabular}{ |l|l|l|l|l|l|l|l|l|l|l|l| }
 \hline
\multirow{2}{*}{Ambiance} &
  \multicolumn{2}{c|}{Neutral} &
  \multicolumn{2}{|c|}{Joyful} &
  \multicolumn{2}{|c|}{'Lombard'} &
  \multirow{2}{*}{DF} &
  \multirow{2}{*}{N} &
  \multirow{2}{*}{F} &
  \multicolumn{2}{c|}{P-value}\\
  
   & $\mu$ & $\sigma$ & $\mu$ & $\sigma$ & $\mu$ & $\sigma$ & & & & One-Way Anova & Tukey HSD\\
    \hline
 \multicolumn{12}{|c|}{\MakeUppercase{Socially Appropriate}} \\
 \hline
 Fine Dining   & \textbf{5.30} & \textbf{1.49} & 4.30 & 1.73 &  3.97 & 1.66 & 2 & 120 & 22.58 & $<0.001$*** & $<0.001$***\\
 Lively Restaurant   &  4.48 & 1.63 & \textbf{5.68} & \textbf{1.14} & 4.82 & 1.59 & 2 & 118 & 18.93 & $<0.001$*** & $<0.001$***\\
 Noisy Bar   & 4.41 & 1.54 & \textbf{5.29} & \textbf{1.17} & 4.61 & 1.47 & 2 & 120 & 20.07 & $<0.001$*** & $<0.001$***\\
 Nightclub   & 3.88 & 1.54 & \textbf{5.46} & \textbf{1.26} & 3.28 & 1.87 & 2 & 120 & 38.28 & $<0.001$*** & $<0.001$***\\
 \hline
 \multicolumn{12}{|c|}{\MakeUppercase{Ambiance Aware}} \\
 \hline
  Fine Dining   &  \textbf{5.01} & \textbf{1.54} & 4.26 & 1.65 & 4.22 & 1.63 & 2 & 120 & 9.43 & $<0.001$*** & $<0.001$***\\
  Lively Restaurant   &  4.31 & 1.42 & \textbf{5.29} & \textbf{1.26} & 4.58 & 1.58 & 2 & 118 & 13.92 & $<0.001$*** & $<0.001$***\\
  Noisy Bar   & 4.41 & 1.42 & \textbf{5.29} & \textbf{1.17} & 4.64 & 1.48 & 2 & 120 & 12.46 & $<0.001$*** & $0.001$**\\
 Nightclub   & 3.73 & 1.59 & \textbf{5.22} & \textbf{1.27} & 3.96 & 1.54 & 2 & 120 & 35.25 & $<0.001$*** & $<0.001$***\\
 \hline
  \multicolumn{12}{|c|}{\MakeUppercase{Comfort}} \\
 \hline
  Nightclub   &  4.35 & 1.45 & \textbf{5.09} & \textbf{1.40} & 4.41 & 1.56 & 2 & 120 & 9.43 & $<0.001$*** & $<0.001$***\\
 \hline
   \multicolumn{12}{|c|}{\MakeUppercase{Human-likeness}} \\
 \hline
  Nightclub   & 3.66 & 1.57 & \textbf{4.37} & \textbf{1.59} & 3.65 & 1.60 & 2 & 120 & 8.05 & $<0.001$*** & $0.002$**\\
 \hline
  \multicolumn{12}{|c|}{\MakeUppercase{Competency}} \\
 \hline
  Noisy Bar   &  5.32 & 1.43 & \textbf{5.87} & \textbf{1.35}  & 5.36 & 1.45 & 2 & 120 & 7.91 & $<0.001$*** & $0.002$**\\
 Nightclub   &  5.13 & 1.37 & \textbf{5.80} & \textbf{1.06}  & 5.08 & 1.43 & 2 & 120 & 11.46 & $<0.001$*** & $<0.001$***\\
 \hline
 
\end{tabular}}
{\parbox{6in}{
\footnotesize 
1. \textbf{Bold} values highlight the preferred voice.\\
2. Due to space limitations, we only report those voices that were found to have a significantly different Likert rating than both other voices. In all cases, the post-hoc P-value was equal for all pairings with the preferred voice.
}
}
\end{table*}

\subsection{RQ 3: Awareness, competency, and human-likeness}
We found that a Neutral voice was perceived as more ambiance-aware for fine dining and that a Joyful voice was perceived as more ambiance-aware in a lively restaurant, noisy bar, and nightclub. In addition, the robot with a Joyful voice was perceived as more human-like and competent in a nightclub and more competent in a noisy bar.

\begin{figure}[]
      \centering

      \includegraphics[width=0.8\linewidth]{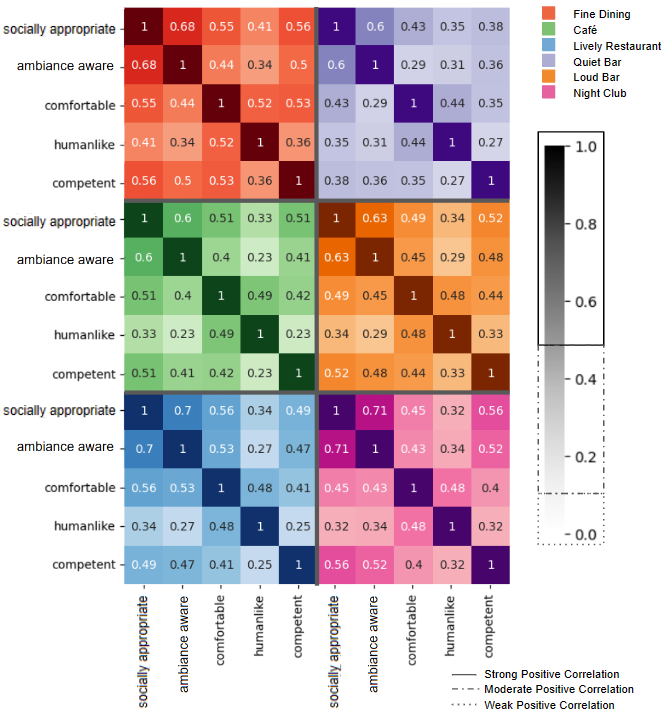}
      \caption{Pearson's R correlations between perception study responses for each ambiance.}
      \label{corr}
\end{figure}

Correlation results can be seen in Fig. \ref{corr}. We find the strongest positive correlations between social appropriateness and ambiance awareness in all ambient contexts. Strong positive correlations are seen between social appropriateness, competency, and comfort for fine dining and the café, with competency in the nightclub and loud bar and with comfort in the lively restaurant. Comfort strongly correlates with competency and human-likeness for fine dining and ambiance awareness in the lively restaurant. Lastly, ambiance awareness shows strong positive correlations with competency in the fine dining and the nightclub, as well as comfort in the lively restaurant. All other correlations are moderately positive, which overall tells us that by increasing social and ambiance appropriateness, we can improve perceptions of social and overall intelligence in all factors.

\section{Discussion}

We aimed to better understand how to increase perceived robot intelligence through selection of ambient adapted voices, specifically using non-loudness features, in opposition to previous literature.

We first developed a protocol for collecting ambiance-modified voices through Zoom to mitigate the difficulties with in-the-wild collection of vocal features in noisy ambiances. We found that we were able to replicate our desired ambiances to a reasonable degree through changes in lighting, sound, and images, as confirmed through both our validation study and the ambiance ratings in our perception study. These results are especially exciting as they suggest that simple, low-cost virtual simulations of ambient environments could be a suitable means for data collection toward usage on real robots. Further studies should be conducted to compare the results of simulated versus in-the-wild contexts.

We proposed clustering the voice features to extract primary voice styles, as opposed to a brute-force method of creating a separate voice for each ambiance. This method requires less data and computational intensity while being more transparent and interpretable. Testing robot voices selected based off these clusters, we found that the presence of a voice style in an ambiance from the Zoom study (Fig. \ref{barplot}) tended to match the preference of the matching voice style in that same ambiance in the on-robot study (Fig. \ref{1st}). This provides evidence supporting our overall procedure.

Our results indicate that the Neutral voice (default on Pepper) was significantly NOT preferred in loud, high-energy environments (Table \ref{chi}). Instead, a higher-pitched voice with high pitch range and speed was preferred. Surprisingly, given both the literature on human vocal adaptation and our human voice analysis results, we did not see our curated ``Lombard'' voice outperforming Pepper's base Joyful voice in these loud environments. However, the designed ``Lombard'' voice tended to have higher comfort or social appropriateness ratings in the loud environments (loud bar, lively restaurant, nightclub) compared to the Neutral voice. One participant commented that the ``Lombard'' voice was ``Robotic sounding but louder'' in the nightclub setting. Since the robot volumes remained constant, this could suggest that the ``Lombard'' voice with increased pitch could indeed increase perceived loudness. Finally, it should be noted that our in-person study did \emph{not} include any negative sentiment sentences. It remains to be tested whether the ``Lombard'' voice could outperform the Joyful Pepper voice in these contexts (e.g., ``Sorry, we don't have that today.''). As expected, the Neutral voice was found to be more socially appropriate and ambiance-aware in the quiet fine dining ambiance.

The lack of preference for the ``Lombard'' voice may have several explanations: first, the Lombard voice is produced by humans in loud environments to increase intelligibility. However, further research needs to be done to see if equally loud and intelligible voices are actually preferred by the listener or are just an inevitable production of vocal force from the speaker. Participants commented that with the Lombard voice in the nightclub environment, ``the interaction was like with a real server'', but ``could be more casual''. 
Another possible explanation is the lack of flexibility in text-to-speech models. Perhaps only being able to modify speech rate and pitch with a slight selection of pitch range is not enough to accurately re-create Lombard speech. Spectral features, specifically spectral slope, which was found to be significantly different in loud environments, may play a key role in the acceptability of pressed Lombard voice. 
Finally, by inspecting Fig.~\ref{barplot}, the preference for the joyful voice over Lombard voice may, in fact, be expected. In our Zoom study, the joyful voice style was used more often in all ambiances compared to the Lombard voice style. However, utterance-specific appropriateness may also be considered. For our perception study, all utterances were positive or neutral. It seems likely that a Lombard voice may outperform a joyful voice in loud ambiances if the utterance had a negative sentiment (e.g., ``Sorry, we don't have that today.'') and should be considered future work.


 Overall, other than the café and quiet bar, including those ambiances where there was no significance in the selection of preferred server, the preferred voice saw a significant increase in social appropriateness and ambiance awareness. This suggests that carefully selecting the correct voice for an ambiance can improve human perception of socially intelligent robots. Additionally, Fig. \ref{corr} depicts strong correlations between an increase in socially appropriate voice and competency.

It is important to note the limitations of the aforementioned study. First, although we collected clear audio data for a variety of different virtual ambient environments, we could not control for how one would adapt their voice virtually versus in a real ambient environment. Second, we could not control for group dynamics in the user study. More specifically, how one would interact, perceive, or interpret Pepper in isolation may differ from when not in isolation. Finally, we were limited to Pepper's vocal features, and thus, our work was tailored to that of Pepper. Introducing a variety of social robots in the future would allow us to observe if embodiment impacts the results. 

\chapter{A Text-to-Speech Model for Second-Language Listeners}

We finally arrived at the question of how to adapt a TTS to improve comprehension for second-language speakers. The perception and production of certain L2 vowels, even those that exist in the speaker's first language, remains challenging for L2 speakers even at a high level of proficiency \cite{kewley2005influence}. In English, differentiating between tense (long) and lax (short) vowels (e.g., beat vs. bit) can be particularly difficult for second-language speakers \cite{iverson12, han2013pronunciation, 2019japanese}. 

Often, first-language speakers consider that slowing down their speech rate or putting emphasis on a difficult sound by raising their pitch will help improve L2 comprehension \cite{AOKI2024101328}; however if vowel length or pitch is a primary marker for vowel differentiation, is this truly the ideal method? 

As already mentioned in Chapter 2, while in the past vowel perception had been viewed as a local process \cite{kuhl97}, it has now become well-understood that the pitch and duration context surrounding vowels plays an important role in their perception \cite{Stilp20}. The existence of such context effects opens the opportunity to use pitch and duration manipulations within and around the target sound strategically to bias the perception of difficult L2 sounds into their correct interpretation. Unfortunately, we lack a precise specification of how one should modify these cues to facilitate vowel comprehension. 

To understand the contribution of pitch and duration of both a target word containing a tense or lax vowel and its proceeding context, in \cite{tuttosi24_interspeech}, we applied the data-driven psychophysical methodology of \emph{reverse correlation}. We follow this up with a validation of these findings, making manual manipulations to a TTS. Finally, we built a TTS that automatically incorporates modifications to tense-lax vowels specifically designed to aid L2 speakers in their comprehension of these difficult vowels.

\section{Reverse correlating context effects on L1 and L2 vowel perception}

\label{sec:revcor}

To explore the effect of pitch and duration on vowel perception, we used a phase-vocoder technique \cite{dolson1986phase} to systematically vary the pitch and speech rate in phrases surrounding pairs of vowels/words that are difficult for L2 speakers: English (\textipa{/i/}-\textipa{/I/}) and French (\textipa{/u/}-\textipa{/y/}), letting participants listen and judge which word they heard, and then used reverse correlation to reconstruct the prosodic profiles that biased the perception of these word pairs in one direction or the other. We did this both for isolated words (Word task) and for words embedded in sentences (Phrase task).

We first collected data from English-L1-French-L2 and French-L1-English-L2 speakers on the same French and English stimuli. This initial pairing was chosen as duration plays a vital role in English vowels, yet plays little role in distinguishing vowels in Parisian French \cite{MonnotMichel1974LPdf, StrangeWinifred2007Avwa, MillerJoanneL.2011DEiS}. Additionally, while pitched stress plays a vital role in English \cite{ulHassanSardarFayyaz2012Tnos}, stress in the French language is still not clearly defined and appears to be more rhythmic in nature \cite{frost09}.

\begin{CJK*}{UTF8}{gbsn}

We then extended the study to Mandarin-L1-English-L2 and Japanese-L1-English-L2 speakers with the same English stimuli as the previous groups. As French has both lesser influences of pitch and duration on the comprehension of vowels than in English, Mandarin, and Japanese were chosen to explore these aspects in languages that rely more extensively on these features for perception. Mandarin lacks vowel length contrasts, and it has been found that Mandarin speakers display reduced precision in determining vowel length in other languages \cite{LU2021101049}. Instead, Mandarin vowels are firmly defined by four discrete tones completely defined by changes in pitch over the vowel, where duration remains only a secondary cue (e.g., 汤 tāng (soup) and 糖 táng (sugar)). The addition of Japanese extends to speakers whose first language relies heavily on duration to differentiate vowels. All 5 vowels in Japanese contain a minimal pair of short/long vowels, which, when varied, can change the meaning of a word \cite{HIRATA2004565}(e.g.  かど \textipa{/kado/} (corner) カード\textipa{/ka:do/} (card). Moreover, Japanese has a pitch accent distinct from, but not dissimilar to, English \cite{kozasa2004interaction}.

\end{CJK*}

\subsection{Stimulus generation}

\subsubsection{Vowel selection and word}

In English, we selected the pair of tense/lax vowels \textipa{/i/} and \textipa{/I/} that are known to be difficult for English L2 speakers both in perception and in production \cite{iverson12}. From the point of view of French-L1 speakers, the vowel \textipa{/i/}, such as in ``beat'' \textipa{/bit/}, is a high, tense vowel and exists in both French and English. In contrast, the vowel \textipa{/I/}, such as in ``bit'' \textipa{/bIt/}, is lower and lax and does not exist within the French lexicon. Often, French-L1 speakers learning English will replace \textipa{/I/} by \textipa{/i/}, for example saying ``sheep'' \textipa{[Sip]} when they mean to say ``ship'' \textipa{/SIp/} \cite{iverson12}. Similarly, \cite{2019japanese} found that Japanese speakers will form the same vowel substitution, i.e., substitute the tense \textipa{/i/} for the lax \textipa{/I/}. In Japanese, as in English and French, \textipa{/i/} is a high front vowel, and, as in French, \textipa{/I/} is non-existent. Mandarin-L1 speakers of English face a similar problem. While Mandarin contains the vowel \textipa{/I/}, Mandarin speakers often lack the ability to distinguish between \textipa{/I/} and \textipa{/i/} as, in Mandarin, the length difference between these two vowels does not alter their meaning. In many cases, Mandarin speakers will replace both vowels in ``sheep'' \textipa{[Sip]} and ``ship'' \textipa{/SIp/} with the Mandarin vowel \textipa{/I/}, which requires a higher and more frontal position of the tongue \cite{han2013pronunciation}. We selected target words that differ only by these vowels to maintain consistency in confounding factors from consonant proximity. The selected words were ``pill'' \textipa{/pIl/} and ``peel'' \textipa{/pil/}.

For French, we selected the vowels \textipa{/u/} and \textipa{/y/}. Once again, this pair of vowels is known to be difficult for English-L1-French-L2 speakers in both perception and production \cite{Flege87, Levy09}. The vowel \textipa{/u/}, such as in ``fou'' \textipa{/fu/} (mad), is a high, back vowel and exists in both French and English. In contrast, the vowel \textipa{/y/}, such as in ``fût'' \textipa{/fy/} (cask), is a high, front vowel and does not exist within the English lexicon. English speakers tend to overcompensate for the lack of \textipa{/y/} vowel in their language production and often replace \textipa{/u/} with \textipa{/y/}, such as mispronouncing ``beaucoup'' \textipa{/boku/} (a lot) with the rather immodest phrase \textipa{[boky]} \cite{STU13}. Once again, we selected words that differ only by these vowels to maintain consistency in confounding factors from consonant proximity. The selected words were ``pull'' \textipa{/pyl/} and ``poule'' \textipa{/pul/}.

\subsubsection{Phrase stimuli} 

To embed target words in a larger context, we used the phrase ``I heard them say'' (FR: ``je l'ai entendu dire'') preceding the word. This was selected in an attempt to avoid that a semantic context could bias the interpretation of the target word in the direction, e.g., of the most frequent alternative. The phrases were generated using the Hugging Face interface for CoquiXTTS\footnote{https://huggingface.co/spaces/coqui/xtts}. The language was set to English for the English phrase and French for the French phrase. An L1 male reference voice was provided, and no other modifications were made to the TTS settings.

\subsubsection{Stimulus manipulation: vowel ambiguity}

To avoid response bias, reverse correlation experiments most often employ a 2-interval one alternative methodology, for example, asking the participant: ``which of the two stimuli is more like \emph{a target}''. However, for sound-based experiments, the experiment length nearly doubles with the addition of a second stimulus. Comparatively, a 1-interval task (``which of these 2 words did you hear'') allows for nearly twice as many trials in the same amount of time. Yet, because reverse correlation presents the participant with random manipulations of an original word/phrase in the hope that noise will sway their decision one way or another, the 1-interval design presents the risk that the modifications made to the phrase will not be enough to sway the participants between responses; if not, there will be a strong bias towards the original word, limiting the number of trials that can be analyzed. Another problem with 1-interval tasks is that we would need two sets of phrases, one for each word. 

To mitigate this bias in our 1-interval, 2-alternative task method, we generated morphed sounds that were perceptually intermediate between each of the vowel pair: \textipa{/i/}-\textipa{/I/} and \textipa{/u/}-\textipa{/y/}. To do this, we extracted the vowels from each of the 4 target words, making sure to split and merge on zero-crossing points, and estimated the frequencies of the first two formants, F1 and F2, for each vowel. We then used the Whisper automatic speech recognition (ASR) \cite{whisper2023}\footnote{v20231117, medium multilingual} to estimate the most probable interpretation at each of a series of steps interpolated between the formants of each alternative and selected the point that Whisper judged most ambiguous. 

\begin{figure}[t]
  \centering
  \includegraphics[width=0.7\linewidth]{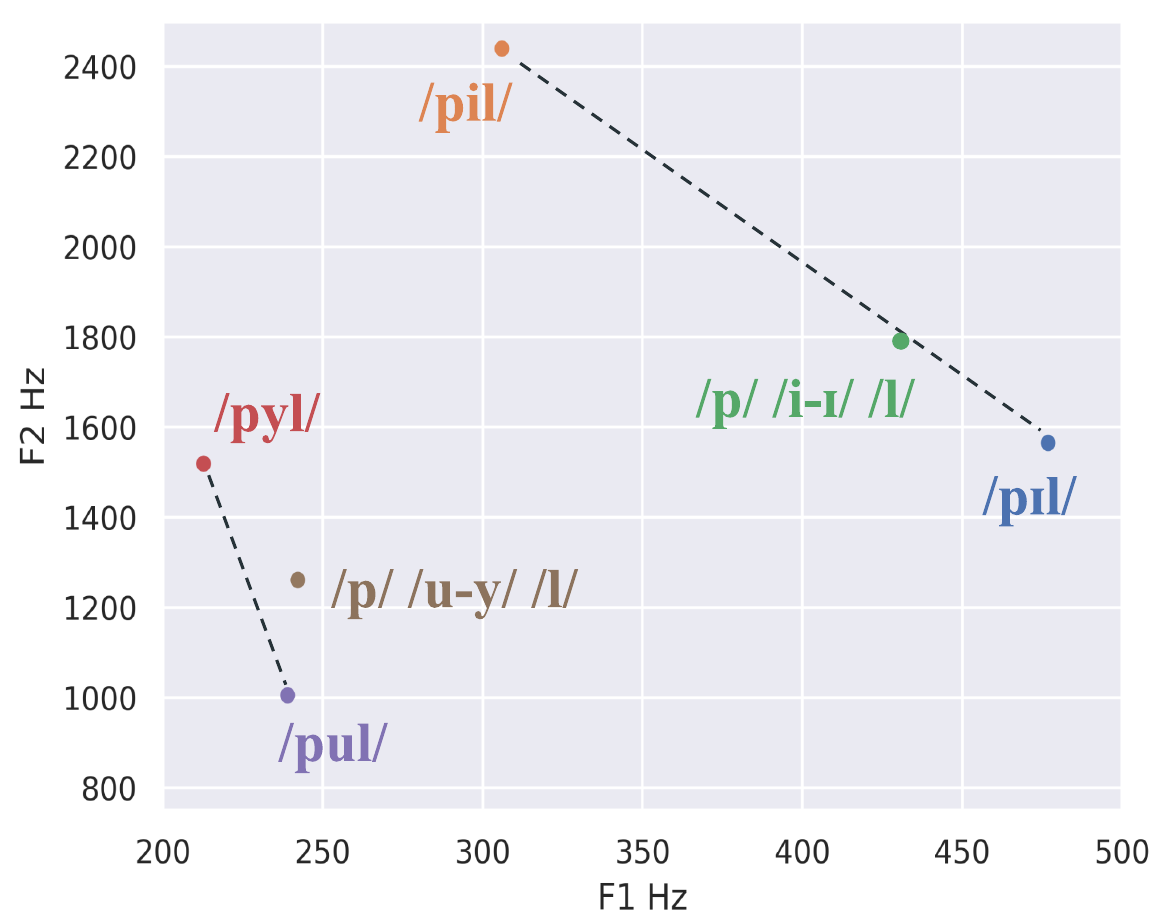}
  \caption{F1-F2 plot of initial and final formants for ambiguous vowels /u-y/ and /i-I/.}
  \label{fig:formants}
\vspace*{-5mm}
\end{figure}

In more detail, we modified Whisper to extract the log-probabilities that a processed audio clip contains each of the target words for a given language rather than simply giving the most likely prediction. A two dimensional search grid was generated using the origin F1 and F2 as follows: \textipa{/i/} F1: 305.89Hz, F2: 2440.77Hz to \textipa{/I/} F1: 476.85Hz, F2: 1565.45Hz, stepsize = 10Hz; \textipa{/u/} F1: 238.90Hz, F2: 1005.67Hz to \textipa{/y/} F1: 212.61Hz, F2: 1519.49Hz. At each iteration, Praat \cite{praat} was used to extract and modify the formants, then re-synthesize the word to be processed by the ASR model. This was completed starting from both vowels, i.e. in English moving both from \textipa{/pil/} (peel) to \textipa{/pIl/} (pill) and from \textipa{/pIl/} (pill) to \textipa{/pil/} (peel), searching for the formant pairing with the smallest difference in log-probabilities for the two target words (i.e. most ambiguous). The resulting formants can be seen in Fig.~\ref{fig:formants} with the final \textipa{/i/}-\textipa{/I/} vowel having F1: 436.77Hz, F2: 1722.68Hz (synthesized from ``peel'' \textipa{/pil/}), and the final \textipa{/u/}-\textipa{/y/} vowel having F1: 238.13Hz, F2: 1258.68Hz (synthesized from ``poule'' \textipa{/pul/}). These ambiguous words served as base stimuli in the Word task (see Experimental Procedure, below) and were then inserted manually in their original phrase (at a zero-crossing 120ms after the end of the last word ``say/dire'') to serve as base stimuli for the Phrase task.

\subsubsection{Stimulus manipulation: randomization}

Next, we generated the reverse-correlation stimuli from these ambiguous base sounds using the open-source CLEESE toolbox \cite{BURR19}, a voice-transformation toolbox that creates random fluctuations around an audio file’s original contour of pitch and speech rate. The pitch contour of the recordings was artificially flattened to a constant 120Hz. We then transformed the stimuli by randomly manipulating their pitch and duration independently, beginning with pitch, then duration, in \emph{n} successive windows of 100ms (\emph{n}=4 for words; \emph{n}=13 for phrases), each of which with a factor sampled from a normal distribution (pitch: $\mu$=0, $\sigma$=100 cents (i.e. 1 semitone); duration: $\mu$=0\%$,\sigma$=100\% (i.e. doubling or halving the window's duration); both distributions clipped at $\pm 2 \sigma$). These values were chosen so as to cover the range observed in naturally produced utterances and were linearly interpolated between successive time points to ensure a natural-sounding transformation.

\subsection{Experimental procedure}

Reverse correlation is an experimental paradigm aimed at discovering the signal features that govern a participant's judgment by analyzing their responses to large sets of stimuli whose acoustic characteristics have been systematically manipulated \cite{ADOL16}. Specifically, we presented participants with a series of 250 trials, each consisting of a single base recording containing one of our ambiguous target words. For each trial, the base recording was manipulated as previously described with a different random profile of pitch and speech rate, and participants were asked which of two target words they heard (1-interval, 2-alternative forced choice). Responses were then analyzed to reconstruct the prosodic profile that maximizes the likelihood of responding to one option or the other (for a review, see \cite{Mur11}). 

This experiment included 4 different conditions, randomized between participants: (A) two involving random manipulations of the isolated target word (English or French), aimed at establishing a baseline for the intrinsic pitch and rate of the two alternatives, and (B) two involving manipulations of phrases containing the target word, aimed at uncovering extrinsic acoustic context effects. Participants recruited online had the option to participate in more than one condition, in which case the order was randomized across language and type of stimuli (word and phrase). 

In each condition and before the reverse correlation task, we collected information about the participant's native language, age, gender, and self-rated English proficiency. Participants then successively listened to all trials (each played only once) and responded which of the target words they heard. The order of response options (e.g., ``pill/peel'' or ``peel/pill'') was randomized across trials and participants. Each condition lasted, on average, 15 minutes.

\subsubsection{Participants} 

N=160 participants took part in the study: N=54 French-L1 speakers (female: 24, M=32.4yo ± 9.7), N=60 English-L1 speakers (female: 33, M=30.0yo ± 10.7), N=30 Mandarin-L1 speakers (female: 15, M=22.2yo ± 5.6), and N=16 Japanese-L1 speakers (female: 5, M=29.9yo ± 12.1). The participants were recruited via the online Prolific platform.

L2 participants had a self-rated English proficiency ranging from 2-5 (1: no proficiency, 5: fluent) with a mode of 4. English participants were recruited primarily from anglophone Canada, French participants from France, Mandarin within China, and Japanese from Japan. As such, the L2 speakers are immersed in their native language rather than their L2.

English and French participants were recruited for all English/French Word/Phrase tasks (4 conditions). The Japanese and Mandarin participants were only recruited for the English vowel tasks (2 conditions: Word and Phrase). Each condition (1-word tasks, 1 phrase tasks) was carried out with n=25 speakers for each L1 aside from Japanese. For Japanese participants, we had n=14 for the word task and n=13 for the phrase task due to difficulties finding Japanese participants. Although conditions were randomized across participants, participants were also left the option to participate in more than one condition - in the following, we treat all samples as independent regardless of repeated measures. 

All participants provided their informed consent and were compensated financially for their time at a standard rate. The procedure was approved by the SFU-REB. 

\subsection{Results}

\subsubsection{Validation of ambiguity}
We first explored how successful we were at creating an ambiguous vowel to confirm the validity of the  PRAAT/Whisper approach. Because random manipulations were centered on zero, we expected a 50\% response rate for the alternative options. For the \emph{English Word} condition, English-L1 speakers answered ``peel'' $52\%$ of the time, French-L1 speakers $54\%$ of the time, Mandarin-L1 speakers $53\%$ of the time, and Japanese $52\%$. For the \emph{English Phrase} condition, English-L1 speakers responded with $59\%$ ``peel'', the French-L1 speakers with $54\%$, Mandarin-L1 speakers with $54\%$ and Japanese-L1 $56\%$. For the  \emph{French Word} condition, English-L1 speakers responded ``poule'' $64\%$ of the time and French-L1 speakers $70\%$ of the time. For the \emph{French Phrase} condition, English-L1 speakers responded with $59\%$ ``poule'', while French-L1 speakers' responses were $66.6\%$ ``poule''. 

In sum, all stimuli appeared sufficiently ambiguous, although to a lesser degree for native speakers and for French. Informal discussions with participants concurred with these results. 

\subsubsection{French reverse correlation}
The first subset of results concerns how L1 (French) and L2 (English) speakers perceived French-language stimuli, either as single words (\emph{``pull''} vs. \emph{``poule''}) or complete sentences (ex. \emph{``je l'ai entendu dire poule'')}. 

\noindent{\bf Analysis procedure:} We computed first-order kernels from reverse-correlation data using the {\it classification image} method \cite{Mur11} for each participant. This is done by computing the average random pitch and speech rate transformation profile of the recordings classified as one response option (e.g., ``pill'') and subtracting it from the average profile of the recordings classified as the other option (e.g., ``peel''). The kernels are then normalized by dividing them by the root-mean-square sum of their values. This resulted in two dim=4 (word) and dim=13 (phrase) vectors of pitch and rate of speech values for each participant. These represent the pitch and speech rate transformations that should be applied to a given base word to increase the likelihood of recognizing one target word or the other. To analyze the differences between the two kernels within participants, we compute paired t-tests at every time point, i.e., every 100ms.

\begin{figure}[t]
  \centering
  \includegraphics[width=\linewidth]{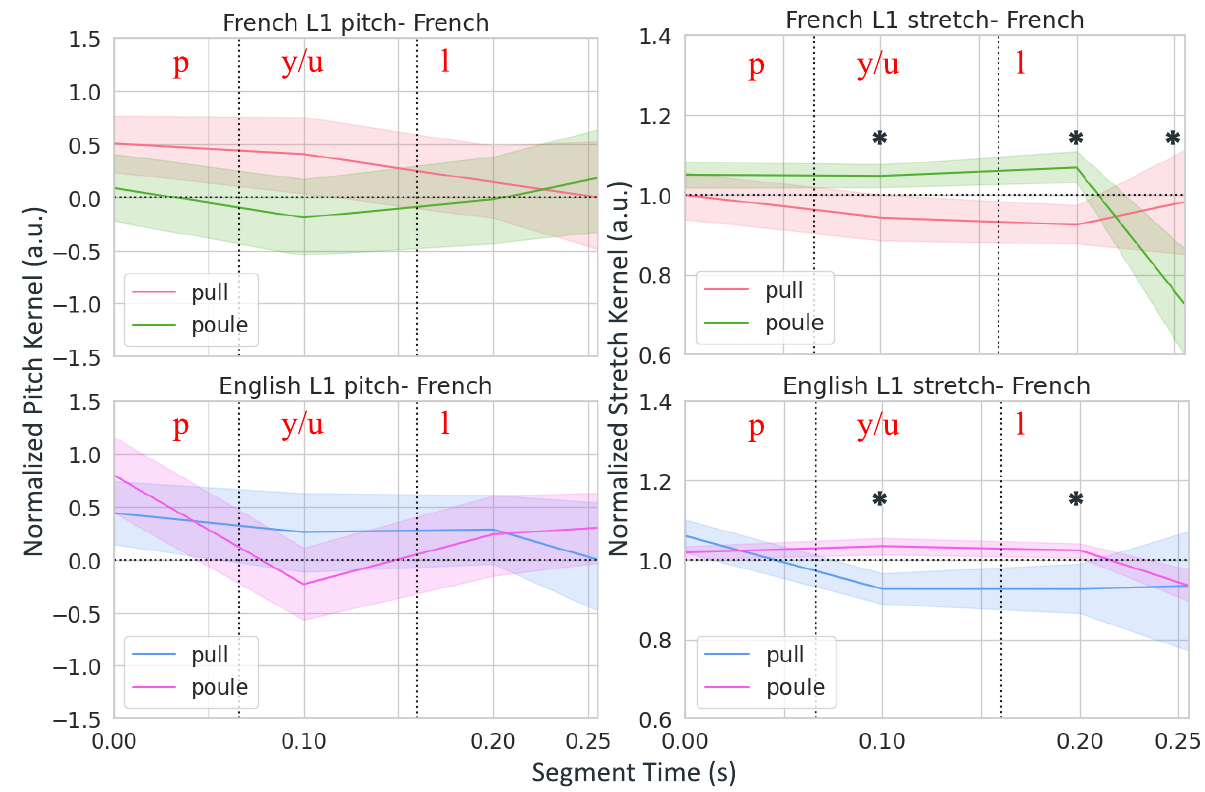}
  \caption{{\bf French words reverse-correlation results:} Pitch (left) and speech rate (right) kernels for French-L1 (top) and L2 (bottom) speakers for the French words ``pull'' and ``poule'', presented in isolation. In all figures, colored areas mark 95\% confidence intervals on the mean, and * marks time segments that differ statistically at $\alpha=0.05$. Smaller values for the stretch kernel mean shorter duration, i.e. {\it faster} speech rate, and for the pitch kernel lower pitch.}
  \label{fig:french-word}
  \vspace*{-5mm}
\end{figure}

\noindent{\bf Words: } Our prediction was that French-L1 (FL1) speakers' perception of \textipa{/y/}, a high/front vowel, would be driven by a higher pitch and faster speech rate compared to \textipa{/u/}. Our results confirm this (Fig.\ref{fig:french-word}) for speech rate (FL1 - 0.1s: t(25)=2.80, p=.010; 0.2s: t(25)=3.52, p=.002), and pitch (at 0s and 0.1s, albeit non-statistically). French-L2 (EL1) shares the same pattern of data, with a non-statistical pitch increase at t=0.1s and faster speech rate at 0.1s: t(25)=4.57, p$<$.001; and 0.2s: t(25)=2.73, p=.012. Although statistically weak, this pattern of results (higher pitch/faster rate for \textipa{/y/}) is confirmed by segments located on the target word in the phrase kernels (see below).

\begin{figure}[t]
  \centering
  \includegraphics[width=\linewidth]{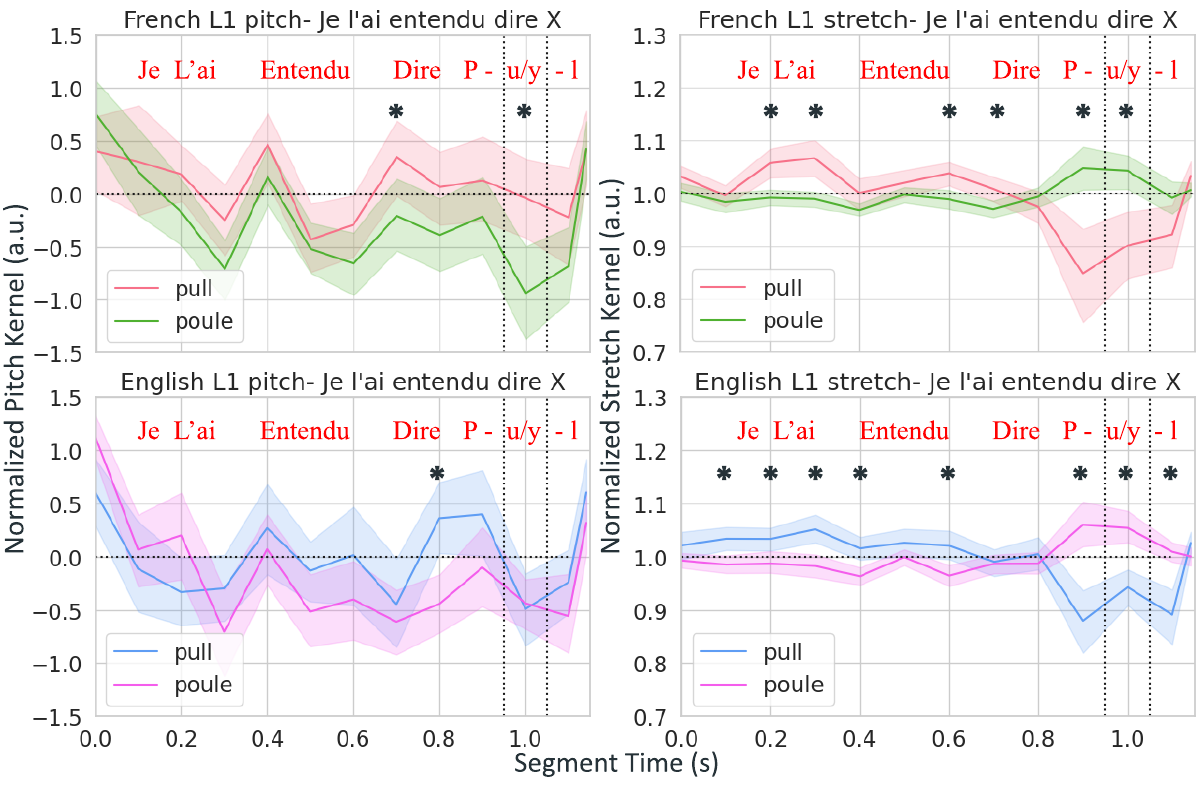}
  \caption{{\bf French phrase reverse-correlation results:} Pitch (left) and speech rate (right) kernels for French-L1 (top) and L2 (bottom) speakers for the French phrases containing ``pull'' and ``poule''. Smaller values for the stretch kernel mean shorter duration, i.e. {\it faster} speech rate, and for the pitch kernel lower pitch.}
  \label{fig:french-phrase}
  \vspace*{-5mm}
\end{figure}

\noindent{\bf Phrases: } Phrase reverse-correlation data first confirms, this time significantly, that, within the target word, hearing \textipa{/y/} is driven by higher pitch in FL1 (1s: t(25)=-2.63, p=.015) and faster speech rate in both FL1 and EL1 speakers (FL1 - 1.0s: t(25)=2.99, p=.006; EL1 - 1.0s: t(25)=4.04, p$<$.001, 1.1s: t(25)=3.71, p=.001). Second, regarding the surrounding context of the word, the literature predicts the existence of contextual contrast effects, i.e., that \textipa{/y/} is driven by {\it lower} pitch and speech rate before the target word. In actuality, our data reveals a mix of contrastive and congruent contextual influences: pitch (Fig. \ref{fig:french-phrase}-left) is not associated with any contrastive effect, but rather a proximal congruent effect 200-300ms pre-target (FL1- 0.7s: t(25)=-2.18, p=.040; EL1- 0.8s: t(25)=-4.05, p$<$.001), i.e. an increase of pitch immediately before the target word biases the response towards the higher-pitch alternative \textipa{/y/}. Speech rate (Fig. \ref{fig:french-phrase}-right) exhibits the expected long-term (distal) contrastive effect (FL1 - 0.2s: t(25)=-3.72, p=.001, 0.3s: t(25)=-3.86, p=.001, 0.6s: t(25)=-2.89, p=.008, 0.7s: t(25)=-2.24, p=.018; EL1 - 0.1s: t(25)=-3.35, p=.003, 0.2s: t(25)=-2.53, p=.018, 0.3s: t(25)=-4.12, p$<$.001, 0.4s: t(25)=-2.99, p=.006, 0.6s: t(25)=-3.46, p=.002), where a slower speech rate at the beginning of a phrase and a proximal congruent effect 100ms pre-target (FL1 - 0.9s: t(25)=3.25, p=.003; EL1 - 0.9s: t(25)=3.76, p=.001) biases the response towards the faster alternative \textipa{/y/}, resulting in an overall scissor-shape profile. This pattern of result was remarkably conserved in L2 speakers (Fig. \ref{fig:french-phrase}-bottom). 

\subsubsection{English reverse correlation}

\begin{figure}[]
  \centering
  \includegraphics[width=0.88\linewidth]{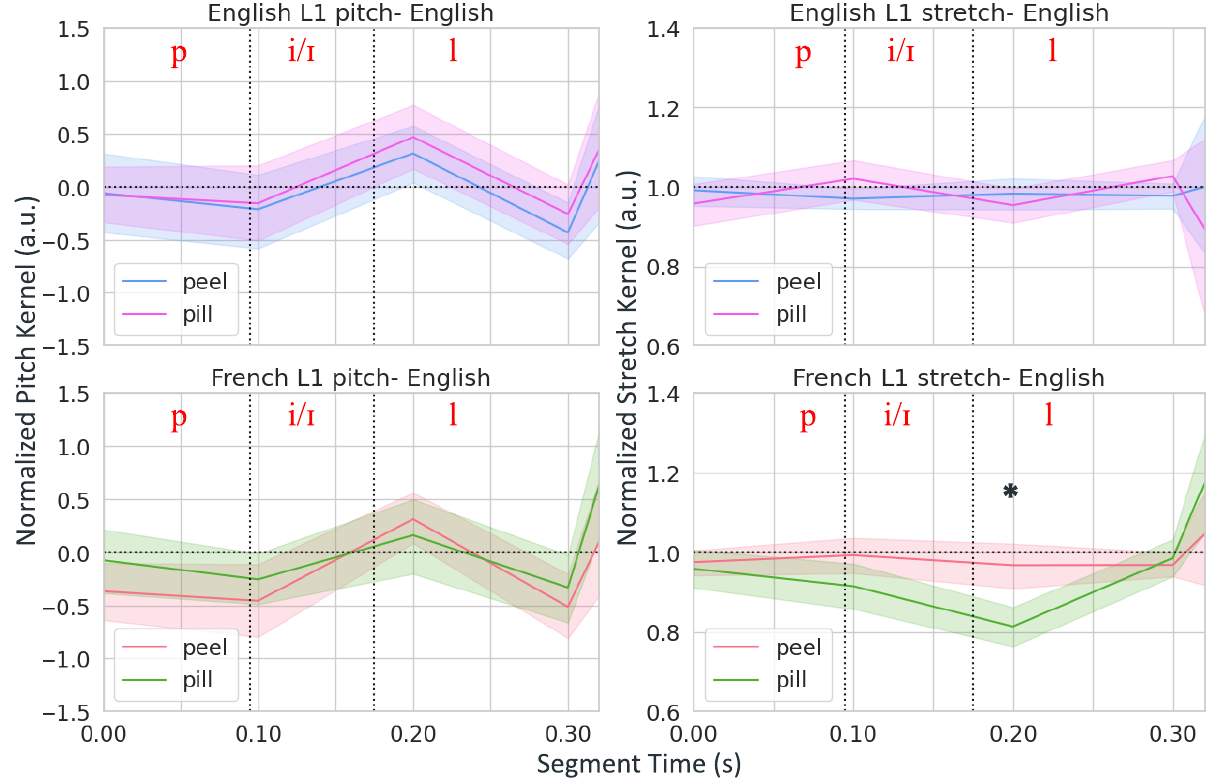}
  \caption{{\bf English words reverse-correlation results (EL1, FL1):} Pitch (left) and speech rate (right) kernels for English-L1 (top) and L2 (bottom) speakers for the English words ``peel'' and ``pill''. Smaller values for the stretch kernel mean shorter duration, i.e. {\it faster} speech rate, and for the pitch kernel lower pitch.}
  \label{fig:english-word}
\end{figure}

\begin{figure}[]
  \centering
  \includegraphics[width=0.88\linewidth]{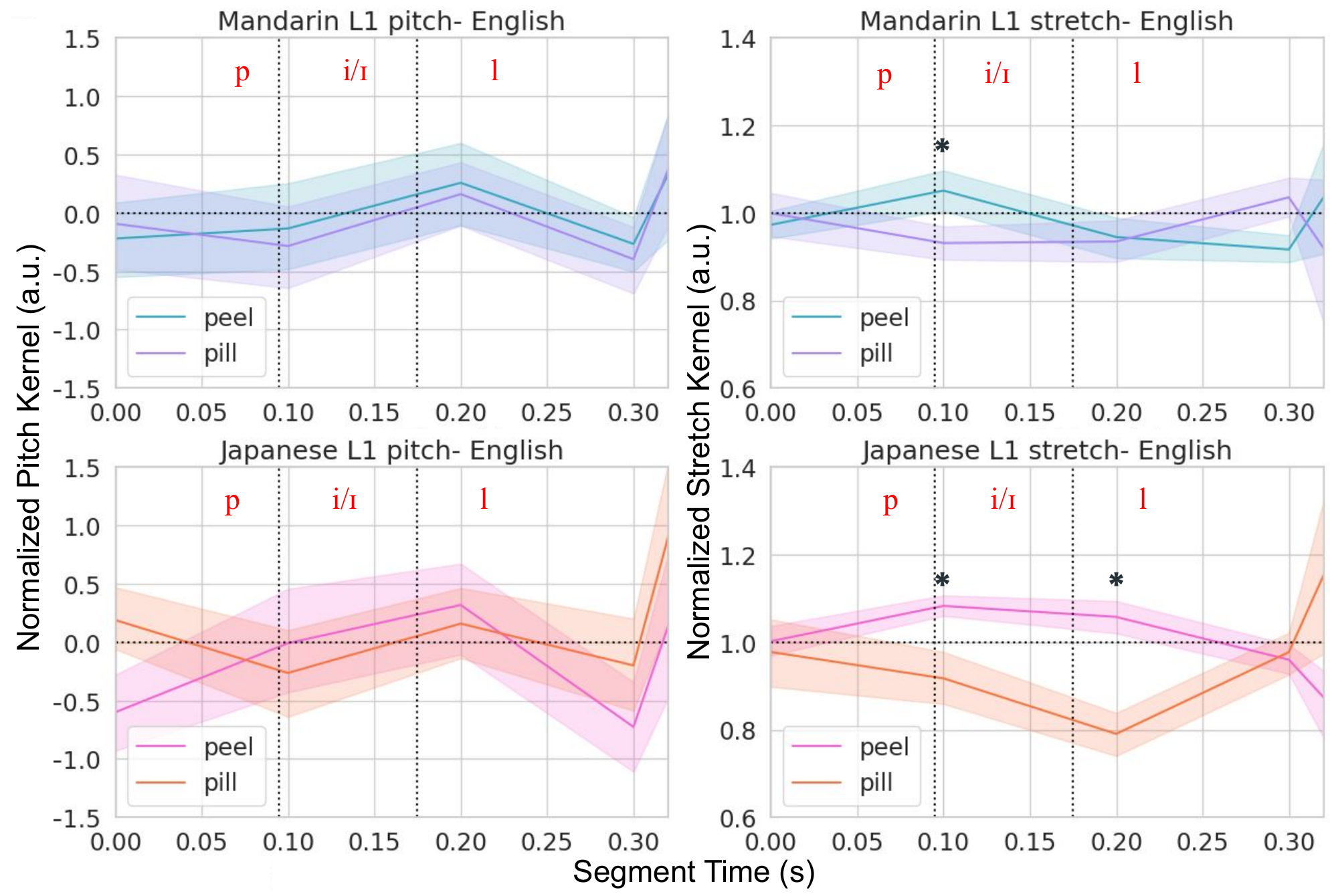}
  \caption{{\bf English words reverse-correlation results, continued (ML1, JL1):} Pitch (left) and speech rate (right) kernels for Mandarin-L1 (top) and Japanese-L1 (bottom) speakers for the English words ``peel'' and ``pill''. Smaller values for the stretch kernel mean shorter duration, i.e. {\it faster} speech rate, and for the pitch kernel lower pitch.}
  \label{fig:mandarin-word}
\end{figure}

A second subset of results concerns how L1 (English) and L2 (French, Chinese, Japanese) speakers perceived English-language stimuli, either as single words (\emph{``peel''} vs \emph{``pill''}) or complete sentences (ex. \emph{``I heard them say pill'')}.
ccepting either extended abstracts (2-pag
\noindent{\bf Words: } We predicted that EL1 speakers' perception of \textipa{/i/}, a high/tense vowel, would be driven by a higher pitch and slower speech rate compared to \textipa{/I/}. Yet, in EL1, our results do not confirm this prediction (Fig.\ref{fig:english-word}, top) as no significant differences are found between the kernels. English L2 speakers do, however, show the expected effect on speech rate - although not on pitch. French L1 speakers display a significantly faster rate of speech for \textipa{/I/} (0.2s: t(25)=3.13, p=.005) (Fig.\ref{fig:english-word}, bottom). Mandarin speakers (ML1) show same expected lengthening in \textipa{/i/} and shortening in \textipa{/I/} (0.1s: t(25)=3.58, p=.001) (Fig.\ref{fig:mandarin-word}- top right) as in FL1, but earlier in the vowel than for FL1 speakers. Lastly, for Japanese speakers (JL1), we see a strong shortening across the entire vowel for \textipa{/I/} (0.1s: t(14)=5.01, p=$<$.001; 0.2s: t(14)=6.95, p$<$.001) (Fig.\ref{fig:mandarin-word}- bottom right). None of the second-language speakers show any significant effect for pitch, 

\noindent{\bf Phrases: } Regarding the surrounding context of the word, the literature predicts the existence of contextual contrast effects, i.e., that \textipa{/I/} is driven by {\it lower} pitch and \emph{faster} speech rate before the target word. In actuality, like in the French phrase reverse correlation, our data reveals a mix of contrastive and congruent contextual influences. 

\begin{figure}[]
  \centering
  \includegraphics[width=0.85\linewidth]{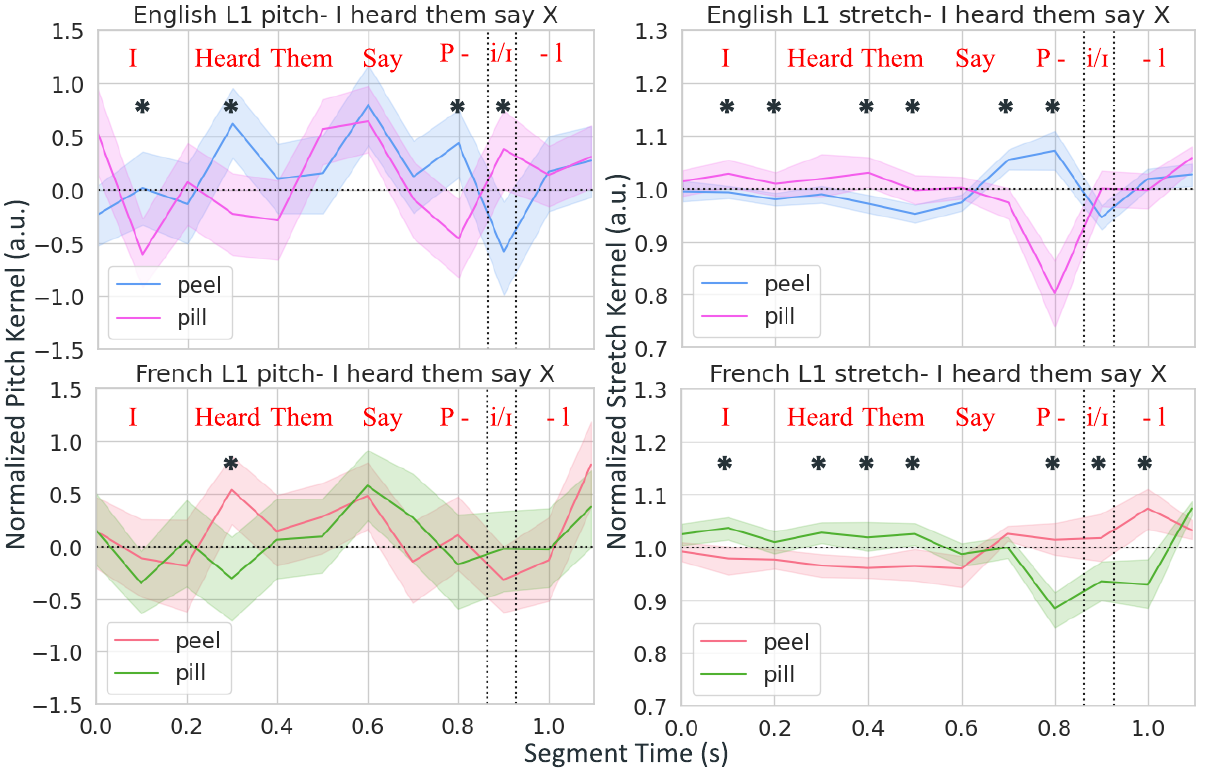}
  \caption{{\bf English phrase reverse-correlation results (EL1, FL1)}: Pitch (left) and speech rate (right) kernels for English-L1 (top) and L2 (bottom) speakers for the English phrases containing ``peel'' and ``pill''. Smaller values for the stretch kernel mean shorter duration, i.e. {\it faster} speech rate, and for the pitch kernel lower pitch.}
  \label{fig:english-phrase}
\end{figure}

\begin{figure}[]
  \centering
  \includegraphics[width=0.85\linewidth]{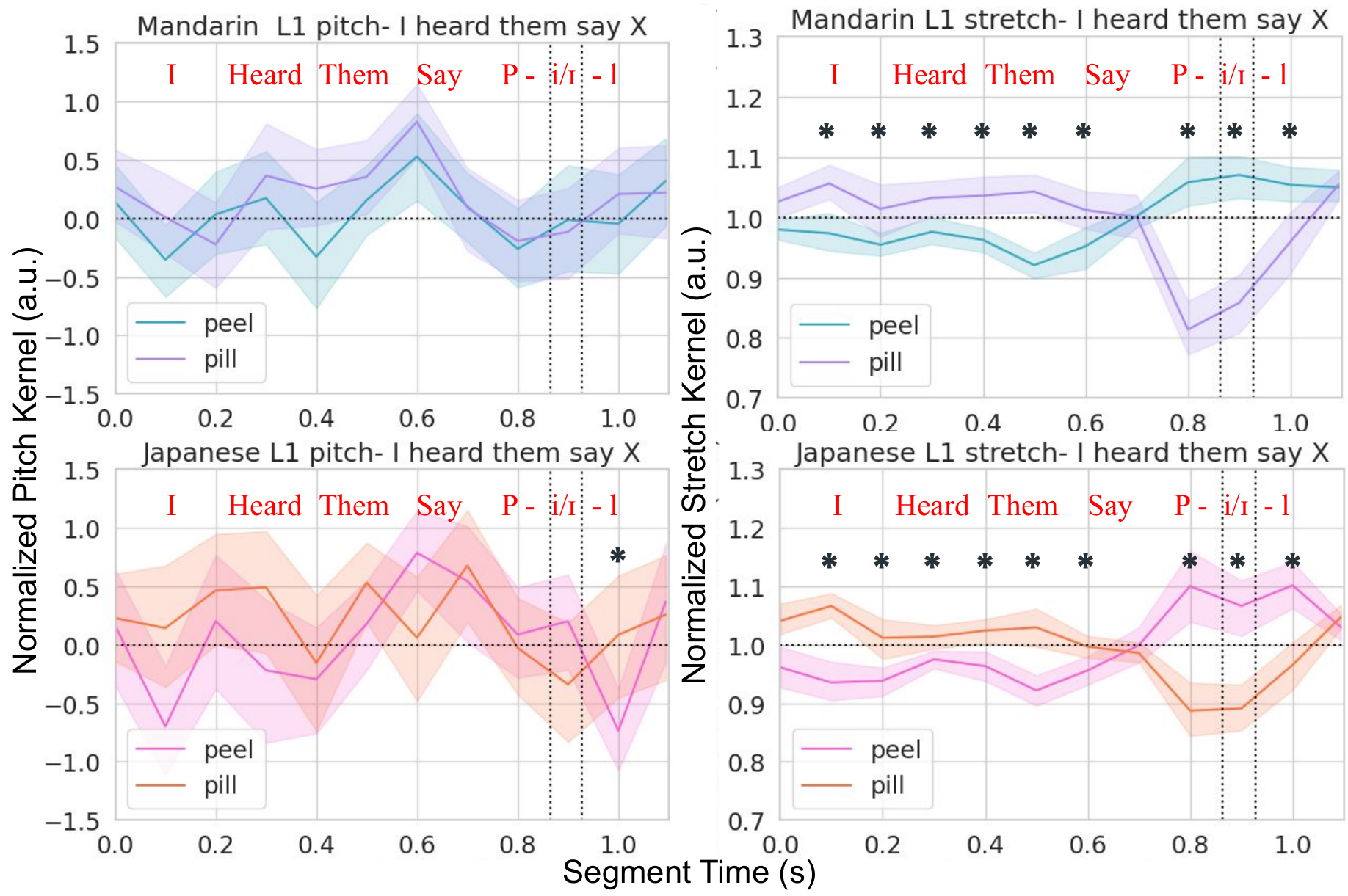}
  \caption{{\bf English phrase reverse-correlation results, continued (ML1, JL1):} Pitch (left) and speech rate (right) kernels for Mandarin-L1 (top) and Japanese-L1 (bottom) speakers for the English phrases containing ``peel'' and ``pill''. Smaller values for the stretch kernel mean shorter duration, i.e. {\it faster} speech rate, and for the pitch kernel lower pitch.}
  \label{fig:mandarin-phrase}
\end{figure}

For pitch, contrary to theoretical predictions, \textipa{/I/} is driven by a significantly higher pitch (0.9s: t(25)=-2.69, p=.013) for EL1 speakers. Moreover, pitch shows contrastive effects within the phrase, both distally (0.1s: t(25)=3.23, p=.003, 0.3s: t(25)=3.68, p=.001) and in the immediate proximity of the target word (0.8s: t(25)=2.91, p=.007) (Fig. \ref{fig:english-phrase}- top left). For English L2 speakers, as in the word, we do not see significant differences in the target word for pitch, aside from Japanese speakers that prefer a slightly lower pitch in ``peel'' following the vowel (1.0s: t(13)=-2.66, p=.019) (Fig. \ref{fig:mandarin-phrase}- bottom left). Instead, there appears to be a preferred pitch pattern over the entire phrase that is interestingly similar across FL1 and ML1 speakers, especially within the target word.

EL1 speech rate (Fig. \ref{fig:english-phrase}-top right) exhibits the expected long-term (distal) contrastive effect (0.1s: t(25)=-2.52, p=.018, 0.2s: t(25)=-2.31, p=.030, 0.4s: t(25)=-3.47, p=.002, 0.5s: t(25)=-2.34, p=.028), where a slower speech rate at the beginning of a phrase and a proximal congruent effect up to 200ms pre-target (EL1: 0.7s: t(25)=3.77, p=.001, 0.8s: t(25)=5.98, p$<$.001), biases the response towards the faster alternative \textipa{/I/}, resulting in an overall scissor-shape profile. However, the rate difference within the target vowel is not significant for EL1. This is not due to the speech rate at the proximal segment, which moves in the expected direction, but rather to the unexpected reversal from the expected speech rate within the vowel. 

Remarkably, we see  the scissor shaped pattern of speech-rate results was conserved in all groups of L2 speakers. As in the word task, we confirm the expected difference in speech rate, i.e. a shorter \textipa{/I/} (0.9s: t(25)=2.15, p=.042, 1.0s: t(25)=3.61, p=.001) for FL1 speakers. The FL1 show the same scissor-shape pattern we saw for EL1 speakers (Fig. \ref{fig:english-phrase}-bottom right), with distal contrastive effects (0.1s: t(25)=-3.02,p=.006, 0.3s: t(25)=-3.53,p=.002, 0.4s: t(25)=-3.07, p=.005, 0.5s: t(25)=-2.75, p=.011), and a strong proximal congruent effect (0.8s: t(25)=4.25, p$<$.001). ML1 speakers show the same scissor pattern with distal contrastive effects and proximal congruent effects (Fig. \ref{fig:mandarin-phrase}- top right) that we see in both the EL1 and FL1 speakers (0.1s: t(25)=-3.39, p=.002, 0.2s: t(25)=-2.11, p=.044, 0.3s: t(25)=-3.01, p=.006, 0.4s: t(25)=-3.65, p=.001, 0.5s: t(25)=-5.66, p$<$.001, 0.6s: t(25)=-2.28, p=.031, 0.8s: t(25)=6.36, p$<$.001, 0.9s: t(25)=5.43, p$<$.001, 1.0s: t(25)=2.44, p=.022). Once again, we see the same pattern in the JL1 group as we do in all other groups with significance across the entire phrase as in ML1 (Fig. \ref{fig:mandarin-phrase}- bottom right), despite the smaller sample size (0.1s: t(13)=-5.25, p$<$.001, 0.2s: t(13)=-2.82, p=.014, 0.3s: t(13)=-2.63, p=.021, 0.4s: t(13)=-3.02, p=.010, 0.5s: t(13)=-4.51, p=.001, 0.6s: t(13)=-2.14, p=.052, 0.8s: t(13)=4.01, p=.001, 0.9s: t(13)=4.08, p=.001, 1.0s: t(13)=3.39, p=.005).

\subsection{Discussion}

This experiment is, to the best of our knowledge, the first to apply a reverse correlation paradigm to examine how pitch and speech rate (i.e., traditionally prosodic cues) influence the perception of vowels in L1 and L2, both in isolated sounds and embedded in phrases (for a related recent study on how these cues influence the perception of word boundaries, see \cite{osses2023prosodic}).

Results in the French and English Word conditions (Figures \ref{fig:french-word}, \ref{fig:english-word} and \ref{fig:mandarin-word}) show that reverse correlation can uncover intrinsic pitch and speech rate characteristics of vowels. Strikingly, we saw the same profile repeated through the experiments, adhering to the known characteristics of the vowel in all cases across four languages, except for the English vowel for L1 speakers. In this case, although \textipa{/pIl/} is generally considered to be longer and have a higher intrinsic pitch than \textipa{/pil/}, they are both high vowels, and the reversal of pitch characteristics is not particularly surprising. For speech rate, it may be that EL1 speakers put more weight on timbre cues than speech rate cues, whereas the L2 speakers still rely heavily on prosody to disambiguate the foreign timbre. 

Results in the Phrase conditions (Figures \ref{fig:french-phrase}, \ref{fig:english-phrase} and \ref{fig:mandarin-phrase}) provide a detailed account of how prosodic contextual effects may bias the perception of vowels when they're embedded in a phrase. We were able to replicate the predicted contrastive context effects in speech rate and weakly in pitch. Reverse correlation, however, helped us to uncover fine-tuned information on these effects. In particular, the effects are distally contrastive but become congruent proximal to the word, with a tipping point around 200ms before the target in French, Mandarin, and Japanese-L1 and 300ms before the target in English-L1. In general, we see weaker effects for pitch than speech rate, especially within the French phrase, and in general, the EL1 speakers saw more exaggerated pitch profiles in both languages. 

Additionally, we found that L2 speakers could map remarkably well to the mental representation of L1 speakers both in terms of the intrinsic characteristics of the vowel and the contrastive effects of the context, in some cases even more significantly than L1 speakers. One reason for this may be that L1 speakers already are at ceiling performance when discriminating vowels based on spectral cues and do not typically rely on additional cues from the context, while context pitch and speech rate appear to constitute robust - and potentially determinant - cues for vowel perception in L2 speakers who may lack sufficiently resolute spectral representations. 

Finally, it is also worth noting that phoneticians often concentrate on examining one single cue at a time to deeply understand its significance for perception. This, however, is not how humans perceive sound, and reverse correlation offers a methodology to explore combinations of cues at once, producing results that may be more akin to everyday human perception. 

Overall, the most prominent effect found here is the scissor-shaped pattern of speech rate. That pattern was shown to bias vowel perception in both L1 and 3 samples of L2 speakers for both French and English vowel pairs. It was replicated with a strikingly similar shape in all experiments. In the rest of this chapter, we therefore focus on this perceptive mechanism, first to validate that this has a behavioural effect in a wider selection of vowels (Section \ref{sec:validation}, below) and, second, to use it as a strategy to improve L2 clarity in a state-of-art TTS architecture (Section \ref{sec:clarity}).

\section{Validation of Durational Vowel Control}
\label{sec:validation}

Reverse correlation allows us to uncover psychoacoustic mechanisms that we use unconsciously to differentiate sounds. Our results suggested a mechanism to improve comprehension when the listener struggles to use the primary formant cues, such as with L2 speakers \cite{kewley2005influence}. However, we need to confirm that these mechanisms translate to macroscopic behavioural changes (i.e., change perceptual responses) through validation experiments. First, we found a strong effect of duration within the target word and a weaker effect within the context. Still, we do not know the exact amount of duration change required to elicit the desired effects, i.e., a change in perception between a minimal tense/lax pair. In the following, we systematically manipulated the scissor manipulation's intensity to establish perception thresholds. Second, reverse correlation kernels have shown effects spanning the complete phrase and vowel but do not provide information on their respective strengths. In the following, we explore duration changes in the word and context, the context only, and the word only. Third, our previous results theoretically only hold for the two vowel pairs investigated. In the following, we chose to focus on English, as English has more clearly defined long and short (tense/lax) vowels that do not rely explicitly on surrounding consonants to determine their duration, and introduce another tense/lax vowel pair: ``full'' (\textipa{/fUl/}) and ``fool'' (\textipa{/ful/}). Finally, while reverse correlation was conducted on manipulated ambiguous vowels for practical reasons (avoiding bias in a 1-interval task), we do not know how they extend to more ecological situations. In the following,  the vowels are no longer ambiguous, allowing us to better understand and confirm whether L1 speakers rely more strongly on timbre cues while L2 rely on prosody.

\subsection{Stimulus generation}

We generated phrase stimuli (e.g. \emph{``I heard them say fool''}) that incorporated the ``scissor-shape'' profile of speech-rate uncovered in the previous experiment. To do this, we again use Matcha-TTS \cite{mehta2024matcha} (see Chapter 3), which has phoneme level duration control and allows us to manipulate the speech rate at specific points of the phrase.  

The fact that the reverse correlation kernels in the previous Phrase results (Fig. \ref{fig:english-phrase} and \ref{fig:mandarin-phrase}) had smaller weights (the $y$ axis of the figures respresent the kernel weights)  outside than within the target word suggests that the duration change required in the context to elicit a perception effect should be larger than in the word (kernel amplitudes are roughly equivalent to sensitivity \cite{Mur11}). However, we found that large duration changes in the context resulted in a highly unnatural speech rate that would not be applicable in real-world use cases. Therefore, we opted to make a smaller duration change to the context than within the target word. Within the word, we tested a duration change ranging from 0.5x speed to 2.0x speed at increments of 0.2, both increasing and decreasing from 1.0x speed, resulting in 11 different stretch manipulations. The corresponding context duration change ranged from 0.67x to 1.5x speed at increments of 0.1, both increasing and decreasing from 1.0x speed. Context and word duration were applied in opposite directions, e.g., when the word was 2.0x stretched, the context was compressed at 0.67x simultaneously. To explore the contribution of duration changes in the word and the context individually, we tested three conditions: duration manipulations on the context and the word, the word only, and the context only. The duration changes for the word-only and context-only conditions were the same as when the word and context were combined.

Using Matcha-TTS, these modifications were made by hand by applying an array of the same length as the phonemized phrase. This method has been shown to result in natural and effective duration changes in TTS systems containing a phonemizer \cite{joly23_ssw}. In this array, the phonemes up until the pause before the target word contained the context multiplier, and phonemes beginning at the space before the target word contained the word multiplier. This \emph{clarity duration} multiplier is applied just after the base speech rate multiplier, where the base rate multiplier is an array the same length as the phonemized phrase containing only the speech rate provided at synthesis (in our case, 0.75). The Matcha-TTS text encoder returns $x_\mu$: the average output of the encoder, $\log w$: the log duration predicted by the duration predictor, and $x_{\text{mask}}$: the mask for the text input. The clarity array $c array$ is applied via Hadamard product to the array resulting from the Hadamard product of $x_{\text{mask}}$, the exponential of $\log w$ and the base speech rate multiplier $speechrate$ (Algorithm~\(\ref{eq:step1}\)). The resulting $y_{\text{lengths}}$ and $y_{\text{max\_length}}$ are then used to calculate the attention alignment map.

 \begin{equation}
\begin{aligned}
    w &= e^{\log w} \odot x_{\text{mask}} \\
    w_{\text{ceil}} &= (\lceil w \rceil \odot \text{speechrate}) \odot \text{c\_array} \\
    y_{\text{max\_length}} &= \max\left(\max\left(1, \sum_{i, j} w_{\text{ceil}}[i, j]\right)\right) \\
\end{aligned}
\label{eq:step1}
\end{equation}

This modification was added to the single-speaker Matcha-TTS, and the LJ Speech checkpoint\footnote{url{https://drive.google.com/drive/folders/17C\_gYgEHOxI5ZypcfE\_k1piKCtyR0isJ}} was used. For each duration step, the phrases ``I heard them say peel'', ``I heard them say pill'', ``I heard them say fool'', and ``I heard them say full'' were generated, resulting in 44 different stimuli (4 phrases $\times$ 11 manipulations). 

\subsection{Experimental procedure}

We used the Gorilla Experiment Builder\footnote{https://gorilla.sc/} to structure our experiments. The participants were first asked to provide demographic information on the language they first learned, their most commonly used daily language, their age and gender, and their self-rated English proficiency. We also collected pronunciation tests; however, this remains to be analyzed in the future. 

\begin{figure}[]
  \centering
  {%
\setlength{\fboxsep}{10pt}%
\setlength{\fboxrule}{1pt}%
\fbox{  \includegraphics[width=0.6\linewidth]{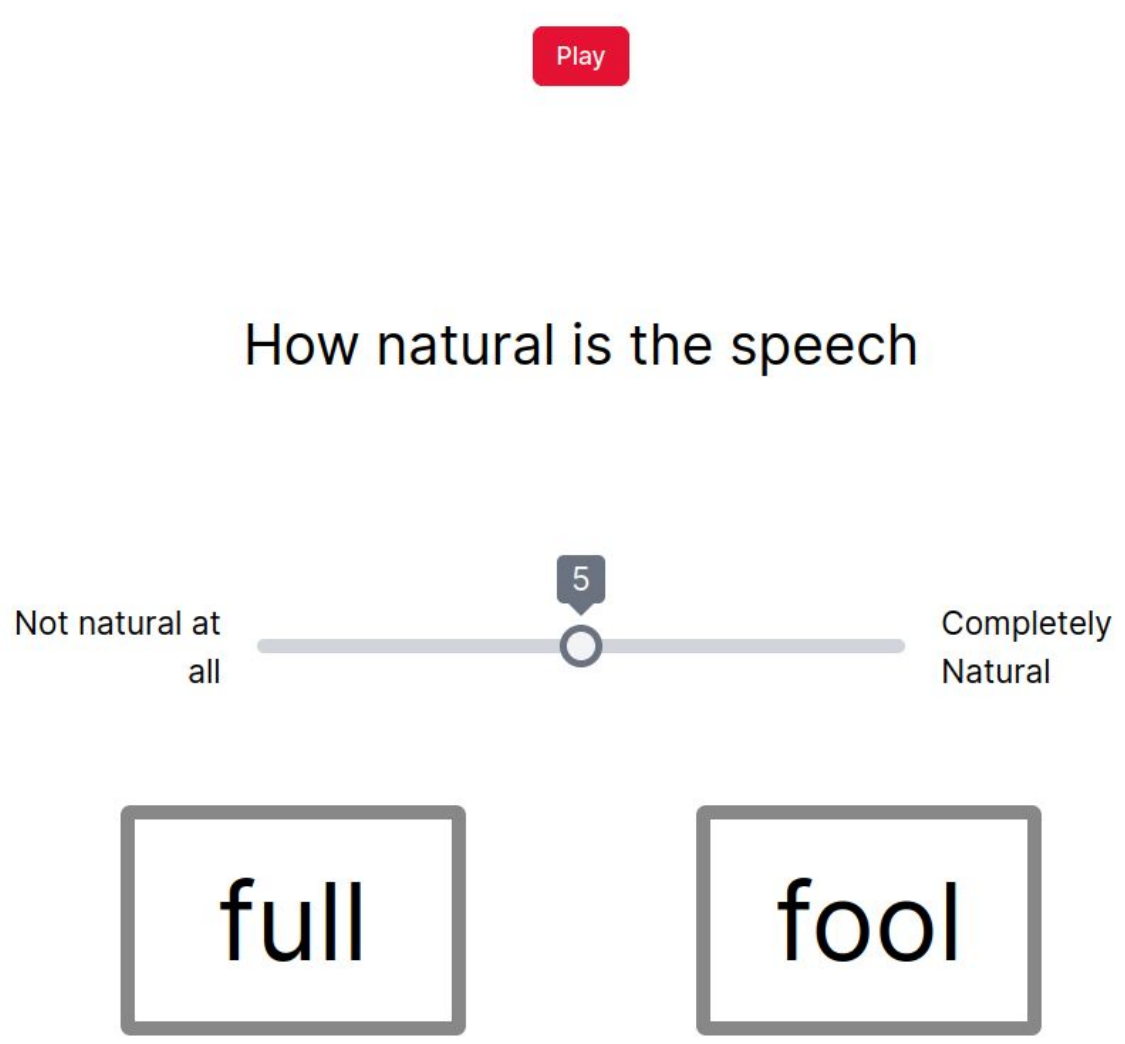}
}%
}%

  \caption{Screenshot of validation experiment set up in Gorilla.}
  \label{valexp}
\end{figure}

Participants were then presented successive trials consisting of a single stimulus, for which they were asked to choose which of 2 alternative words (``pill'' or ``peel'', or ``full'' or ``fool'') they thought it included (1-interval, 2-alternative forced choice). Each of the 44 stimuli (4 phrases, 11 manipulations) was presented 5 times, in random order, resulting in 220 trials. The three conditions, word and context, word only, and context only, were varied between-participants. There were two attention checks, and the experiment took roughly 30 minutes. In each trial, participants could listen to the phrase once. They were also asked to provide a naturalness MOS (nMOS) score from the MOS-X2 \cite{Lewis2018InvestigatingMR} at each trial. An example of the setup can be seen in Fig. \ref{valexp}.

\subsubsection{Participants}
We recruited N = 145 participants via Prolific. We maintained French as our target L2 population but recruited a smaller set of Mandarin and Japanese L1 speakers to confirm our reverse correlation results. There were N = 50 (17F, age = 34.36$\pm$9.74) English-L1 speakers (25 context + word, 25 context + word + noise), N = 75 (36F, age = 34.54$\pm$11.06) French-L1 speakers (25 context + word, 25 context only, 25 word only), N = 10 (7F, age = 28.70$\pm$7.82) Mandarin-L1 speakers (context + word), and N = 10 (8F, age = 36.40$\pm$13.44) Japanese-L1 speakers (context + word). The participant language demographics can be seen in Table \ref{tab:participant_demographics-validation}.

\begin{table}[]
    \centering
    \caption{Participant language demographics for durational vowel control validation study}
    \begin{tabular}{lc}
        \textbf{Category} & \textbf{Count} \\
        \midrule
        \midrule
        \multicolumn{2}{l}{\textbf{English-L1}} \\
        \midrule
        \midrule
        \multicolumn{2}{l}{\textbf{Daily Language}} \\
        English & 50 \\
        \midrule
        \multicolumn{2}{l}{\textbf{Self-Rated English Proficiency}} \\
        \multicolumn{2}{l}{\textbf{1(no proficiency) - 5(fluent)}} \\
        Score 4 & 4 \\
        Score 5 & 46 \\
        \midrule
        \midrule
        \multicolumn{2}{l}{\textbf{French-L1}} \\
        \midrule
        \midrule
        \multicolumn{2}{l}{\textbf{Daily Language}} \\
        French & 52\\
        English & 21 \\
        French \& English & 2\\
        \midrule
        \multicolumn{2}{l}{\textbf{Self-Rated English Proficiency}} \\
        \multicolumn{2}{l}{\textbf{1(no proficiency) - 5(fluent)}} \\
        Score 1 & 2 \\
        Score 2 & 4 \\
        Score 3 & 11 \\
        Score 4 & 18 \\
        Score 5 & 40 \\
        \midrule
        \midrule
        \multicolumn{2}{l}{\textbf{Mandarin-L1}} \\
        \midrule
        \midrule
        \multicolumn{2}{l}{\textbf{Daily Language}} \\
        English & 7 \\
        Mandarin & 2\\
        English \& Mandarin & 1\\
        \midrule
        \multicolumn{2}{l}{\textbf{Self-Rated English Proficiency}} \\
        \multicolumn{2}{l}{\textbf{1(no proficiency) - 5(fluent)}} \\
        Score 2 & 1 \\
        Score 3 & 2 \\
        Score 4 & 5\\
        Score 5 & 2 \\
        \midrule
        \midrule
        \multicolumn{2}{l}{\textbf{Japanese-L1}} \\
        \midrule
        \midrule
        \multicolumn{2}{l}{\textbf{Daily Language}} \\
        Japanese & 5 \\
        English & 5\\
        \midrule
        \multicolumn{2}{l}{\textbf{Self-Rated English Proficiency}} \\
        \multicolumn{2}{l}{\textbf{1(no proficiency) - 5(fluent)}} \\
        Score 3 & 1 \\
        Score 4 & 2 \\
        Score 5 & 5 \\
    \end{tabular}
    \label{tab:participant_demographics-validation}
\end{table}

\subsection{Results}
\subsubsection{Manipulation of both context and word}

\emph{French-L1}. Fig. \ref{frenchcwresult} {shows how different levels of manipulations of both context and word duration affect the normalized gain of accuracy (proportion of correct response) of French-L1 participants, relative to the accuracy obtained when no manipulation is applied. Baseline accuracy is relatively high, at 80.8\% for ``pill'', 76.8\% for ``peel'', 83.2\% for ``full'' and 32.0\% for ``fool''. As the French speakers had high English proficiency, it is unsurprising that participants perform relatively well on the baseline. Yet, we see a bias towards the lax vowel (left) (80.8\% accuracy ``pill'' vs. 76.8\% accuracy ``peel''; 83.2\% accuracy ``full'' vs 32.0\% accuracy ``fool'') that we would like to overcome to improve comprehension. For ``fool'' (bottom right), specifically, the TTS struggled to synthesize clear formants, and even the EL1 speakers struggle to identify the correct word at the baseline (see \emph{English-L1}).

We observe that we can overcome the bias towards lax vowels by increasing the length of tense vowels (right) (``peel'': 1.6x/w-0.77x/c: t(25)=2.57, p=.017, 1.8x/w-0.71x/c: t(25)=2.87, p=.008, 2.0x/w-0.67x/c: t(25)=3.36, p=.003; ``fool'': 1.4x/w-0.83x/c: t(25)=2.67, p=.013, 1.6x/w-0.77x/c: t(25)=4.38, p$<$.001, 1.8x/w-0.71x/c: t(25)=7.24, p$<$.001, 2.0x/w-0.67x/c: t(25)=5.28, p$<$.001). We do not observe a significant improvement in performance for the lax vowels as the word becomes shorter. Still, we do see that, although the participants have high proficiency, they can be convinced to hear the tense vowel if the duration of a lax vowel becomes long (left) (``pill'': 1.2x/w-0.9x/c: t(25)=-2.75, p=.011, 1.4x/w-0.83x/c: t(25)=-2.30,p=.031, 1.6x/w-0.77x/c: t(25)=-3.32, p=.003, 1.8x/w-0.71x/c: t(25)=-3.29, p=.003, 2.0x/w-0.67x/c: t(25)=-4.11, p$<$.001; ``full'': 1.2x/w-0.9x/c: t(25)=-2.78, p=.010, 1.6x/w-0.77x/c: t(25)=-3.36, p=.003, 1.8x/w-0.71x/c: t(25)=-4.68, p$<$.001, 2.0x/w-0.67x/c: t(25)=-4.87, p$<$.001).

\begin{figure}[]
  \centering
  \includegraphics[width=\linewidth]{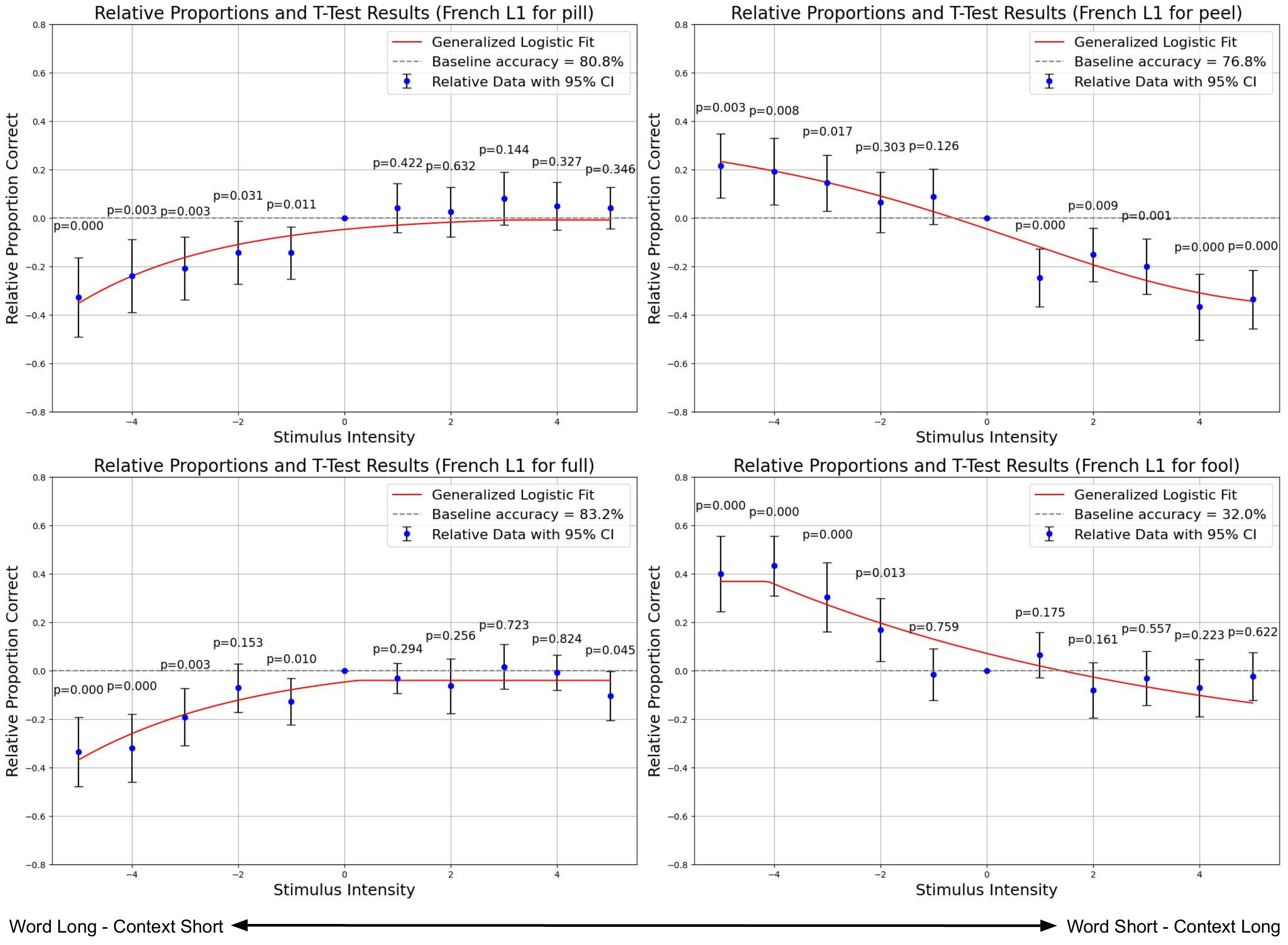}
  \caption{{\bf Word identification performance in French-L1 participants, when both context and word are manipulated}: Proportion of correct responses for each level of duration multiplier, from 2.0x (left)  to 0.5x (right) in the word and from 0.67x (left) to 1.5x (right) in the context, normalized relative to no manipulation (0.75x). P-values correspond to one-sample t-tests for the difference to the baseline. The red curve is a generalized logistic curve fit.}
  \label{frenchcwresult}
\end{figure}

\emph{English-L1}. Fig. \ref{englishcwresult} shows how different levels of manipulations affect the gain in accuracy of English-L1 participants relative to baseline. Baseline accuracy is 96\%  for ``pill'', 94.4\% for ``peel'' and ``full'', and, as above, 32\% for ``fool''. We observe that contrary to the previous experiment where vowels were spectrally manipulated to be ambiguous, English speakers are now able to perform with nearly 100\% accuracy for the ``pill'', ``peel'' and ``full'' and that, contrary to French-L1, they are mostly insensitive to the manipulation. It is possible that because they have more robust auditory representations of the spectral content of vowels in their native language, English speakers are able to use the formants of the vowel to differentiate the words and are not easily swayed by changes in duration. While this pattern of results seems to contradict the English-L1 kernels seen in the previous reverse-correlation experiment (Fig \ref{fig:english-phrase}), we do, in fact, see evidence that even English-L1 participants may start relying on duration cues in situations where timbral cues are not easily processed. First, we observe that extreme shortening of ``peel'' does begin to affect the performance (top right) (0.55x/w-1.4x/c: t(25)=-2.61, p=.015, 0.5x/w-1.5x/c: t(25)=-2.37, p=.026). Second, and most strikingly, when the TTS struggled to clearly generate the word ``fool'' (resulting in a much lower baseline performance), we see English speakers using the same duration cues as the French-L1 speakers, with performance increasing as the duration of ``fool'' increases (bottom right) (1.4x/w-0.83x/c: t(25)=2.70, p=.013, 1.6x/w-0.77x/c: t(25)=4.38, p$<$.001, 1.8x/w-0.71x/c: t(25)=5.01, p$<$.001, 2.0x/w-0.67x/c: t(25)=5.58, p$<$.001).

\begin{figure}[]
  \centering
  \includegraphics[width=\linewidth]{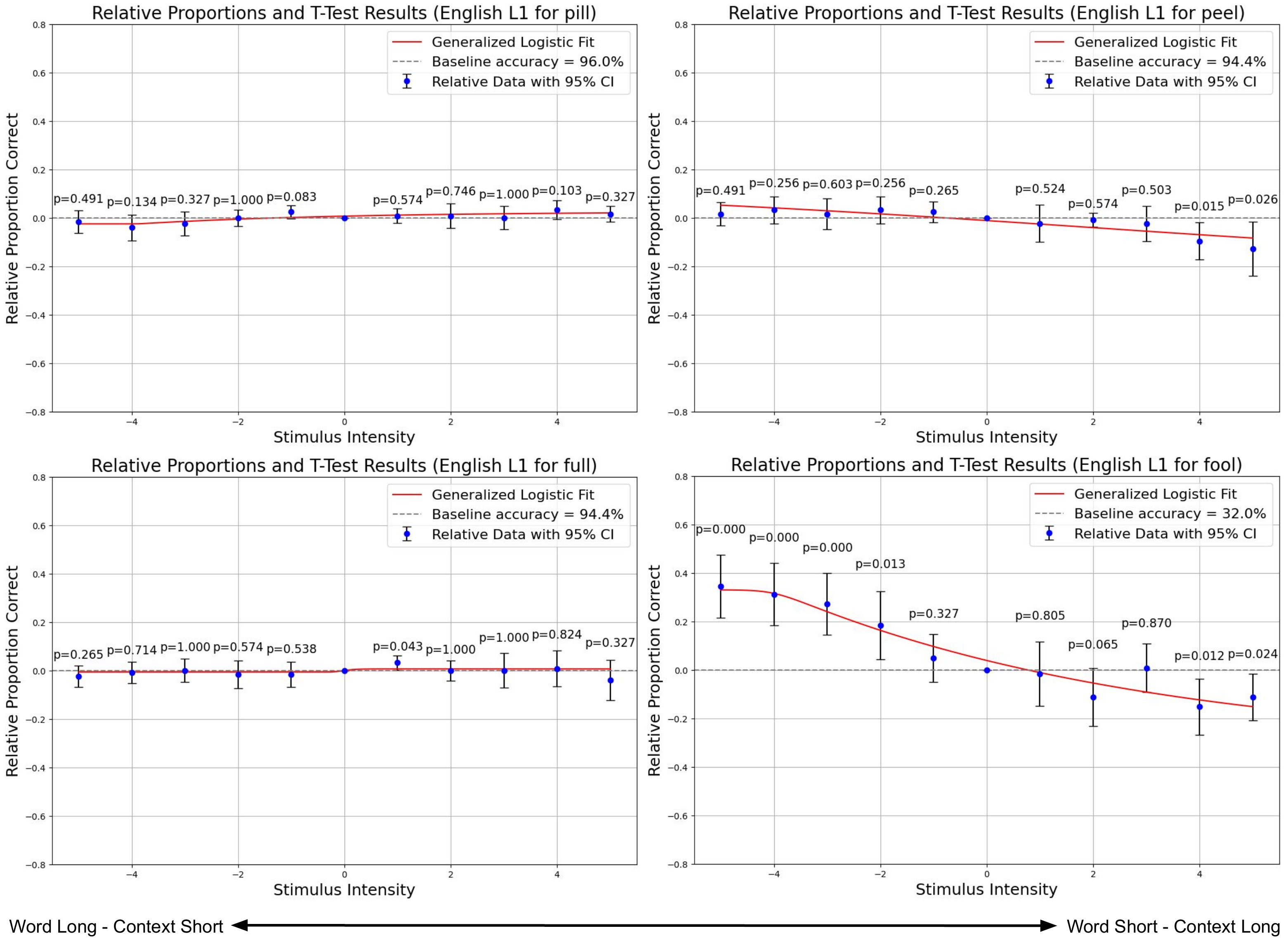}
  \caption{{\bf Word identification performance in English-L1 participants, when both context and word are manipulated:} Proportion of correct responses for each level of duration multiplier, from 2.0x (left)  to 0.5x (right) in the word and from 0.67x (left) to 1.5x (right) in the context, normalized relative to no manipulation (0.75x). P-values correspond to one-sample t-tests for the difference to the baseline. The red curve is a generalized logistic curve fit.}
  \label{englishcwresult}
\end{figure}

\emph{Mandarin-L1}.  Word identification performance in Mandarin-L1 participants can be seen in Fig. \ref{mandarincwresult}. Like FL1 speakers, ML1 speakers exhibited a slight bias towards lax vowels (left) (88.0\% accuracy ``pill'' vs. 82.5\% accuracy ``peel''; 78.0\% accuracy ``full'' vs 17.0\% accuracy ``fool''), albeit less statistically (note the smaller sample size). As in FL1 and EL1 for ``fool'', we observe an improvement in performance for tense vowels as the duration is lengthened (right). Interestingly, for ``fool,'' the performance on the baseline was particularly low, even compared to the shortened stimuli (bottom right); the rest of the curve displayed the same trend as we saw in FL1 and EL1.

\begin{figure}[]
  \centering
  \includegraphics[width=\linewidth]{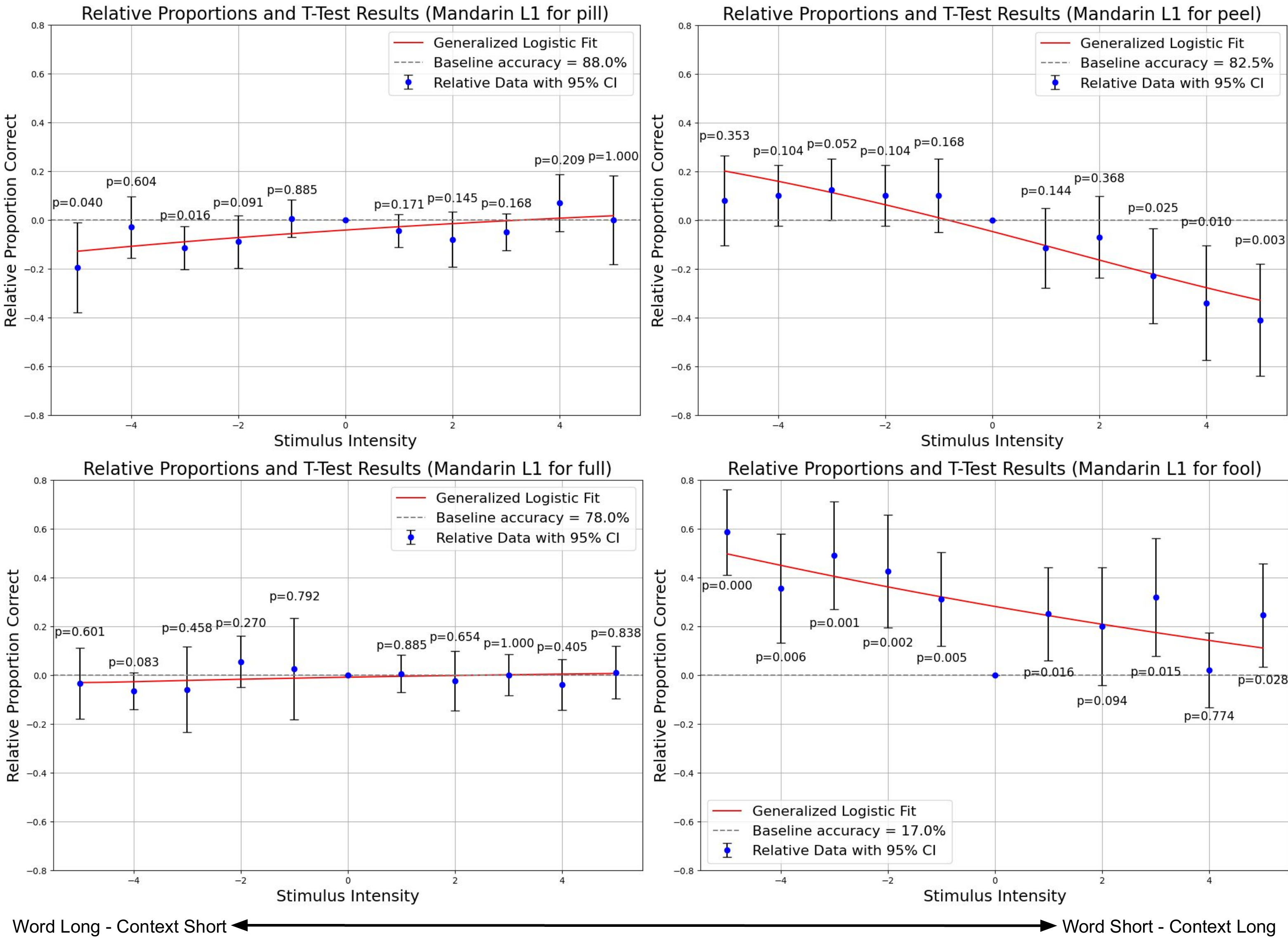}
  \caption{{\bf Word identification performance in Mandarin-L1 participants, when both context and word are manipulated:} Proportion of correct responses for each level of duration multiplier, from 2.0x (left)  to 0.5x (right) in the word and from 0.67x (left) to 1.5x (right) in the context, normalized relative to no manipulation (0.75x). P-values correspond to one-sample t-tests for the difference to the baseline. The red curve is a generalized logistic curve fit.}
  \label{mandarincwresult}
\end{figure}

\emph{Japanese-L1}.  Word identification performance in Japanese-L1 participants can be seen in Fig. \ref{japanesecwresult}. Despite reporting lower proficiency than the other L2 speakers, the Japanese speakers perform better than other L2 speakers on the baseline (98.0\% accuracy for ``pill'', 88.0\% for ``peel''; 98.0\%  for ``full''; 28.0\% for ``fool''), with a similar bias towards the lax vowel (left). Very similarly to French-L1 and, to some extent, to Mandarin-L1, Japanese speakers are sensitive to duration manipulations in the stimuli and can be easily swayed to hear the incorrect vowel when the duration is changed in the wrong direction, e.g. when ``peel'' becomes too short (top right) or ``pill'' becomes too long (top left) (``pill'': 1.6x/w-0.77x/c: t(10)=-2.54, p=.032, 2.0x/w-0.67x/c: t(10)=-2.45, p=.037; ``full'': 1.2x/w-0.9x/c: t(10)=-2.69, p=.025, 1.4x/w-0.83x/c: t(10)=-3.54, p=.006, 1.6x/w-0.77x/c: t(10)=-3.58, p=.006, 1.8x/w-0.71x/c: t(10)=-6.47, p$<$.001, 2.0x/w-0.67x/c: t(10)=-5.30, p$<$.001, ``peel'': 0.83x/w-1.1x/c: t(10)=-6.33, $<$.001, 0.71x/w-1.2x/c: t(10)=-7.58, p$<$.001, 0.63x/w-1.3x/c: t(10)=-6.43, p$<$.001,  0.55x/w-1.4x/c: t(10)=-7.83, p$<$.001, 0.5x/w-1.5x/c: t(10)=-7.75, p$<$.001). For ``peel,'' we do not see a significant increase in accuracy by lengthening the word, which may be because of a ceiling effect (baseline was already at 88\%). In ``fool'', as in all other language groups, we see a strong increase in performance with lengthening of the target word (bottom right) (1.2x/w-0.9x/c: t(10)=2.62, p=.028, 1.4x/w-0.83x/c: t(10)=3.99, p=.003, 1.6x/w-0.77x/c: t(10)=4.61, p=.001, 1.8x/w-0.71x/c: t(10)=4.44, p=.002, 2.0x/w-0.67x/c: t(10)=3.15, p=.012).

\begin{figure}[]
  \centering
  \includegraphics[width=\linewidth]{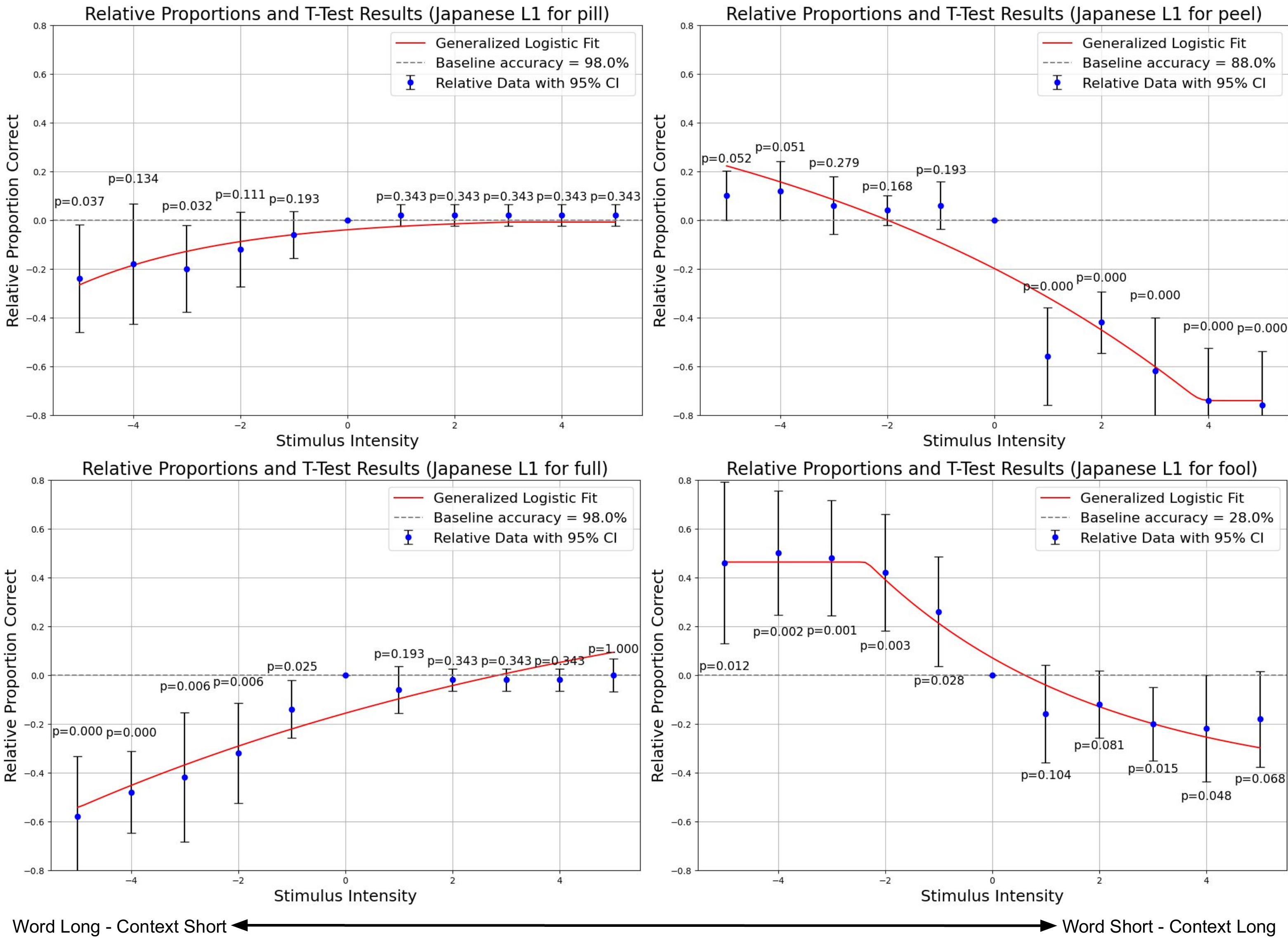}
  \caption{{\bf Word identification performance in Japanese-L1 participants, when both context and word are manipulated:} Proportion of correct responses for each level of duration multiplier, from 2.0x (left)  to 0.5x (right) in the word and from 0.67x (left) to 1.5x (right) in the context, normalized relative to no manipulation (0.75x). P-values correspond to one-sample t-tests for the difference to the baseline. The red curve is a generalized logistic curve fit.}
  \label{japanesecwresult}
\end{figure}

\emph{English-L1, with added noise}. The previous results suggest that L2 speakers in 3 languages rely consequentially on duration cues for word identification. However, L1 speakers are already at ceiling performance based on spectral cues and may not need or even use duration cues for the task in typical conditions. However, English-L1 results for ``fool'' stimuli, a word which, for technical reasons, was found to have poor synthesis results with our TTS system, suggest that, in more difficult or ambiguous listening conditions, even native speakers would default back to the same mechanism used as L2 speakers. To confirm this hypothesis,  we wanted to explore if it is possible to mask formantic information with distortion and background noise and force L1 English speakers to behave like L2 for ``peel'' and ``pill''. ``Fool'' and ``full'' were not tested in this condition as we already observed the duration mechanism without the need for background noise.

We first attempted to overlay a loud background noise, yet through a pilot study, we found that this simply resulted in a loss of the target word, and participants resorted to random guessing. We then aimed to create a sound similar to loud-speakers in a metro station, i.e., a distorted and distanced sound, with a crowd in the background. To achieve this sound, we used Audacity\footnote{https://www.audacityteam.org/}. First, a rectifier distortion was applied at 45\%; then reverberation was added with a room size of 22\%, a pre-delay of 10ms, a reverberance and dampening percentage of 50\%, a tone low, tone high, and stereo width of 100\%, and a wet and dry gain of -1bB. Finally, a background crowd noise was added so that all word duration in the phrase could be perceived, but it was difficult to recognize all of the speech clearly.

While baseline performance remains relatively high (84.5\% ``pill'' and 81.8\% ``peel''), results, seen in Fig. \ref{englishnoisecwresult}, show an apparent effect of duration manipulations on word recognition performance both for ``pill'' and `peel'', which mirrors what is seen in L2 speakers. These results, therefore, confirm that the duration perception mechanism evidenced in reverse-correlation experiments is one used by both L1 and L2 speakers (with French, Mandarin and Japanese L1s) of English as a fall-back strategy when spectral information on the word itself is unreliable, both for internal (L2 speakers) or external (L1 with noise) reasons. 

\begin{figure}[]
  \centering
  \includegraphics[width=\linewidth]{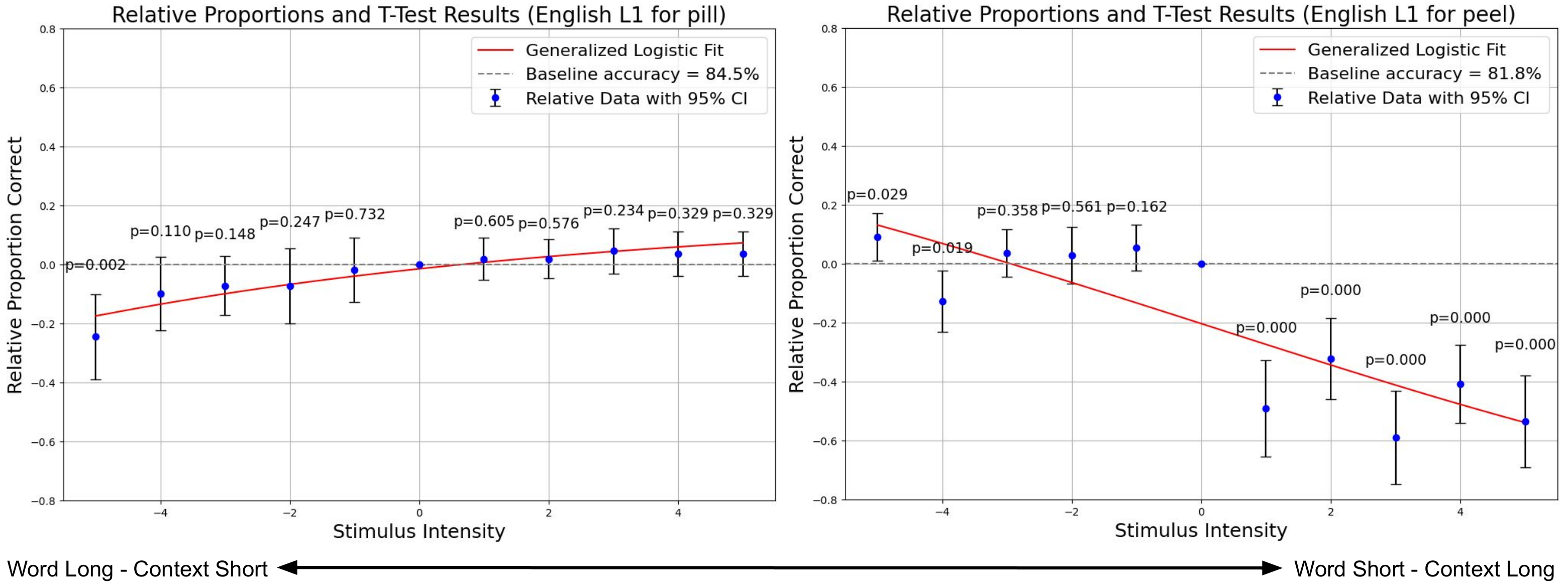}
  \caption{{\bf Word identification performance in English-L1 participants, when both context and word are manipulated, in the addition of background noise and distortion.} Proportion of correct responses for each level of duration multiplier, from 2.0x (left)  to 0.5x (right) in the word and from 0.67x (left) to 1.5x (right) in the context, normalized relative to no manipulation (0.75x). P-values correspond to one-sample t-tests for the difference to the baseline. The red curve is a generalized logistic curve fit.}
  \label{englishnoisecwresult}
\end{figure}

\subsubsection{Manipulation of context-only and word-only}

Finally, after confirming reverse-correlation results for simultaneous context and word manipulations of duration, we explored the contribution of manipulating only the context or only the word (i.e. applying either 0.67x  to 1.5x in the context and 1x in the word or 2.0x to 0.5x in the word and 1x in the context). We focused only on L1 French speakers, as the results above showed the effect could be seen identically in the other populations.

We observed that although reverse correlation kernels showed effects of duration both outside and inside the word, word identification performance was more strongly driven by word duration than context duration. The context manipulations resulted in only very slight improvements over the baseline (Fig. \ref{frenchcresult}), and modifying the word had as strong of an effect as modifying both the context and the word simultaneously (Fig. \ref{frenchcwresult}). 

\begin{figure}[]
  \centering
  \includegraphics[width=\linewidth]{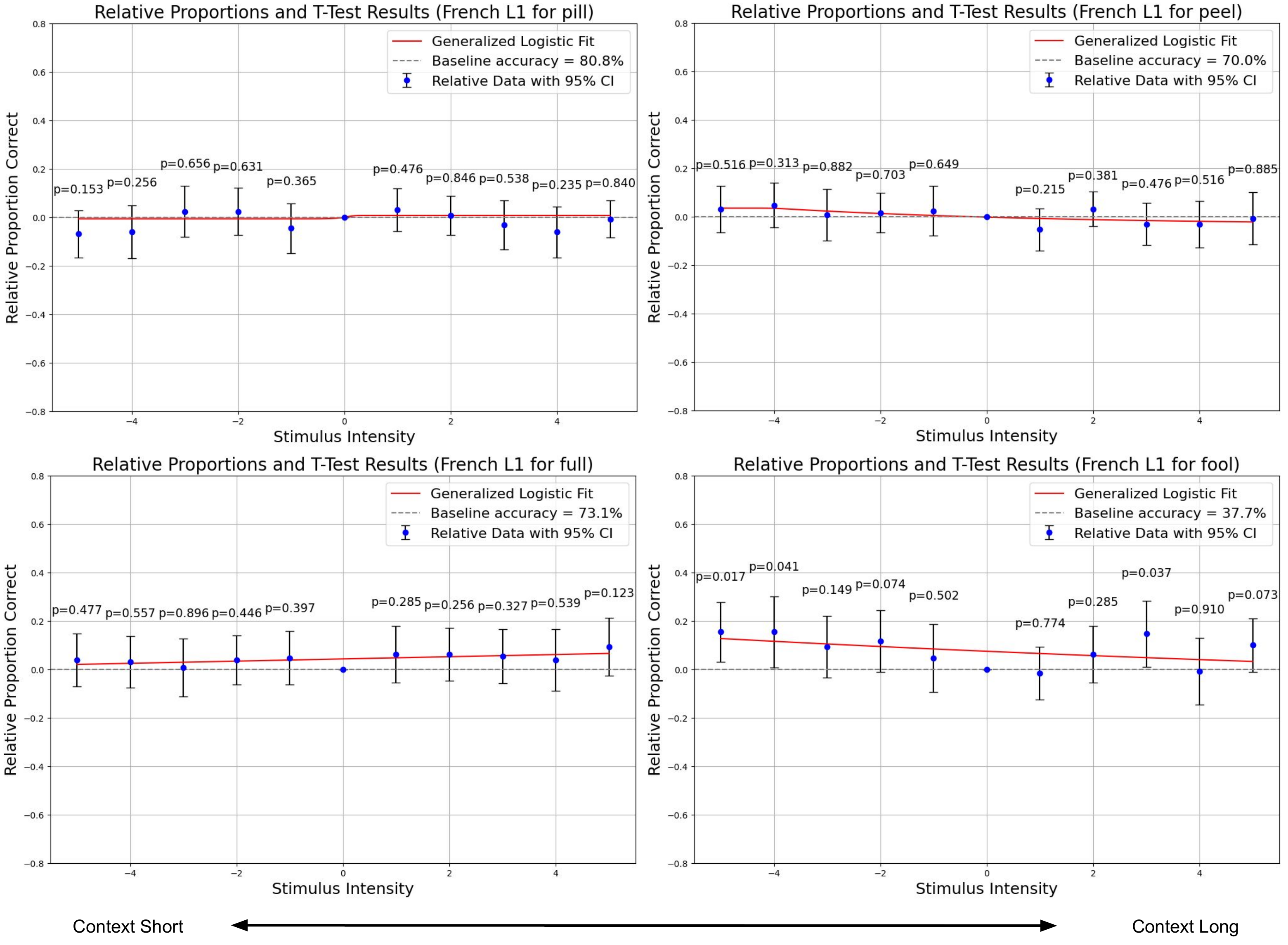}
  \caption{{\bf Word identification performance in French-L1 participants, when both only the context is manipulated:} Proportion of correct responses for each level of duration multiplier from 0.67x (left) to 1.5x (right) in the context, normalized relative to no manipulation (0.75x). P-values correspond to one-sample t-tests for the difference to the baseline. The red curve is a generalized logistic curve fit.}
  \label{frenchcresult}
\end{figure}

\begin{figure}[]
  \centering
  \includegraphics[width=\linewidth]{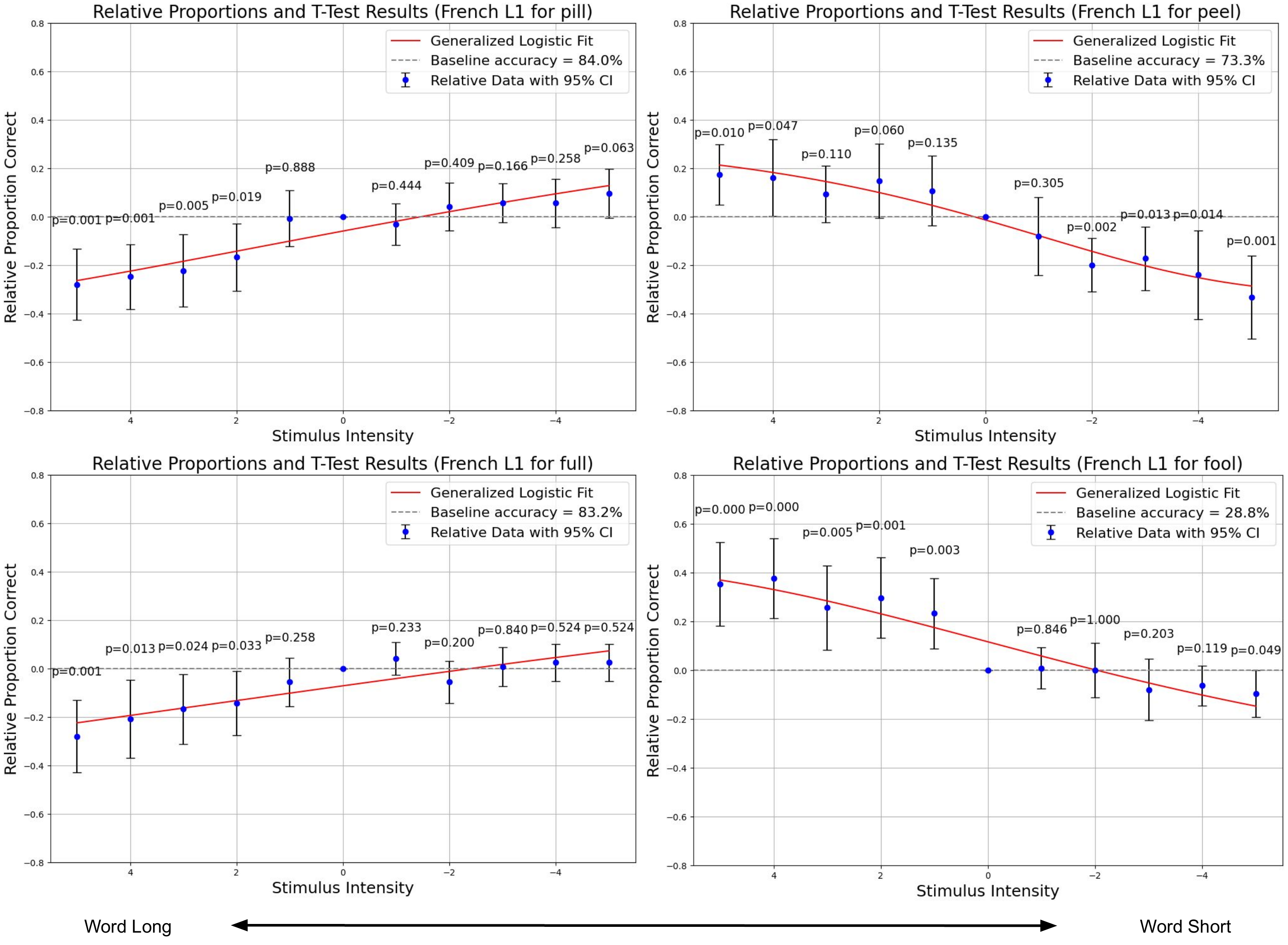}
  \caption{{\bf Word identification performance in French-L1 participants, when both only the word is manipulated:} Proportion of correct responses for each level of duration multiplier from 2.0x (left) to 0.5x (right) in the word, normalized relative to no manipulation (0.75x). P-values correspond to one-sample t-tests for the difference to the baseline. The red curve is a generalized logistic curve fit.}
  \label{frenchwresult}
\end{figure}

\subsection{Discussion}

Reverse correlation experiments in Section \ref{sec:revcor} showed that vowel perception was driven psychophysically by duration cues both in the target word and in its preceding context, but the question as to whether this mechanism could be used causally to drive word identification performance in ecological situations remained open. The present results showed that this was indeed the case, using non-ambiguous vowels and expanding the results to another tense/lax vowel pair (`full'' (\textipa{/fUl/}) and ``fool'' (\textipa{/ful/}). 

The mechanism was most strongly observed in the case of tense vowels, where there is generally a lower level of perception accuracy on the baseline. For these vowels, applying a lengthening allowed for an improvement in word identification, and a shortening of tense vowels decreased performance. In comparison, little advantage was seen when shortening lax vowels. This may be due to the bias towards lax vowels already existing in the baseline or to the fact that the words may become too short and difficult to hear. Instead, we observed that it is important not to lengthen words containing lax vowels, as this can lead to confusion with their tense counterparts. 

Strikingly, this was confirmed with remarkable consistency for all non-native FL1, ML1, and JL1 speakers. This is especially remarkable given the vast difference in reliance on duration cues for vowel differentiation in these three languages, i.e. Japanese relies primarily on duration, Mandarin relies primarily on pitch but uses duration as a secondary cue, and French is nearly void of a reliance on duration cues (see Sec. \ref{sec:revcor}). In EL1 speakers, eliciting this mechanism was more difficult, but we were able to evidence it when spectral cues were either poorly synthesized (``fool'') or masked with distortion and noise. This pattern of result is consistent with a universal duration-perception mechanism which acts as a fall-back strategy, which EL1 speakers who can easily use the formants to differentiate minimal pairs will only use in challenging listening conditions but may be used routinely, and perhaps even principally, by L2 speakers.

Although we collected MOS scores for naturalness and do not report on their detailed analysis here, we saw no significant differences in naturalness ratings from the baseline as the duration changes were applied. We also did not see differences in naturalness ratings between context and word, context-only and word-only. However, we do note that EL1 speakers began to trend towards lower naturalness in both extremes of stretch and compression and had a lower standard deviation across their responses overall. It appears that L2 speakers have a wide range of expectations and acceptance of naturalness that may relate to their English proficiency. This remains future work to explore.

Taken together, these results provide a promising strategy to improve the clarity of synthesized speech for L2 speakers by manipulating duration cues to enforce the correct perception of tense/lax alternatives. In more detail, as we did not see an improvement in either performance or naturalness through the addition of duration changes in the context, we hypothesize that a word-only modification should be sufficient to improve comprehension for L2 listeners. Moreover, as the duration effects seen above were asymmetric, we make the hypothesis that lax vowels should remain at the base speaking rate, but words containing a tense vowel that can be easily confused with a lax vowel minimal pair should be lengthened relative to the rest of the phrase. In the following, we implement this strategy in a complete TTS system, using a parsing technique to automatically identify portions of the phrase that should benefit from durational changes and validate that this strategy improves speech comprehension compared to two other control strategies (slowing down the difficult word or slowing down the whole sentence).

\section{L2 Clarity TTS}
\label{sec:clarity}

Once we had confirmed we could elicit behavioural changes to improve the perception of tense/lax vowels using simple linguistically driven duration changes, we added the changes we had applied manually as a new ``L2 clarity'' mode for Matcha-TTS in \cite{l2tts}. Moreover, we add the last tense/lax vowel pair:  \textipa{/A/} (cot) and \textipa{/2/} (cut).

\subsection{Pilot clarity mode}\label{pilot}

Initially, to enable L2 clarity mode in Matcha-TTS, we added a clarity flag that could be set to ``True'' or ``False'' at synthesis. If set to true, it automatically applied a stretch to words containing a tense vowel to increase clarity. This, however, becomes complicated when there are multiple vowels in a word, and the speech rate of a phrase can quickly become awkward with too many stretched words. We decided to eliminate context changes to minimize the number of duration changes in a phrase as we previously saw that the perception was driven primarily by the target word, and as we discovered through our validation, a lax vowel's duration does not need to be modified to increase clarity.

In order to reduce unnecessary and frequent duration changes we created a list of function words such as ``be'' and ``he'' that were ignored for clarity treatment. We ensured that no function words in this ``ignore'' list had a minimal pair, such as ``been'' (\textipa{/bin/}), which can be confused with ``bin'' (\textipa{/bIn/}). Moreover, diphthongs and words ending in \textipa{/i/} were ignored. The model first parsed the phrase through several steps:

\begin{enumerate}
    \item Parse each word to see if it contains a tense or lax vowel
    \item If the word is in the list of function words, it is ignored
    \item If the word contains tense vowels but no lax vowels, the clarity modification is applied
    \item If the word contains both tense and lax vowels, if the tense vowel has primary stress, the clarity modification is applied
\end{enumerate}

The clarity modification was applied as in Algorithm \ref{eq:step1}. A 1.6x stretch was applied across the entire word with a gradual ramp up and down to the base speech rate over the 6 phonemized items (if there are at least 6 phonemized items between target words, otherwise only as many as are available)  preceding and following the target word. Six items were chosen as this encompasses two phonemes preceding and following the target word (approximately 200-300ms \cite{ma21_interspeech} as per our reverse correlation results). We also apply a 1x stretch (the base speech rate) to lax words simultaneously, following the same parsing above to ensure that lax vowels are not stretched by surrounding tense vowel words. A 1.6x stretch was chosen to minimize duration changes, because as in our validation, we saw that perception performance quickly improved from the baseline by increasing the duration of the target word, and there was minimal improvement in performance between a 1.6x and 2.0x stretch.

We, however, discovered through a pilot study of French-L1-English-L2 speakers (N=50) that although we attempted to control unnecessary changes in duration, there still remained many duration changes in the context leading up to and following target words. Because of this, our results showed little improvement over the baseline, and the overall results were inconsistent. Interestingly, although in our validation study (Sec. \ref{sec:validation}) we found that the context changes alone were not enough to influence perception, these pilot results echo our reverse correlation findings that we need a contrasting and consistent difference in duration between the context and the target word to induce improvement in tense vowel comprehension, i.e., having too many duration changes in the phrase does not create a strong enough contrast in the target word to illicit improvements in comprehension. As such, we needed to control the duration changes in the TTS in a more defined manner.

\subsection{Final L2 clarity mode}
Based on our findings in the pilot experiment, we decided to include a markup to control which words are emphasized. As the text for a TTS must be generated either by a human or a large language model (LLM), these markups can be added to words that are either perceived as important, difficult, or lacking context in the sentence and should be emphasized by the TTS at the time of the text generation. This markup means that we no longer need to ignore function words, nor parse the entire sentence, instead, only the flagged words are parsed for clarity treatment. Once the clarity flag is enabled, as in the pilot, the user then surrounds difficult words with exclamation points, e.g., ``!peel!'', allowing the TTS to parse the only words to be treated for clarity. The parsing and duration modifications were then applied as follows:

\begin{enumerate}
    \item Parse each flagged word to see if it contains a tense or lax vowel
    \item If the word contains tense vowels but no lax vowels, the clarity modification is applied
    \item If the word contains both tense and lax vowels, if the tense vowel has primary stress, the clarity modification is applied
\end{enumerate}

\subsection{Stimulus generation}\label{gen}
We tested 4 different TTS styles: 1. \textbf{Base}: the base Matcha-TTS (0.75x speech rate), 2. \textbf{Stretch}: 1.2x ($0.75\times1.6)$ speech rate applied across the entire phrase, 3. \textbf{Emphasis}: 1.6x stretch applied across all target words (0.75x speech rate elsewhere), and 4. \textbf{Clarity}: 1.6x stretch applied across the target words containing a tense vowel (0.75x speech rate elsewhere).

In our reverse correlation and validation experiments, only a \textbf{single target word}, always at the end of a phrase, was tested. To test the robustness of the L2 clarity TTS when the target words occurred in other parts of the phrase, we tested clarity mode in several contexts. First, single target word phrases where the target word was in the middle of the phrase were tested. We then tested \textbf{double target word} phrases (e.g., ``Write down dull and doll''), where the targets could be at the end of the phrase, in the middle of the phrase, or at the beginning of the phrase. Additionally, we explored the performance when the TTS had a combination of tense and lax vowels in a single phrase; we tested a tense and a lax minimal pair, a tense and a lax that are not minimal pairs, two tense and two lax vowels. We aimed to use a variety of starting and ending consonants surrounding the target vowels, and all words had a minimal pair in English. We tested all tense/lax vowel pairs: \textipa{/i/} (peel) and \textipa{/I/} (pill), \textipa{/u/} (fool) and \textipa{/U/} (full),  \textipa{/A/} (cot) and \textipa{/2/} (cut). Lastly, we aimed not to semantically bias phrase meanings towards any word, that is, given the context outside of the target word neither word in the minimal pair made more logical or semantic sense. To do this, we asked ChatGPT-4o to provide a starting list of neutral phrases with varying lengths. These were then selected and modified to create our final list of 16 phrases and insert our target words (Appendix C).

Because participants would hear each of the phrases 4 times (each TTS style had the same phrases), we added 1 confusion phrase (randomly assigning a TTS style) for each of the phrases. A confusion phrase was the same phrase context using the opposite minimal pair, for example: ``She kept mentioning cot during the conversation.'' and the confusion phrase ``She kept mentioning cut during the conversation.'' Lastly, for the single target word TTS, ``clarity'' TTS was the same as ``emphasis'' for a single tense vowel target and ``base'' for a single lax vowel target. As such, we assessed the TTS in terms of the length of the target words rather than TTS styles and ensured no repeated phrases with the same treatment.

\subsection{Experimental procedure}

\begin{figure}[]
  \centering
  \includegraphics[width=0.8\linewidth]{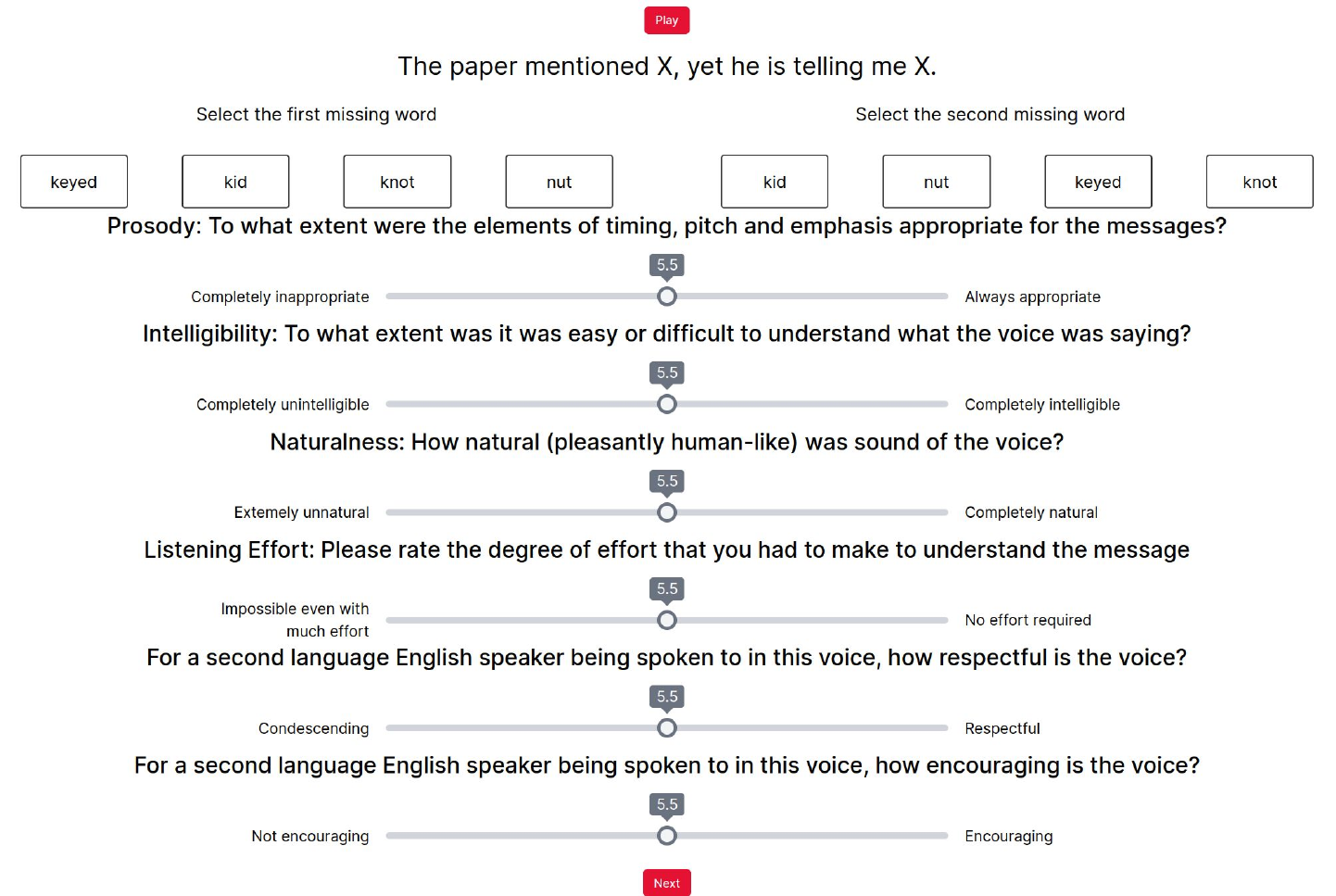}
  \caption{Screenshot of L2-TTS experiment set up in Gorilla.}
  \label{ttsexp}
\end{figure}
We conducted an experiment to assess the objective (through word error rate) and subjective (through mean opinion scores) performance of French L1 English L2 listeners using our ``clarity mode'' compared to the baseline models (i.e., ``base'', ``stretch'', ``emphasis''). In an online experiment, participants chose to receive instructions in either English or French. They then provided demographic information on the language they first learned, their most commonly used daily language, their age and gender, and their self-rated English proficiency. Each participant was randomly assigned to start in either the single- or double-word trial.

The participants were shown the phrase with the target words removed, e.g., ``Write down the word X followed by the word X on the paper,'' and could play the audio only once. They then selected which words they heard and were told (in the double-word case) that it was possible to hear the same word twice. It was never the case that they would hear the same word, however, this instruction was added to encourage participants not to base their choice for the second word on what they believe they heard in the first word. The missing word was selected from a list of 4 words: for the single-word trial, 2 words were the tense/lax vowel minimal pair (e.g., beat, bit); the third word was a minimal pair word with another vowel (e.g., bat); and the final word was dissimilar from the other three choices (e.g., shop). The dissimilar choice functioned as an attention check. In the double word case, when the two words were a minimal pair, the selection for the remaining two words was as per the single-word trial; if the two words were not a minimal pair, the 4 choices were the two words and their minimal pairs. The order of the phrases was randomized for each participant. The experiment set up can be seen in Fig. \ref{ttsexp}.

Once the participants selected which word they heard, they responded to a questionnaire containing the MOS-X2 \cite{Lewis2018InvestigatingMR} naturalness (nMOS), intelligibility (iMOS), and prosody (pMOS) scores, as well as an intelligibility question for listening effort (eMOS) (``Please rate the degree of effort you had to make to understand the message'') from MOS-X \cite{Lewis2018InvestigatingMR}. The listening effort question was added to understand if there is a difference between being able to understand the phrase and how much effort was required to understand the phrase. They then responded to two questions from L2-directed speech research \cite{ROTHERMICH201922}: ``For an English second language speaker being spoken to with this voice, how respectful is the voice?'' (Resp.: condescending - respectful), ``For an English second language speaker being spoken to with this voice, how encouraging is the voice?'' (Enc: not encouraging - encouraging). These questions were to help us understand if slowing down or adding emphasis makes the TTS sound condescending, as can be the case with L2-directed speech \cite{AOKI2024101328, ROTHERMICH201922}. All scores were on a 10-point Likert scale.

\subsubsection{Participants}
We recruited N=56 participants (29F, age = 34.59$\pm$10.67) via Prolific. Once again we maintained French as our target L2 population and the participant language demographics can be seen in Table \ref{tab:participant_demographics-tts}. Additionally, 71.4\% of participants chose to have the instructions in French.

\begin{table}[]
    \centering
    \caption{Participant language demographics for L2-TTS study}
    \begin{tabular}{lc}
        \toprule
        \textbf{Category} & \textbf{Count} \\
        \midrule
        \multicolumn{2}{l}{\textbf{Daily Language}} \\
        French & 40 \\
        English & 14\\
        French \& English & 2\\
        \midrule
        \multicolumn{2}{l}{\textbf{Self-Rated English Proficiency}} \\
        \multicolumn{2}{l}{\textbf{1(no proficiency) - 5(fluent)}} \\
        Score 2 & 2\\
        Score 3 & 5 \\
        Score 4 & 20 \\
        Score 5 & 29 \\
        \bottomrule
    \end{tabular}
    \label{tab:participant_demographics-tts}
\end{table}

\subsection{Results}
\subsubsection{Single word}

To confirm our previous findings, we expect improved performance in WER with lengthened tense vowels but decreased performance for the same treatment in lax vowels. We computed one-way ANOVAs (Type II) followed by a post-hoc Tukey HSD as well as word error rates (WER) as the proportion of incorrect target words for each type of target word treatment: ``base'' has the baseline duration on the target, ``target stretch'' is 1.6x the duration of the baseline on the target, and ``full stretch'' is 1.2x the duration on the entire phrase.

We observed that, through our ``clarity'' mode stretch applied to tense-vowel-containing words, we could overcome the bias towards lax vowels, i.e., the fact that participants were more likely to respond, for example, ``pill'' than ``peel,'' which we observed in our validation study. By lengthening the target tense-vowel-containing word, the WER on the baseline  (``base tense'') was reduced from 60.23\% to 29.48\% for ``target stretch tense'' (Table \ref{single_result}, Fig. \ref{wer-single-fig}). Importantly, we also confirmed that, indeed, stretching lax vowels, as is typical in L2-directed speech, results in more errors (28.90\% vs. 16.86\%). Lastly, the ``stretch'' TTS resulted in a slight reduction in WER over the baseline for tense vowels, yet not as much as for only emphasizing the target word. This suggests that a difference between the target word and its context was important for determining vowel perception \cite{tuttosi24_interspeech}. 

\begin{figure}[]
  \centering
  \includegraphics[width=0.75\linewidth]{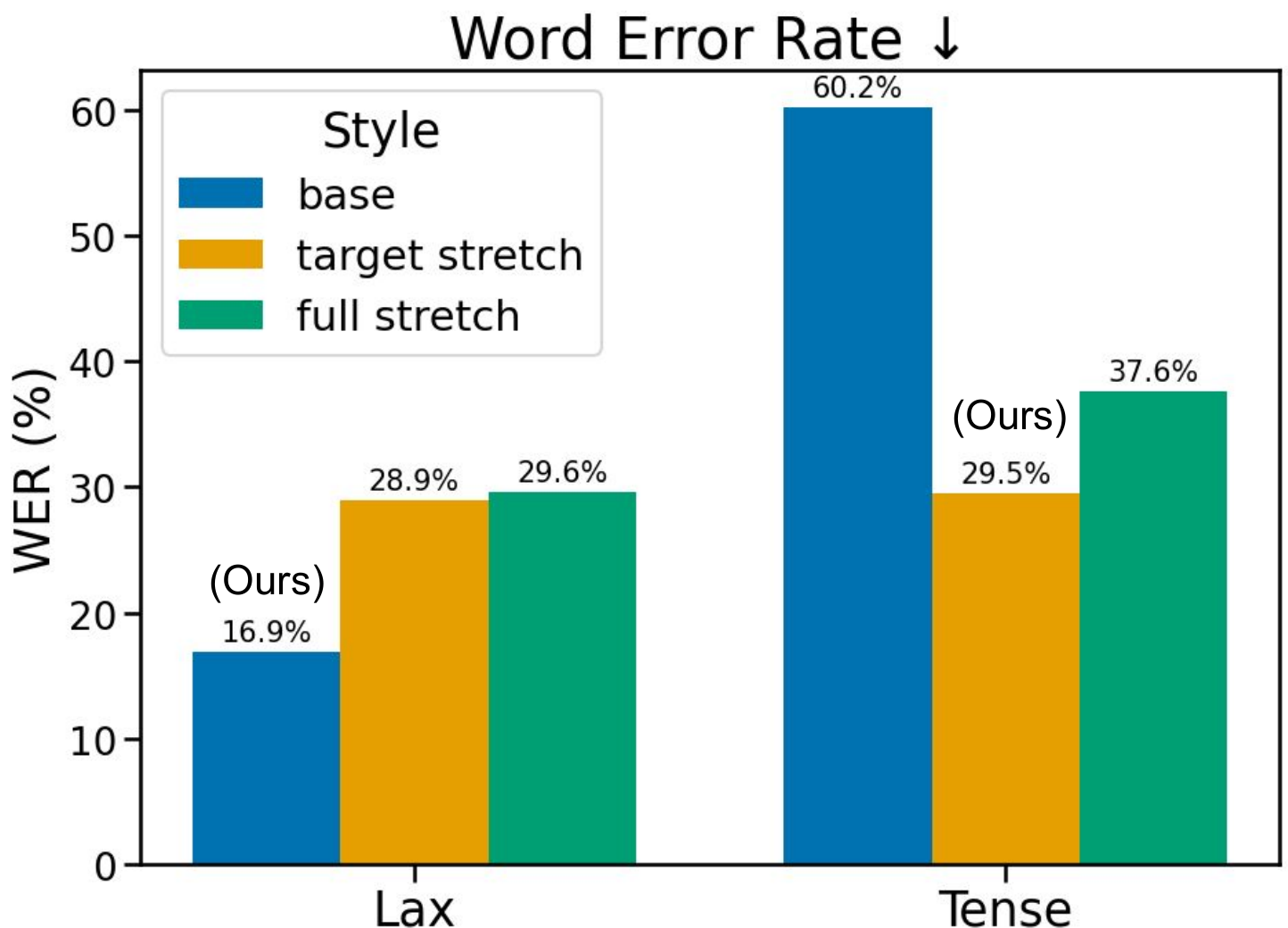}
  \caption{Single target word results: our clarity mode has the lowest WER for lax-base (left) and tense-target stretch  (right).}
  \label{wer-single-fig}
\vspace{-7mm}
\end{figure}

Surprisingly, despite the objective improvements seen in the WER, the participants perceived a ``target stretch'' target word as significantly more intelligible for lax vowels (Tables \ref{single_result} and \ref{single_stats}, Figs. \ref{mos-single-fig-lax}, \ref{mos-single-fig-tense}). Moreover, ``target stretch'' also required less listening effort (although not significantly). Furthermore, although the ``full stretch'' TTS shows improvements in WER over the baseline for tense vowels, it was found to be significantly less respectful and encouraging in all cases. We also saw that the ``full stretch'' TTS was considered to be less natural and had worse prosody than the other TTS styles. This confirms our expectation that maintaining a natural (for L1 speakers) speech rate outside of difficult words can make the L2 listeners feel less patronized by the voice. Interestingly, the ``target stretch'' lax was rated as more natural and as having better prosody than ``base'' lax vowel, once again, despite have lower objective comprehension scores. These findings are particularly interesting as they suggest that L2 speakers may be overconfident in their ability to use the formant cues when the vowel is lengthened.

\begin{table}[]
  \centering
  \caption{Single target word results: word error rate, intelligibility (iMOS), naturalness (nMOS), effort (eMOS), prosody (pMOS), encouragement (Enc) and respect (Resp.) Our clarity mode, using base lax and target stretch tense, achieves the lowest WER.}
  \vspace{-3mm}
  \resizebox{\textwidth}{!}{
  \begin{tabular}{| l | l | l | l |  l |  l |  l |  l |}
    \hline
    {\textbf{TTS Style}} & {\textbf{WER}} $\downarrow$ & {\textbf{iMOS}} $\uparrow$ & {\textbf{nMOS}} $\uparrow$ & {\textbf{eMOS}} $\uparrow$ & {\textbf{pMOS}} $\uparrow$ & {\textbf{Enc.}} $\uparrow$ & {\textbf{Resp.}} $\uparrow$\\
    \hline
    Base lax (Ours)  & \textbf{16.86}\% & 7.63  & 7.11/*** & 7.33 & 7.20/*** & 6.44/*** & 6.63/** \\
    Target stretch lax & 28.90\% & \textbf{8.18}*/** & \textbf{7.91}**/*** & \textbf{7.73} & \textbf{7.87}*/*** & \textbf{7.03}*/*** & \textbf{7.32}*/***\\
    Full stretch lax &  29.65\% & 7.53 & 4.18 & 7.40 & 5.83 & 5.54 & 5.76\\
    \hline
    Base tense  & 60.23\% & 7.57 & 7.07/***  & 7.18 & 7.22/*** & 6.45/*** & 6.53/** \\
    Target stretch tense (Ours) & \textbf{29.48}\% & \textbf{8.18}*/ & \textbf{7.71}/*** & \textbf{7.69} & \textbf{7.80}/*** & \textbf{6.84}/*** & \textbf{7.00}/***\\
    Full stretch tense &  37.57\% & 7.68 & 4.27 & 7.61 & 6.13 & 5.49 & 5.68\\
    \hline
  \end{tabular}}
    \label{single_result}
    {\parbox{6in}{
    \centering
\footnotesize 
Significance values are presented as: \{significance over long or short TTS\}/\{significance over stretch TTS\}
}}
\vspace{-3mm}
\end{table}

\begin{table*}[]
  \centering
  \caption{Single target word results: ANOVA and Tukey test statistics on MOS Likert scores.}
  \vspace{-3mm}
  \resizebox{\textwidth}{!}{
  \begin{tabular}{|  l |  l |  l  | p{9cm}|}
    \hline
    {\textbf{TTS Style}} & {\textbf{F-Statistic (5,1029)}} & {\textbf{P-Value}} & {\textbf{Significant Tukey Results}}\\
    \hline
    iMOS  &  5.12 & $<$.001 & Lax: targetS/base,fullS p=.043, p=.008; Tense: targetS/base p=.016 \\
    nMOS &  110.83 & $<$.001 & Lax: all/fullS p$<$.001 targetS/base p=.006; Tense: all/fullS p$<$.001 \\
    eMOS & 2.21 & 0.051 & N/A \\
    pMOS & 30.24 & $<$.001 & Lax: all/fullS p$<$.001 targetS/base p=.028; Tense: all/fullS p $<$.001 \\
    Enc. & 21.45 & $<$.001 & Lax: all/fullS p$<$.001 targetS/base p=.036; Tense: all/Stretch p$<$.001\\
    Resp. & 17.90 & $<$.001 & Lax: targetS/fullS,base p$<$.001, p=0.022 base/fullS p=0.001; targetS/fullS p$<$.001 base/fullS p=0.001\\
    \hline
  \end{tabular}}
    \label{single_stats}
\centering
    {\parbox{6in}{
\footnotesize 
\centering
Tukey results are presented as: \{higher value\}/\{lower value\}\\
fullS = full stretch, targetS = target stretch
}}
\vspace{-5mm}
\end{table*}

\begin{figure}[]
\vspace{-2mm}
  \centering
  \includegraphics[width=\linewidth]{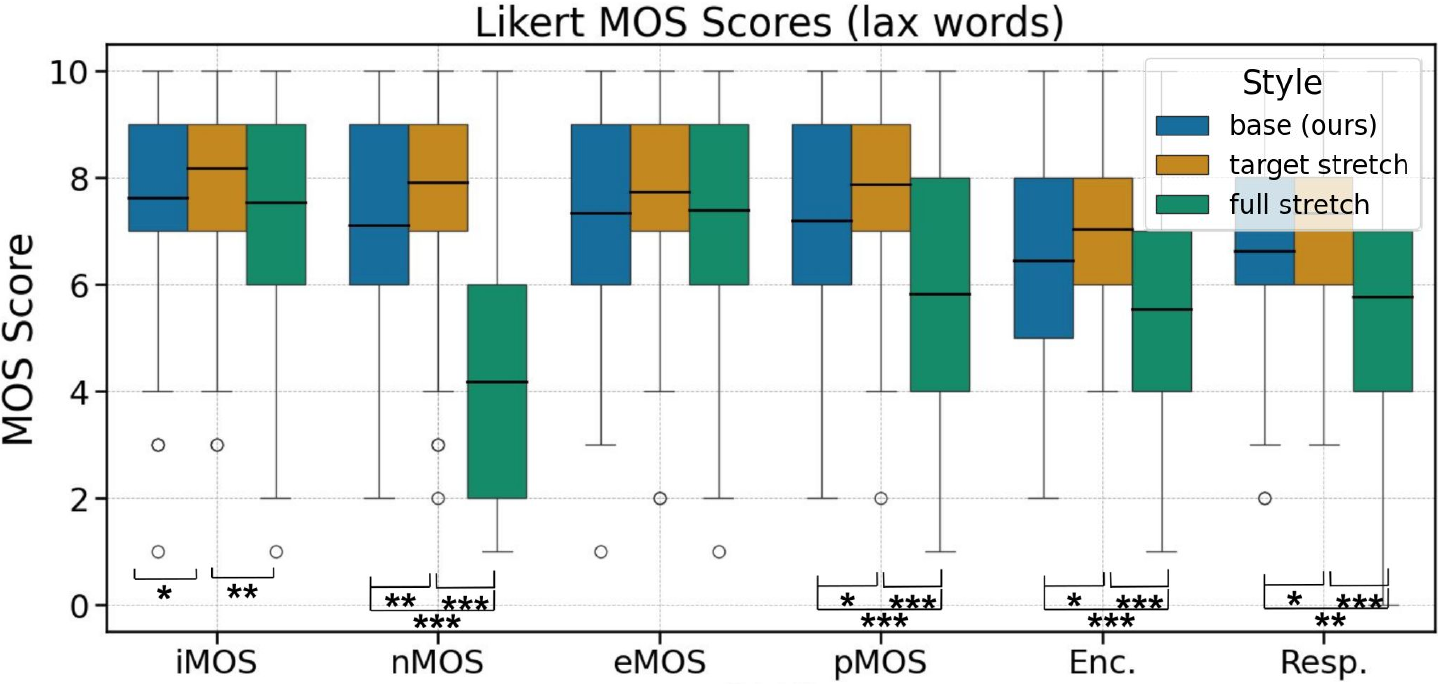}
  \caption{Single target word results: Likert MOS scores of lax vowel containing words for intelligibility (iMOS), naturalness (nMOS), effort (eMOS), prosody (pMOS), encouragement (Enc) and respect (Resp.) for base (ours), target stretch, and full stretch. Stretching the target word has significantly higher perceived intelligibility than both full stretch as the baseline, despite having lower objective comprehension performance. Both the baseline and target word stretch has significantly higher naturalness, prosody, encouragement and respectfulness over full stretch, and target word stretch over the baseline.}
  \label{mos-single-fig-lax}
\vspace{-5mm}
\end{figure}

\begin{figure}[]
\vspace{-2mm}
  \centering
  \includegraphics[width=\linewidth]{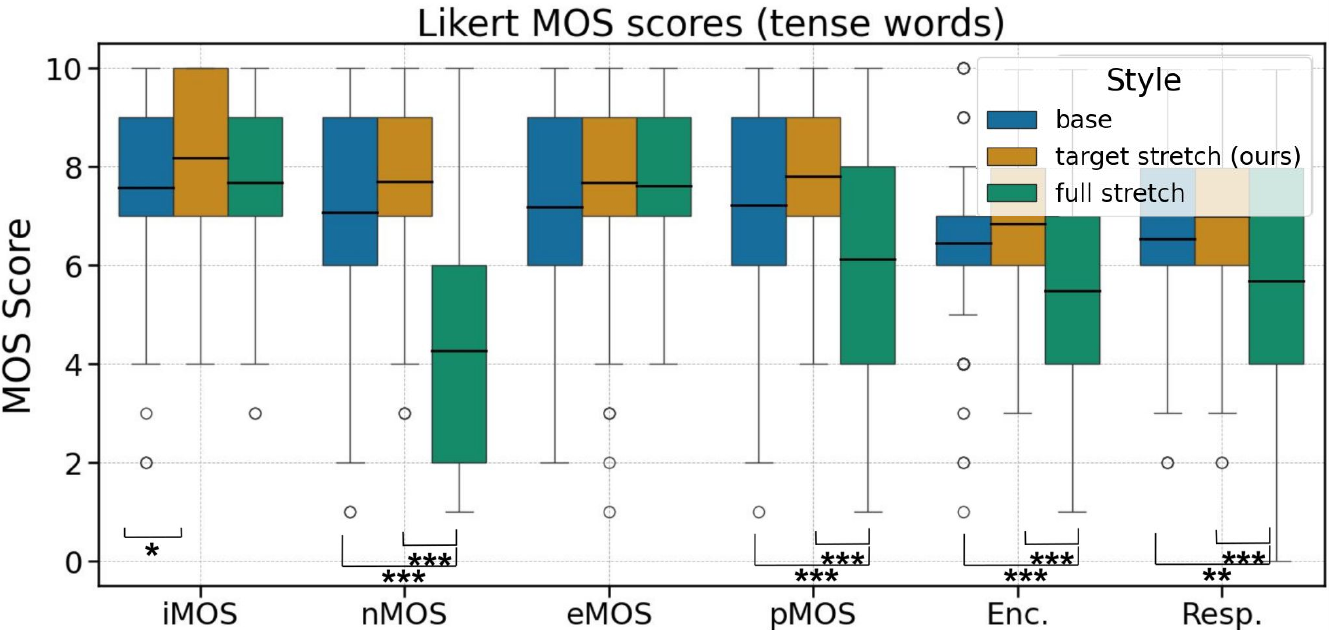}
  \caption{Single target word results: Likert MOS scores of tense vowel containing words for intelligibility (iMOS), naturalness (nMOS), effort (eMOS), prosody (pMOS), encouragement (Enc) and respect (Resp.) for base, target stretch (ours) and full stretch. Stretching the target word has significantly higher perceived intelligibility over the baseline. Both the baseline and stretching the target word have significantly higher naturalness, prosody, encouragement and respectfulness than the full stretch.}
  \label{mos-single-fig-tense}
\vspace{-5mm}
\end{figure}

\subsubsection{Double word}

Once again, we computed WER as the proportion of correct target words for each TTS style. We also assessed differences between tense (tWER) and lax (lWER) vowels.

We observed the lowest WER with our proposed ``clarity'' TTS, both overall and for lax vowels, while vastly improving the performance for tense vowels over the baseline (relative 50\% improvement) (Table \ref{double_wer}, Fig. \ref{wer-double-fig}). Again, the ``base'' TTS showed a bias towards lax vowels (tWER: 30.00\%, lWER: 18.66\%), that we were able to overcome with our ``clarity'' mode. Additionally, we see that simply emphasizing all difficult words that had a tense/lax minimal pair decreases performance in lax vowels. Lastly, we observed ``stretch'' having improved performance over both ``emphasis'' and ``base'' for tense vowels. Yet, as in the ``emphasis'' condition, we saw participants struggle to understand the short, lax vowels.  Although the tense vowel words between ``emphasis'' and ``clarity'' and lax vowel words between ``clarity'' and base had the same duration, ``clarity'' mode still had a lower WER when specifically comparing these words. This likely resulted from the phrases containing both a tense and a lax vowel, where the L2 participants could use duration differences between the two words to more easily differentiate the words. A more in-depth exploration of these differences remains a topic for future work. 

We once again observed the fascinating result that the participants rate the ``emphasis'' TTS the highest in all categories (although in this case, the scores are not significantly higher than those for the ``clarity'' TTS) (Tables \ref{double_likert} and \ref{double_stats}, and Fig. \ref{mos-double-fig}), despite the WER being higher for this TTS than both ``clarity'' and ``stretch'' TTS styles. We observed that ``stretch'' is less natural and has poorer prosody than all other TTS styles, despite showing objective improvements in WER over ``emphasis'' and ``base''. Lastly, we found that L2 participants rated both speaking too fast (``base'') and too slow (``stretch'') as being less respectful and less encouraging.

\begin{table}[]
  \centering
  \caption{Double target word results: total word error rate, tense word error rate, and lax word error rate.}
  \vspace{-1mm}
  \begin{tabular}{| l | l | l | l | }
    \hline
    {\textbf{TTS Style}} & {\textbf{WER}} $\downarrow$ &  {\textbf{tWER}} $\downarrow$ &  {\textbf{lWER}} $\downarrow$\\
    \hline
    Base  & 24.30\% & 30.00\%  & 18.66\% \\
    Stretch & 19.82\% & 17.99\% & 21.65\%\\
    Emphasis &  24.44\% & 20.06\% & 28.82\% \\
    Clarity (Ours) & \textbf{15.15}\% & \textbf{14.38}\% & \textbf{15.92}\% \\
    \hline
  \end{tabular}
    \label{double_wer}
    \vspace{-3mm}
\end{table}

\begin{figure}[]
  \centering
  \includegraphics[width=0.75\linewidth]{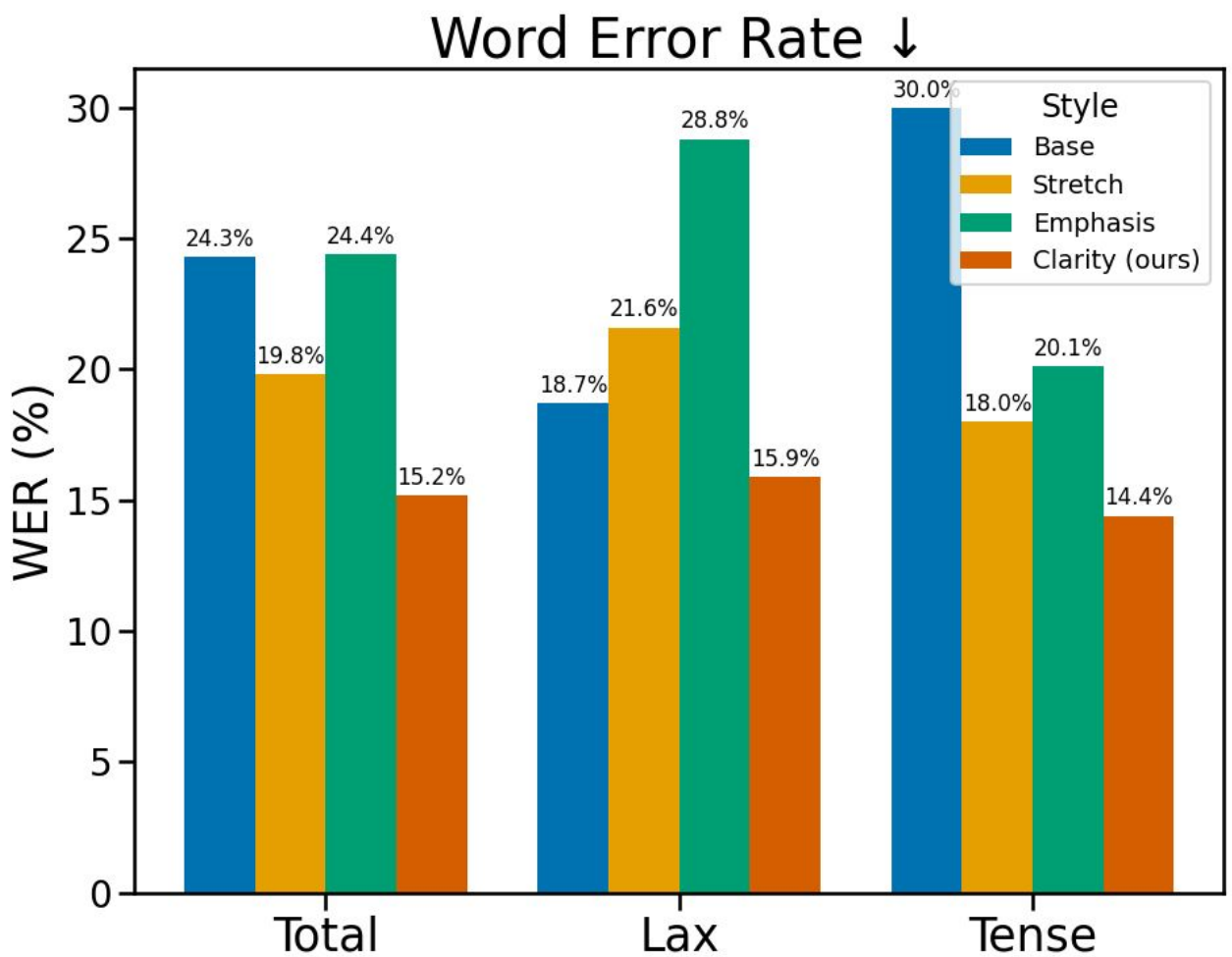}
  \caption{Double target word results: our clarity mode has the lowest total WER (left), lax WER (center) and tense WER (right).}
  \label{wer-double-fig}
\vspace{-5mm}
\end{figure}

\begin{table}[]
  \centering
  \caption{Double target word results: intelligibility, naturalness, effort, prosody, encouragement and respect.}
  \vspace{-3mm}
  \resizebox{\textwidth}{!}{
  \begin{tabular}{|  l |  l |  l |  l | l | l | l |}
    \hline
    {\textbf{TTS Style}} & {\textbf{iMOS}} $\uparrow$ & {\textbf{nMOS}} $\uparrow$ & {\textbf{eMOS}} $\uparrow$ & {\textbf{pMOS}} $\uparrow$ & {\textbf{Enc.}} $\uparrow$ & {\textbf{Resp.}} $\uparrow$\\
    \hline
    Base  &  7.30 & 6.95/*** & 6.60 & 7.16/*** & 6.25 & 6.70 \\
    Stretch & 7.94***/ & 4.93 & \textbf{7.53}***/ & 6.40 & 6.02 & 6.46\\
    Emphasis & \textbf{8.06}***/ & \textbf{7.71}***/*** & 7.43***/ & \textbf{7.98}***/*** & \textbf{6.97}***/*** & \textbf{7.42}***/***\\
    Clarity (Ours) & 7.83***/ & 7.54***/*** & 7.25***/ & 7.77***/*** & 6.72***/*** & 7.16***/***\\
    \hline
  \end{tabular}}
    \label{double_likert}
    {\parbox{6in}{
\footnotesize 
\centering
Significance values are presented as: \{significance over base TTS\}/\{significance over stretch TTS\}
}}
\vspace{-3mm}
\end{table}

\begin{table}[]
  \centering
  \caption{Double target word results: ANOVA and Tukey test statistics on MOS Likert scores.}
  \vspace{-3mm}
  \resizebox{\textwidth}{!}{
  \begin{tabular}{|  l |  l |  l |  l |}
    \hline
    {\textbf{TTS Style}} & {\textbf{F-Statistic (3,2504)}} & {\textbf{P-Value}} & {\textbf{Significant Tukey Results}}\\
    \hline
    iMOS  &  20.15 & $<$.001 & all/Base p$<$.001 \\
    nMOS &  20.15 & $<$.001 & all/Stretch p$<$.001; Emphasis,Clarity/Base p$<$.001 \\
    eMOS & 24.15 & $<$.001 & all/Base p$<$.001 \\
    pMOS & 77.15 & $<$.001 & all/Stretch $<$.001; Emphasis,Clarity/Base p$<$.001 \\
    Enc. & 35.74 & $<$.001 & Emphasis,Clarity/Stretch,Base p$<$.001 \\
    Resp. & 35.87 & $<$.001 & Emphasis,Clarity/Stretch, Base p$<$.001 \\
    \hline
  \end{tabular}}
    \label{double_stats}
\centering
    {\parbox{6in}{
\footnotesize 
\centering
Tukey results are presented as: \{higher value\}/\{lower value\}
}}
\vspace{-5mm}
\end{table}

\begin{figure}[]
\vspace{-1mm}
  \centering
  \includegraphics[width=\linewidth]{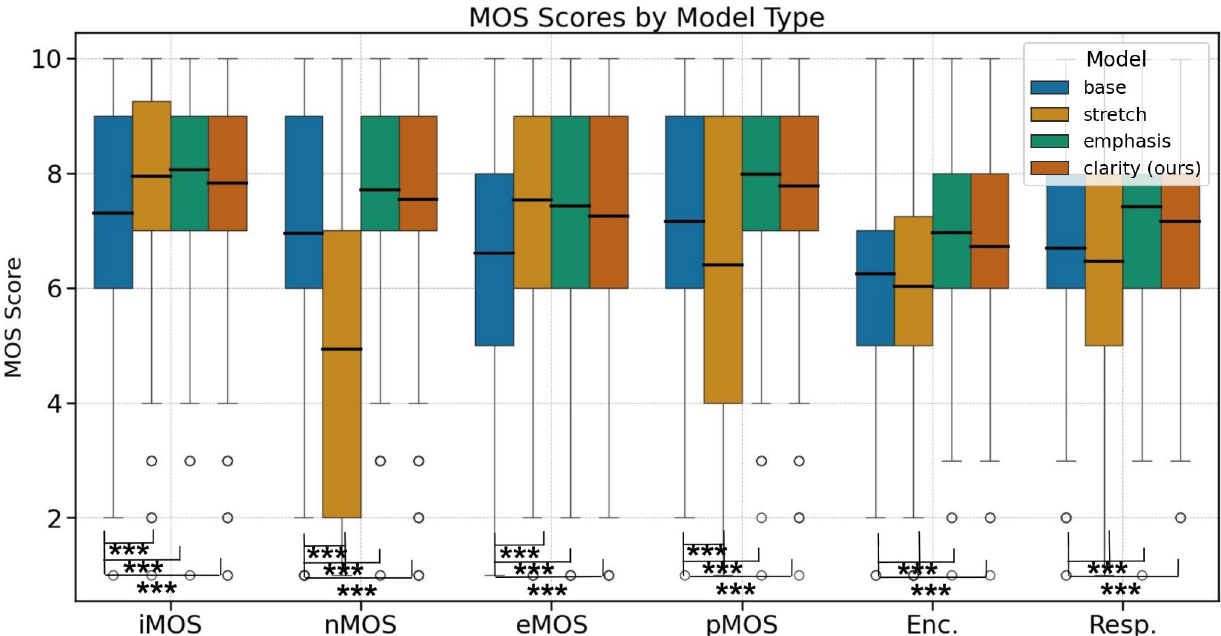}
  \caption{Double target word results: Likert MOS scores of words containing tense vowels for intelligibility (iMOS), naturalness (nMOS), effort (eMOS), prosody (pMOS), encouragement (Enc) and respect (Resp.). The baseline is perceived as  significantly less intelligible and requiring more listening effort than all of stretch, emphasis and clarity. Baseline and stretch are perceived as significantly less natural  with worse prosody than both emphasis and clarity, with stretch lower than baseline. Both baseline and stretch (speaking too fast or too slow) are perceived as significantly less encouraging and respectful.}
  \label{mos-double-fig}
\vspace{-5mm}
\end{figure}

\subsubsection{Whisper ASR}
Speech synthesis studies often use human MOS scores but rely on ASR for transcription accuracy, as it correlates well with L1 intelligibility \cite{taylor21_interspeech}. In this section, we explore how ASR relates to L2 performance and whether it uses the same duration cues as L2 participants.

We used Whisper ASR \cite{whisper2023}\footnote{v20231117, medium multilingual} with 72 phrases (generated in the same manner as for the human experiments) and calculated overall WER (WERt) and WER on only the target words. The phrases included those from the human study, and the additional phrases were constructed as in Sec. \ref{gen}. We also included the percentage of errors in the target word resulting from minimal tense/lax pair substitution (sub) and what percentage of these substitutions were lax substituted for a tense (t-sub) and tense substituted for a lax (l-sub) vowel. Additionally, we ensured that homophones with the target words were accepted as correct transcriptions.

Similar to L2 speakers, we saw for the ``base'' TTS, Whisper struggled to predict the correct target words that lack context in the phrase (21.4\% vs an expected 5-10\% WER for this model), although the performance was slightly higher than that for the L2 speakers (Table \ref{whisper_wer}, Fig \ref{wer-asr-fig}). We did not see the same improvements in WER with the ``clarity'' TTS that we saw in the L2 participants. Instead, we saw an overall slowing down (``stretch'') of the TTS decreases the WER in the target words. Yet, we also saw that while the WER on the target words decreased with this slowing down, the proportion of errors in the target word stemming from the minimal pair substitutions was much higher for the ``base'' TTS (71.42\%, all other TTS $<$ 37\%), while the overall difference in WER both on the whole phrase and the target words was within 3\% for all TTS styles. This suggests that while slowing down slightly reduces the overall number of errors, the target words were being predicted as even further from the target, e.g. where ``peel'' was replaced with ``pill'' in the ``base'' TTS was replaced with ``peaked'' in the ``stretch'' TTS. Therefore, the ASR does not use the same duration mechanisms as humans when facing difficult predicting words.

\begin{table}[]
  \centering
  \caption{Whisper ASR results: overall word error rate, target word error rate, tense/lax substitutions, lax substituted for a tense, tense substituted for a lax}
    \vspace{-2mm}
  \footnotesize
  \begin{tabular}{| l | l | l | l | l | l | }
    \hline
    {\textbf{TTS Style}} & {\textbf{WERt}} $\downarrow$  & {\textbf{WER}} $\downarrow$ &  {\textbf{sub}} & {\textbf{t-sub}} &  {\textbf{l-sub}}\\
    \hline
    Base  & 17.10\% & 21.4\% & 71.42\% & 61.9\%  & 9.52\% \\
    Stretch & \textbf{15.98}\% & \textbf{19.38}\% & 36.83\% & 31.57\% & 5.26\%\\
    Emphasis & 16.26\% & 22.4\% & 31.81\% & 22.72\% & 9.09\% \\
    Clarity (Ours) & 17.68\% & 24.49\% & 29.16\%  & 29.16\% & 0\% \\
    \hline
  \end{tabular}
    \label{whisper_wer}
    \vspace{-2mm}
\end{table}

\begin{figure}[]
  \centering
  \includegraphics[width=0.75\linewidth]{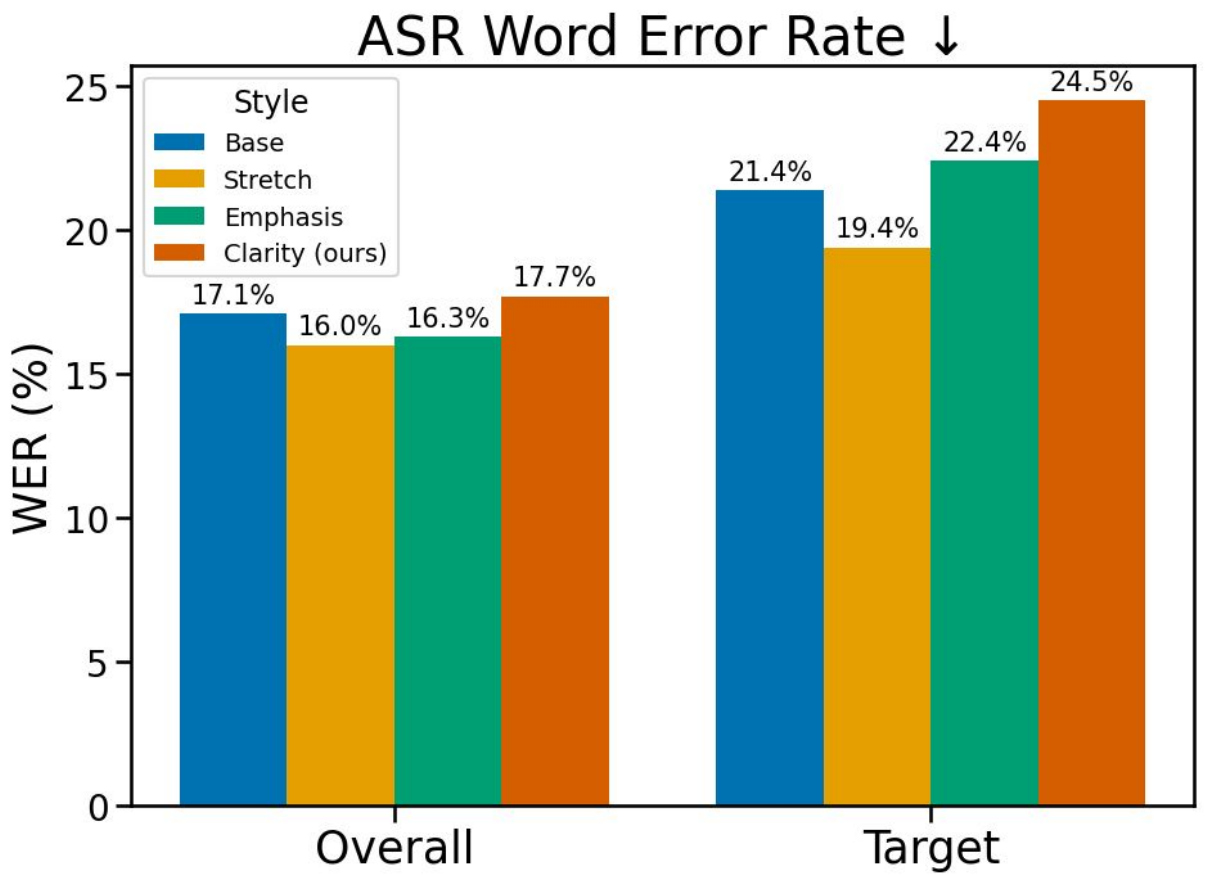}
  \caption{Whisper ASR results: We observe that ASR does not align with L2 perception (Fig. \ref{wer-double-fig}). Stretch had the lowest WER both for the whole phrase (left) and the target words (right).}
  \label{wer-asr-fig}
\vspace{-5mm}
\end{figure}

\subsection{Discussion}
This study resulted in multiple interesting findings. First, we provided a ``clarity mode'' as an open-source addition to Matcha-TTS and confirmed that, by applying a stretch to tense vowels for difficult target words, L2 speakers' transcription errors for these words were reduced. Indeed, emphasizing all target words or simply slowing the whole phrase without consideration of the linguistic properties of the vowel reduced the transcription performance of L2 speakers (from French L1 backgrounds), who primarily use duration to determine the difference between English tense and lax vowels. Moreover, we confirm prior findings that slowing down an entire phrase can be seen as less respectful and encouraging to L2 listeners. 

Second, we found that our sample of L2 speakers are unaware they are using this duration mechanism. Through MOS scores, these L2 speakers indicated that they could more easily identify a stretched word, perhaps because they believed they could use the longer vowels (more stable formants) to transcribe a word more easily. This suggests that L2 listeners are limited in their ability to evaluate the effectiveness of adaptive TTS systems, and that objective metrics such as word error rate should serve as a gold standard in future work.

Lastly, we found that Whisper ASR does not use the same duration mechanisms as L2 speakers and, therefore, does not present an adequate replacement for determining the transcription accuracy of synthesized speech for these individuals. This suggests, again, that future development of adaptive TTS systems for L2 speakers must, at the point in time, rely more on data from real human listeners.

Overall, these results have important consequences for TTS assessments, especially for L2 speakers. We cannot simply rely on the same methods used for L1 speaker TTS assessments. Neither user self-rated intelligibility assessments nor automatic systems reflect the true accuracy of difficult vowel perception in L2 speakers, even for those with high proficiency as in our sample, and improved accuracy does not necessarily reflect positive perceptions of the voice (as for ``stretch''). Researchers must use both objective and subjective assessments with human participants to ensure they are building inclusive and accessible speech synthesis systems.

\chapter{Conclusions and Future Directions}
\section{Previous Recommendations For Robot Voices}\label{rec}
In this thesis we made an effort to adhere to the recommendations for spoken language interaction with robots as presented in \cite{MARGE2022101255}. There are 25 recommendations organized into seven categories: user experience design; audio processing, speech recognition, and language understanding; speech synthesis and language generation; dialogue; other sensory processes; robustness and adaptability; and infrastructure. We reflect on which relevant recommendations we aimed to achieve in this thesis.

In the category of \textbf{user experience design}, we met all but the final recommendation for multi-party interaction. \emph{Focus on language not only as a way to achieve human-like behaviors but also as a way to support limited but highly usable communications abilities.}--We explored and provided recommendations for use cases for our voices both in terms of tasks (Ch. 3 and 5) and physical and social environments (Ch. 4). \emph{Deliberately engineer user perceptions and expectations.}--We focused on making minimal changes to TTS systems and focused on the task at hand, e.g., expressivity (Ch. 3), rather than building natural and human-like voices. This allowed us to ensure we meet our goals for HRI without enforcing unrealistic expectations from the robot. \emph{Work to better characterize the list of communicative competencies most needed for robots in various scenarios.}--Through this work, we strove to understand and define the need for clear and comprehensive voices in specific use cases and for specific users.

We pushed forward all the recommendations for \textbf{speech synthesis and language generation}. \emph{Develop the ability to tune speech synthesizers to convey a desired tone, personality, and identity.}--One of the reasons we have chosen Matcha-TTS is because it can be easily controlled for clarity and fine-tuned for expressivity. \emph{Extend the pragmatic repertoire of speech synthesizers.}--We Introduced the use of emojis as an interpretable way to control speech and to represent a robot's internal state (Ch. 3). We also made use of known linguistic properties to improve L2 comprehension (Ch. 5). \emph{Create synthesizers that support real-time control of the voice.}--Once again, Matcha-TTS was chosen for its speed and real-time synthesis abilities. \emph{Develop speech generators that support multimodal interaction.}--In expressive case studies, we have chosen animations that support the communication of expression in our voices (Ch.3).

We addressed all recommendations for \textbf{dialogue}. \emph{Focus on highly interactive dialogue.}--We have designed both of our conversational case studies to have short sentences to maintain the flow of the dialogue and user engagement (Ch. 3 and 4). \emph{Make every component able to support real-time responsiveness.}--In our speech-to-speech agent, we made an effort to have each component run in real-time (Ch. 3). However, there is still room for improvement; for further information, see Section \ref{s2s-model}.

Recommendations for other dialogue, sensory processes, robustness and adaptability, audio processing, speech recognition, language understanding, and infrastructure are outside this research's scope and remain open research questions for the community.

\section{Summary}
Our goal was to create a voice that can be used to teach second languages in HRI applications. To achieve this, we first identified a lightweight TTS that could reliably run in real-time, without hallucinations, on a robot platform: Matcha-TTS. We then modified Matcha-TTS to take an emoji prompt to make a temporally expressive TTS to help improve engagement in expressive teaching tasks. We found that this voice is best applied in long-term expressive speech where the variability in the expression is best utilized. Moreover, we provided a toolbox that helps other HRI researchers quickly and easily build their own expressive TTS to their own use case with only 3 minutes of data per expressive style. At the same time, we explored how we could make the voice appropriate and understandable in different social and physical ambient contexts so that the voice can be used in a variety of environments. First, we found that by selecting the correct voice for the environment, we could increase the perception of awareness and appropriateness in the robot. Additionally, for loud environments, a higher pitch could be used as a substitute to increase clarity, where increases in intensity may cause saturation issues in a robot's speakers. Lastly, unlike in human Lombard speech, in a high-energy (e.g., expressive) social ambiance, a large pitch range was preferred alongside this increase in pitch. Finally, we aimed to understand how, from a perception-based perspective, we could improve a TTS for second language comprehension. We explored known linguistic properties of tense and lax vowels in English that are known to be difficult for L2 speakers. We found that, indeed, we could use lengthening of long-tense vowels to aid L2 speakers to avoid confusing these words with their lax minimal pair. In addition, we found that more work needs to be done to make inclusive speech synthesis systems, and in order to assess these systems, we must have humans with a variety of comprehension abilities provide both objective and subjective measures rather than relying on automatic assessments and purely first language speakers.

\section{Future Work}
Future work is required to understand how to combine each of these TTS modifications effectively (see Sec. \ref{combine}). In addition, the robustness of each of the methods themselves merit future work.

\subsection{Expressive voice}
For EmojiVoice, we observed, both through the data recording process and the case studies, that although the system seems to be appropriate for long-form interactions with variations from phrase to phrase, we still observe awkwardness in very long phrases. Indeed, humans often will use multiple expressive tones over the course of a long phrase. Our system is limited to one emoji per phrase, and it remains future work to implement, in real-time, multiple tones into a single sentence similar to \cite{storytts}, but in a controllable manner. Moreover, in early explorations, it appeared that this effect was stronger for emojis that may focus on a particular word rather than on an entire phrase, e.g., a ``wink'' emoji. Future work should explore how emojis can be used as a form of markup for emphasis and tone around specific words in a phrase.

In addition, we found that the high intensity of some of the emotions can cause over-saturation on the robot speakers, which is not noticed when testing the voices through headphones. This was particularly evident in the Miroka trials and may have impacted the results, and is evidence for the importance of testing voices for HRI in person on real robots \cite{wang24t_interspeech}. The training data was normalized for intensity and new checkpoints are available, eliminating this problem of saturation.

Interestingly, for both case studies, we had 3 participants who preferred the Baseline voice for the interactions and stated that they preferred a flat, robotic-sounding voice for a robot. This shows us that, despite the task, some individuals still want a classic, robotic voice for a robot embodiment. Moreover, all three of these individuals were male, and in general, the male participants seemed to provide lower scores to the expressive voices. Future work should be done to better understand gender preferences for expressive voices and tasks.

\subsection{Ambient appropriate voice}
Although we saw significant improvements in perceptions, most of these were an increase of less than a Likert scale point. Moreover, the average of all of our Likert ratings (social appropriateness, ambiance awareness, comfort, human-likeness, competency) fell between somewhat disagree and agree, suggesting that humans are not perceiving large changes in appropriateness of a voice and are still somewhat uneasy and unopinionated given a robot voice. Further work must be done to expand our abilities to synthesize speech that is adaptable and appropriate if we hope to integrate robots into a multitude of varying everyday spaces. 

Real-time systems able to adapt a voice using a sensing-production loop (such as is \cite{renclarity}) are an exciting area for future study; for instance, the robot may assess the distribution of frequencies in the ambiance and adapt its voice to avoid the frequency ranges already filled by background noise. Furthermore, the use of virtual reality could be explored to overcome the limitations in simulating the closed/open characteristics of space, which were not well differentiated in our perception study.

\subsection{L2 directed TTS}
Although we were able to increase the comprehension of difficult words for L2 speakers, our work remains focused on duration as a cue for tense/lax minimal pairs. Further investigations should be launched into how to increase clarity for other vowels, including manipulations to spectral cues, like formants, and later, extending this approach to consonants. For example, manipulating pauses and stress to cue a difficult word may aid L2 speakers to pay attention and hence increase comprehension \cite{pausefocus}.

Further, through explorations of the results on the participant level in our validation study, we found that the duration mechanism was not clearly universal; rather, it appears to be very strong in certain participants and weak or non-existent in others, which echoes work showing inter-individual differences in cue-weighting \cite{schertz2015individual}. Future work may harness pronunciation data to further customize TTS. Moreover, since our work used forced choice to limit participant responses, a topic of future work is understanding how increasing clarity in a target word affects the rest of the phrase and how robust this duration mechanism is to broader linguistic contexts. Additionally, since our participants had a high level of English proficiency, future work should explore ``L2 clarity TTS'' with participants of lower proficiency levels, which could explore the modification of words that are less likely to be in the vocabulary of L2 speakers (e.g., ``cooed'' vs. ``could''), using individual vocabularies as a factor predicting perception. Lastly, even more fine-grained control over a TTS, such as in \cite{tannander24_interspeech}, could aid in uncovering other possible mechanisms to aid L2 perception, specifically for how formant control and duration control can be combined.

\subsection{An ambient-adaptive expressive L2 teaching voice} \label{combine}
Each of these elements has been studied individually, and a demo has been created to incorporate all of them simultaneously. In this demo, the robot reads short stories for children using EmojiVoice, and the user can select to hear the voices in L2 clarity mode. Moreover, the pitch increases as the room's noise levels increase. However, an HRI study has not been run to test the interaction of each of these elements in the voice. This remains important future work. For example, when the pitch is increased to maintain clarity in a noisy environment, how does this affect the expressive voice that is already at a higher pitch than a neutral voice? Perhaps this increase needs to be more subtle and more quickly reach a ceiling when using expressive speech. Moreover, we interestingly noticed that the pitch tends to decrease when the duration lengthening is applied in L2 clarity mode. It is unclear if this is a physical function of the audio manipulation, or rather, if the slower training data also exhibited lower pitch and the model is mimicking slow human voices. As we did not find that pitch played a clear role in manipulating user perception, it remains interesting future work to understand what role, if any, this pitch modification plays in vowel comprehension. Therefore, the next question is how the increase in pitch to reflect environmental clarity interacts with the duration changes for L2 clarity. Lastly, the interaction between the expressive voice and the L2 clarity must be better understood. Each emoji styling has its own unique speech rate, and it may be possible that the L2 clarity duration changes need to change relative to this base speaking rate rather than maintaining a constant multiplicative. For example, if angry speech has a faster speech rate, perhaps a lengthening of 1.6x is not enough to increase clarity in difficult vowels.

%
%
%
%
%

\backmatter%
    \addtoToC{Bibliography}
    \bibliographystyle{IEEEtran}
    \bibliography{references}

\begin{appendices} 

\chapter{Ambiance Zoom Experiment Script}\label{ambiance-script}

\captionsetup[figure]{list=no}
\begin{figure}[h]
  \centering
  \includegraphics[width=\linewidth]{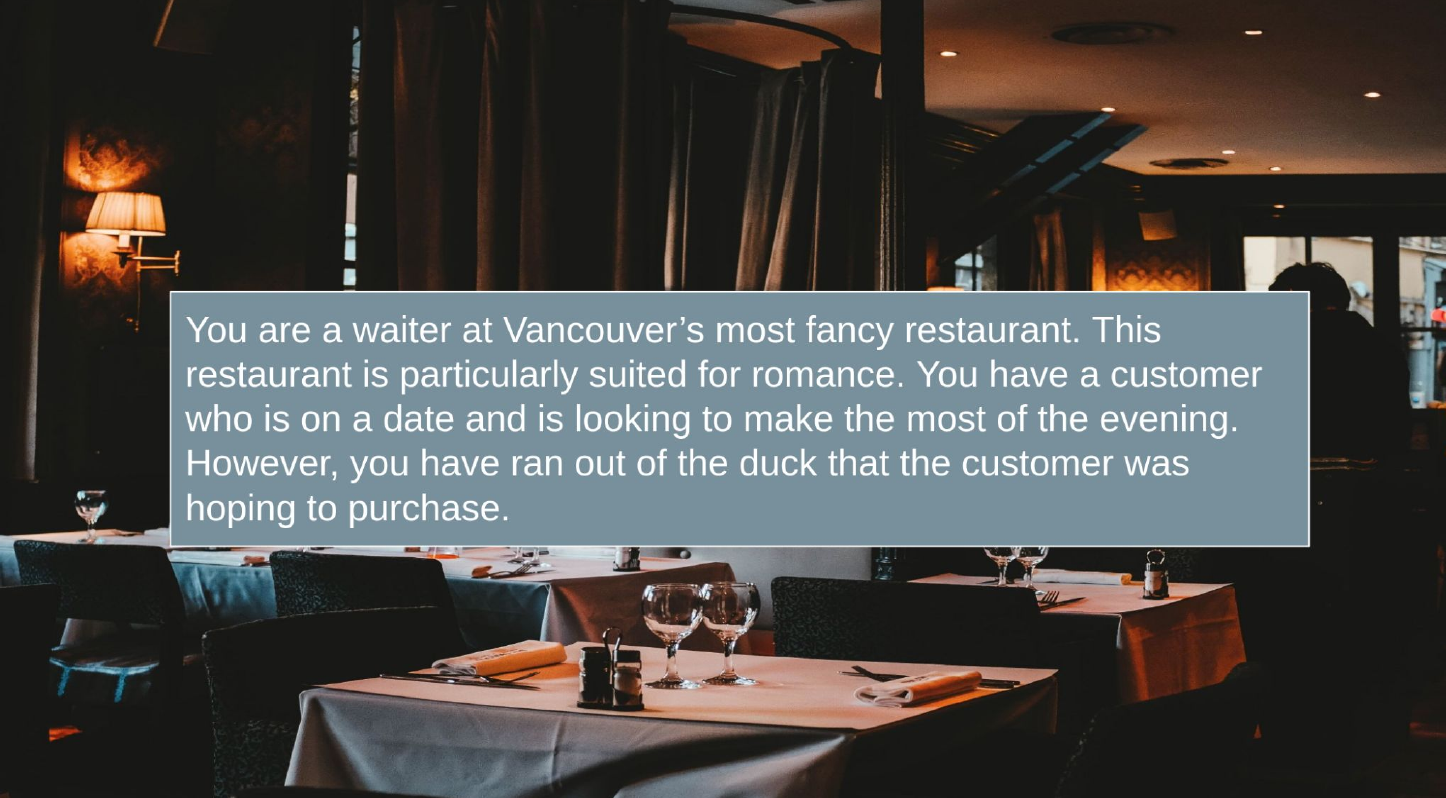}
  \caption{Zoom background image and prompt for the ``fancy restaurant'' ambiance in the Chapter 3.}
\end{figure}

Waiter: Hi there, I hope you’re doing well. I hope you have had a chance to look at the menu. I thought I would let you know, today we do have steak frites as our special. Anyways what can I start you off with?

Customer: Can I get a bottle of your finest Prosecco?

Waiter: Sure! I can definitely do that!

Customer: Wait! Can you also make sure the Prosecco is nice and cold and served in champagne glasses?  

Waiter: Of course, we can definitely do that. Would you also like some water for the table?

Customer: Hmm. What kind of water do you have?

Waiter: We have still, sparkling, and tap water.

Customer:  We will go with the sparking, please!

Waiter: Alrighty, then! I’ll be right back with your Prosecco with two champagne glasses and sparkling water!

A few minutes pass by...

Waiter comes and brings customer order...
Waiter: Hi again, here is your prosecco. Are you both ready to order food?

Customer: Thanks. I think we would both like to have the Duck? 

Waiter: Oh no, unfortunately we just ran out of the Duck! Is there anything else that you see?

Customer: How do you not have any Duck? That's your staple item?

Waiter: Uh, it has been a pretty hectic evening; busier than normal!

Customer: Hmm. Fine, I guess we'll just have the steak frites then.

A few minutes pass by...

Waiter returns with order.
Waiter: Here are your steak frites. I hope you enjoy. Let me know if there is anything else I can get you.

Customer: I am not going to lie to you, I would have much rather preferred the Duck. But the frites is good enough.

Waiter:  I am so sorry we couldn’t get you the Duck. I am sure next time we will have more available.

Customer: Ya, it is a bit odd. Usually I thought businesses like this are more prepared and make enough items just in case. 

Anyways, can we get the bill?

Waiter: For sure, I will be right back with your bill!

\chapter{Battery for Speaker Ambiance Validation study}\label{battery}
\begin{itemize}
\footnotesize 
    \item This person's voice is socially appropriate for the scene.
    \item This person knows how they should present themselves in this ambient context.
    \item This person is aware of the surrounding ambiance. 
    \item This voice sounds like it was originally recorded in this ambiance. 
    \item This person knows how to adapt their voice this ambiance. 
    \item This person makes me feel comfortable. 
    \item I feel  uneasy listening to this person. 
    \item I feel at easy when conversing with this person. 
    \item It took a great deal of effort to understand what this person was saying.
    \item Certain words were difficult to understand.
    \item This person's speech was very clear. 
\end{itemize}

\chapter{L2 TTS Experiment Word List}\label{wordlist}

The word cut seemed important to the instructions.
She kept mentioning cot during the conversation.
The speaker mentioned pill, or at least something similar.
The word pill was what she was trying to write.
The phrase had fool somewhere in the middle of it.
I saw full written on the note pad.
The sign mentioned sin, but the person said scene.
He wrote down bought, but remembered it as but.
In his talk he kept using could, but I am pretty sure he meant cooed.
The paper mentioned kid, yet he is telling me knot.
There was confusion between pull and bean in their speech.
I am not sure if the word was pool or if cup was the right one.
Sheep goes on the top of the page and dull goes on the bottom.
Bit was the first word he said, then nut followed.
Actually hut is the correct word, it was replaced with should by accident.
Maybe he said hot, but I really thought keyed was what he said.
Reap was a more important word in the story than wooed.

\chapter{L2 TTS Experiment Word List - Whisper ASR}\label{wordlistwhisper}

Ship was what I heard, but maybe I misunderstood.\\
Sheep might have been what they meant, but it wasn’t clear.\\
Pool was the word I caught, though it could’ve been something else.\\
Pull might have been what they were talking about, but I’m unsure\\
Cut was the word used, but it seemed odd in context.\\
Cot could have been the intention, but I wasn’t sure.\\
I thought they said something about peel.\\
I think the phrase ended with full.\\
The word on the sign was pill.\\
No, the text message definitely said fool.\\
I thought they ended their sentence with rut.\\
It sounded like the last word they said was rot.\\
The word bean seemed important to the instructions.\\
She kept mentioning bin during the conversation.\\
The speaker mentioned wood, or at least something similar.\\
The word wooed was what she was trying to write.\\
The phrase had dull somewhere in the middle of it.\\
I saw doll written on the note pad.\\
Doll was what I heard, but maybe I misunderstood.\\
Dull might have been what they meant, but it wasn’t clear.\\
Wood was the word I caught, though it could’ve been something else.\\
Wooed might’ve been what they were talking about, but I’m unsure.\\
Bean was the word used, but it seemed odd in context.\\
Bin could’ve been the intention, but I wasn’t sure.\\
I thought they said something about rot.\\
I think the phrase ended with rut.\\
The word on the sign was ship.\\
No, the text message definitely said sheep.\\
I thought they ended their sentence with pool.\\
It sounded like the last word they said was pull.\\
The word cut seemed important to the instructions.\\
She kept mentioning cot during the conversation.\\
The speaker mentioned peel, or at least something similar.\\
The word pill was what she was trying to write.\\
The phrase had fool somewhere in the middle of it.\\
I saw full written on the note pad.\\
The sign mentioned sin, but the person said scene.\\
The sign mentioned scene, but the person said sin.\\
There was confusion between cup and cop in their speech.\\
There was confusion between cop and cup in their speech.\\
He wrote down luke, but remembered it as look.\\
He wrote down look, but remembered it as luke.\\
The paper mentioned bit, yet he is telling me knot.\\
The paper mentioned beat, yet he is telling me nut.\\
I am not sure if the word was could or if keyed was the right one.\\
I am not sure if the word was cooed or if kid was the right one.\\
Now note down heat and the word hut on the bottom.\\
Now note down hit and the word hot and the bottom.\\
The sign mentioned could, but the person said cooed.\\
The sign mentioned cooed, but the person said could.\\
There was confusion between kid and keyed in their speech.\\
There was confusion between keyed and kid in their speech.\\
He wrote down hut, but remembered it as hot.\\
He wrote down hot, but remembered it as hut.\\
The paper mentioned nut, yet he is telling me heat.\\
The paper mentioned knot, yet he is telling me hit.\\
I am not sure if the word was scene or if cup was the right one.\\
I am not sure if the word was sin or if cop was the right one.\\
Now note down beat and the word look on the bottom.\\
Now note down bit and the word luke on the bottom.\\
He whispered about hearing someone say pill and cup nearby.\\
He whispered about hearing someone say peel and cop nearby.\\
The presenter said dull but I think he meant but.\\
The presenter said doll but I think he meant bought.\\
Please write down the words reap and rot on your paper.\\
Please write down the words rip and rut on your paper.\\
He whispered about hearing someone say but and rip nearby.\\
He whispered about hearing someone say bought and reap nearby.\\
The presenter said peel but I think he meant rot.\\
The presenter said pill but I think he meant rut.\\
Please write down the words cup and dull on your paper.\\
Please write down the words cop and doll on your paper.\\


\end{appendices}
\end{document}